\documentclass[11pt]{article}

\usepackage[preprint]{acl}

\usepackage{times}
\usepackage{latexsym}

\usepackage[T1]{fontenc}

\usepackage[utf8]{inputenc}

\usepackage{microtype}

\usepackage{inconsolata}

\usepackage{graphicx}
\usepackage{amsmath}
\usepackage{tabularx}
\usepackage{array}
\usepackage{multirow}
\usepackage{enumitem}
\usepackage{xcolor}
\usepackage{tabularx}
\usepackage{makecell}

\definecolor{mairuigreen}{rgb}{0.0,0.5,0.0}

\title{The Asymmetric Harms of LLM Compression}

\author{
  Yuan Wu\thanks{Equal contribution.} \\
  Rice University \\
  \texttt{yw223@rice.edu}
  \And
  Mairui Li\footnotemark[1] \\
  University of North Carolina at Chapel Hill \\
  \texttt{mairuili@unc.edu}
  \AND
  Lesia Semenova \\
  Rutgers University \\
  \texttt{lesia.semenova@rutgers.edu}
  \And
  Chudi Zhong \\
  University of North Carolina at Chapel Hill \\
  \texttt{chudi@unc.edu}
}

\begin{document}
\maketitle
\begin{abstract}
Large language models (LLMs) compression reduces deployment costs, but standard aggregate metrics like perplexity and  accuracy often mask underlying behavioral shifts. In this work, we systematically evaluate 3 LLMs across 11 compression methods to investigate the effects of compression on knowledge retention, model confidence, and social bias. We find that compression disproportionately reduces the relative retention of head knowledge compared to tail knowledge. Furthermore, compressed models often remain substantially confident in their incorrect answers on newly lost knowledge. Finally, we demonstrate that stable aggregate bias scores can conceal substantial, opposing shifts in stereotypical preferences across demographic subgroups. Together, these findings reveal asymmetric behavioral changes that aggregate performance measures fail to capture, highlighting the need for granular evaluation of compressed models before deployment.

\end{abstract}

\section{Introduction}

Large language models (LLMs) are increasingly deployed under tight memory,
latency, and energy budgets, and model compression has become a standard
tool for meeting these constraints. Specifically, two families of methods dominate
practice: quantization, which stores weights and activations at reduced
precision~\cite{frantar2023gptq,lin2026awq,shao2024omniquant,pmlr-v235-egiazarian24a};
pruning, such as layer dropping, removes individual weights or larger structural
units~\cite{frantar2023sparsegpt,sun2024wanda,ashkboos2024slicegpt}. These methods are
attractive because they deliver large efficiency gains while
appearing to preserve model quality. However, this apparent preservation is often validated
through aggregate measures such as perplexity and average downstream
accuracy, which typically remain close to those of the full-precision
model at low compression levels~\cite{,sun2024wanda,men2024shortgpt}.

A growing body of work has begun to probe behaviors that these aggregate
measures do not capture, reporting that \textit{compression can degrade
trustworthiness and fairness} in model-, task-, and metric-dependent
ways. These results, however, remain fragmented. They are often mixed or even
contradictory: some show that pruning and quantization consistently amplify errors on underrepresented subgroups across all compression levels~\cite{hooker2020characterising,tran2022pruning}, while others report fairness improved among moderately quantized models~\cite{kamal2024beyond,hong2024comptrust}. The measure is also metric or method dependent: under two separate constructs of fairness, quantization is reported to reduce \emph{intrinsic} stereotype scores~\cite{goncalves-strubell-2023-understanding} yet to worsen \emph{extrinsic}, task-level fairness~\cite{ramesh-etal-2023-comparative}. Additionally, 
little is known about how compression affects knowledge retention across different levels of fact popularity. That is, whether compressed models preserve well-known facts better than rare ones. Existing work is sparse and has largely been limited to evaluating a single compression method at a single bit width using accuracy alone \cite{chang2025inputs}. Yet accuracy alone is a blind spot that demands to be jointly examined with calibration, bias,  especially across different compression methods and levels

What is missing is a systematic study of \emph{if} and \emph{where} a model
fails when it is compressed. An aggregate bias score can mask harms that fall
unevenly across demographic subgroups~\cite{hua2026uncertaintydrivessocialbias,marcuzzi-etal-2026-quantization}; a drop in accuracy does not reveal
whether a model is now confidently wrong~\cite{zhang-etal-2024-calibrating}; and average
utility says nothing about whether the facts a model forgets are common or
rare~\cite{mallen-etal-2023-trust,sun-etal-2024-head}. Because prior work measures these effects in isolation and with inconsistent metrics, it cannot answer whether they co-occur or point in the
same direction\cite{rath2026quantizationundoesalignmentbias}.

In this work, we study these questions directly under a common protocol across 11 compression methods spanning quantization and pruning with three research questions (RQs):
   \textbf{RQ1 (Accuracy on popularity subgroups).} Does model compression disproportionately degrade knowledge of less popular (``tai'') entities compared with widely known (``head'') entities that are more robustly retained by base model? 
   \textbf{RQ2 (Confidence on lost knowledge).} After compression causes a model to lose
  knowledge it previously had, does the model remain confident in its now
  incorrect answers, and does this confidence vary with knowledge popularity?
  \textbf{RQ3 (Bias).} Does compression change stereotypical
  preferences in both overall and across demographic subgroups?

To answer these questions we evaluate three widely used base models
(\texttt{Llama-3.1-8B-Instruct}, \texttt{Qwen-3-8B}, and \texttt{Gemma-2-9B-it}), using
knowledge retention by popularity benchmarks
(PopQA~\cite{mallen-etal-2023-trust}, 
Head-to-Tail~\cite{sun-etal-2024-head}) and the bias benchmark of WinoBias~\cite{zhao-etal-2018-gender} and BBQ~\cite{parrish-etal-2022-bbq}. We introduce an evaluation protocol that applies the same measures across quantization, pruning, so they can be compared on a common ground. 

Our analysis shows that the behavioral effects of compression are indeed asymmetric and are frequently invisible to aggregate metrics. For RQ1, we find that the effect of compression does not follow a consistent pattern across popularity groups: moderate compression preserves head, middle, and tail
knowledge at comparable rates, while more aggressive compression non-uniformly redistributes retention across popularity groups rather
than simply degrading the tail. For RQ2, we find that models often retain moderate to high confidence in answers that become incorrect after compression, although the pattern varies across models, compression methods and levels. In particular, under severe compression, confidence may either collapse or remain high despite substantial accuracy degradation.
Investigating RQ3, we discover that compression-induced changes in stereotypical preferences vary across demographic subgroups, models, and benchmarks.

\section{Related Works}
\label{sec:related_works}
There are two main bodies of related works, including LLM compression, where we focus on post-training quantization and pruning, and social bias under compression that we discuss below.
Overall, LLM compression aims to reduce the storage, memory, and computational costs of deploying LLMs while retaining their predictive utility.


\textbf{Quantization.} Quantization reduces memory and computational costs by representing weights and activations at lower precision. Post-training quantization methods improve accuracy by optimizing reconstruction-, loss-, or sensitivity-aware calibration objectives~\cite{frantar2023gptq,shao2024omniquant,ding2025cbq,wang2026sliderquant,zhang2024leanquant,kim2025guidedquant}; protecting salient weights or groups from quantization error~\cite{lin2026awq,huang2024slim}; or reshaping weight and activation distributions through scaling, channel reassembly, rotations, or affine transformations~\cite{xiao2023smoothquant,liu2024qllm,ma2024affinequant,liu2025spinquant,hu2025ostquant,sun2024flatquant,liang2025paroquant}. Other methods develop specialized codebook, frame, lattice, vector, binary, or ternary formulations for extreme low-bit compression~\cite{pmlr-v235-egiazarian24a,adepu2024framequant,savkin2025nestquant,xu2026rsavq,chong2026nanoquant,huang2025tequila}.

\textbf{Pruning.}
Pruning deletes the parts of a model that contribute least to its output, leaving a smaller network behind~\cite{zhu2024survey}.
\emph{Structured} pruning removes whole building blocks, such as entire neurons, layers, or hidden dimensions, so the model becomes physically smaller and runs faster ~\cite{ma2023llmpruner,xia2024sheared,li2025tyr,guo2025slimllm}.
Layer dropping, for one, exploits the finding that many of an LLM's middle layers only slightly change the hidden representation, which can be dropped with little loss ~\cite{gromov2025unreasonable,hu2025trimllm}. Prior work typically decides which blocks to drop using an importance score computed on the calibration data and then removes the lowest-scoring blocks in a single pass without retraining~\cite{zhong2025blockpruner,sandri2025twossp, men2024shortgpt}.

\emph{Unstructured} pruning selects weights individually under an overall sparsity budget ~\cite{yang2025wandaplus,zhao2025fistapruner}, whereas \emph{semi-structured} pruning requires each small group to retain a fixed number of weights. For example, the common 2:4 pattern keeps 2 of every 4 consecutive weights
~\cite{fang2024maskllm,liu2025proxsparse}.


In our work, we use the representative compression methods from each of the quantization and pruning categories described above to examine the behavioral effects of compression.

\textbf{Behavioral Effects of LLM Compression.} 
Studies of quantization and pruning show that their effects on social bias are heterogeneous rather than uniformly harmful or beneficial
\cite{ramesh-etal-2023-comparative,hong2024comptrust,
xu-etal-2024-beyond-perplexity}.
Recent work shows that aggregate bias scores can conceal finer-grained changes across demographic groups, including opposing subgroup shifts and newly amplified stereotypes after quantization or pruning
\cite{hua2026uncertaintydrivessocialbias,marcuzzi-etal-2026-quantization,
rath2026quantizationundoesalignmentbias,rath2026weightpruningamplifiesbias}.
These findings suggest that evaluating compression requires examining not only aggregate model behavior, but also how its effects vary across subgroups and compression levels.
For these reasons, we study how different compression methods affect knowledge retention across popularity groups, confidence on lost knowledge, and stereotypical preferences across demographic subgroups.

\section{Evaluation Framework and Research Questions}

We study how model compression affects knowledge retention, model confidence, and social bias. We first introduce shared notation and then present each RQ alongside the corresponding evaluations.

\subsection{Notations}


Let $M_0$ denote the full-precision base model and $M_c$ denote a
compressed version of the same model.
When a definition applies to either model, we write
$M \in \{M_0,M_c\}$.
Let $\mathcal{D}=\{x_i\}_{i=1}^{N}$ denote an evaluation dataset, where
$x_i\in\mathcal{X}$ is a natural-language input. 
Let $\mathcal{I}\subseteq\{1,\ldots,N\}$ denote the indices for which a
reference answer is available, and let $y_i\in\mathcal{Y}$ denote the
task-specific reference answer for $i\in\mathcal{I}$, where $\mathcal{Y}$
is the answer space.
Let $M(x_i)\in\mathcal{R}$ denote the model response to $x_i$, where $\mathcal{R}$ is the response space and
$\mathcal{Y}\subseteq\mathcal{R}$.
Let
$\mathrm{Match}(M(x_i),y_i)\in\{0,1\},
\label{eq:match}$
denote the task-specific evaluation function, where
$\mathrm{Match}(M(x_i),y_i)=1$ if response $M(x_i)$ matches reference answer $y_i$
under the task-specific evaluation rule, and $0$ otherwise.
The rule may use exact, substring, or multiple-choice matching depending on the task.

Given a model $M\in\{M_0,M_c\}$, we evaluate its perplexity and overall accuracy.
For a tokenized corpus of $T$ tokens, we compute the perplexity as
$\mathrm{PPL}(M)
=
\exp\left(
-\frac{1}{T}
\sum_{t=1}^{T}
\log p_M(w_t\mid w_{<t})
\right),$
where $p_M(w_t\mid w_{<t})$ denotes the probability assigned by model $M$ to token $w_t$ given its preceding tokens.
For an input $x_i$, let
$\hat{y}_i=(\hat{y}_{i,1},\ldots,\hat{y}_{i,T_i})$
denote the answer generated by model $M$, where $T_i$ is the length of
generated tokens. 
The overall accuracy of model $M$ is
$\mathrm{Acc}
=
\frac{1}{|\mathcal{I}|}
\sum_{i\in\mathcal{I}}
\mathrm{Match}\bigl(M(x_i),y_i\bigr).$ When distinguishing between models, we use
\(\mathrm{Acc}_0\) and \(\mathrm{Acc}_c\) to denote the accuracies of
\(M_0\) and \(M_c\), respectively.

To evaluate model behavior across data subsets, we introduce a unified notation. Let $\mathcal{G}$ denote the set of groups under consideration, which may represent knowledge popularity (Sections~\ref{subsec:rq1_framework} and~\ref{subsec:rq2_framework}) or demographic categories (Section~\ref{subsec:rq3_framework}). For each example $i$, let $G(i)\in\mathcal{G}$ denote its group assignment. The specific subset of examples belonging to group $g$ is defined as $\mathcal{I}_g=\{i\in\mathcal{I}:G(i)=g\}$, with its total size denoted by $N_g=\vert{}\mathcal{I}_g\vert{}$.


Next, we focus on methodology of each of our three questions separately.

\subsection{RQ1: Knowledge Retention by Popularity}
\label{subsec:rq1_framework}

Factual knowledge varies in popularity, and language models generally struggle to recall less popular facts \cite{mallen-etal-2023-trust}. Since aggregate accuracy can mask non-uniform compression effects \cite{chang-etal-2025-inputs}, it remains unclear whether compressed models preserve knowledge equally across the popularity spectrum. We therefore ask:

\textbf{RQ1: Does model compression disproportionately reduce the retention
of tail knowledge relative to head knowledge?}
To answer this, we evaluate knowledge retention across three popularity groups, $\mathcal{G}=\{\mathrm{head},\mathrm{middle},\mathrm{tail}\}$, corresponding to high-, medium-, and low-popularity facts, respectively.

\textbf{Measures.} For RQ1, we consider base model and group-level accuracy, retention rate, and relative retention shift, all of which we define next.

For each group $g\in\mathcal{G}$, the group-level accuracy of model $M$ is
\begin{equation}
\mathrm{Acc}_{g}
=
\frac{1}{N_g}
\sum_{i\in\mathcal{I}_g}
\mathrm{Match}\bigl(M(x_i),y_i\bigr).
\label{eq:group_accuracy}
\end{equation}

Because base model performance varies across popularity groups, comparing absolute accuracy can be misleading. To standardize the degradation, we compute the accuracy retention rate \cite{laborde2025}, $r=\mathrm{Acc}_{c}/\mathrm{Acc}_{0}$, alongside the group-level accuracy retention rate, $r_g=\mathrm{Acc}_{c,g}/\mathrm{Acc}_{0,g}$.


While $r_g$ measures isolated group degradation, it does not reveal whether a group is harmed disproportionately compared to the model's average. To quantify this disparity, we introduce the relative retention shift (in percentage points):
\begin{equation}
\mathrm{RS}_g
=
\left(
r_g-r
\right)
\times 100.
\label{eq:relative_retention_shift}
\end{equation}
A negative (positive) $\mathrm{RS}_g$ indicates that group $g$ retains less (more) of its base-model accuracy than the dataset overall, meaning it is more (less) affected by compression.

\textbf{Mathematical Formulation.}
Formally, RQ1 tests whether compression disproportionately damages tail knowledge compared to head knowledge, which would result in
$r_{\mathrm{tail}}
<
r_{\mathrm{head}}$ or $r_{\mathrm{head}}
<
r_{\mathrm{tail}}.$
Using our normalized metric, this  is equivalent to comparing shifts between the tail ($\mathrm{RS}_{\mathrm{tail}}$) and head knowledge ($\mathrm{RS}_{\mathrm{head}}$).
We
 examine $\mathrm{RS}_g$ across all three popularity groups 
$g\in\{\mathrm{head},\mathrm{middle},\mathrm{tail}\}$ and
 report the results in Section \ref{subsec:RQ1_results}. We describe methodology for RQ2 next.

\subsection{RQ2: Confidence after Knowledge Loss}
\label{subsec:rq2_framework}

Model compression may lead to knowledge loss, turning answers that are correct in the base model into incorrect ones. 
However, it's unclear whether the compressed model's confidence decreases accordingly or remains high, 
given that LLMs are often overconfident in their responses \cite{zhang-etal-2024-calibrating}.
We therefore ask:

 \textbf{RQ2: How confident are compressed models when answering questions involving lost knowledge?} 
We define the lost-knowledge set as examples answered correctly by the base model but incorrectly by the compressed model: $\mathcal{I}_{\mathrm{loss}}=\{i\in\mathcal{I}: \mathrm{Match}(M_{0}(x_{i}),y_{i})=1, \mathrm{Match}(M_{c}(x_{i}),y_{i})= 0\}$. To determine if this overconfidence varies with fact popularity, we evaluate across the same popularity groups used in RQ1. For each popularity group $g$, the corresponding lost-knowledge subset is $\mathcal{I}_{\mathrm{loss},g}= \{i\in\mathcal{I}_{\mathrm{loss}}:G(i)=g\}$. 

\textbf{Measures.}
We consider knowledge-loss rate, confidence on lost knowledge, and
expected calibration error (ECE), which we describe next.

To account for varying group sizes, we first measure the frequency of knowledge loss via the knowledge-loss rate:
\begin{equation}\mathrm{LR}g
=
|\mathcal{I}_{\mathrm{loss},g}|/N_g.
\label{eq:knowledge_loss_rate}\end{equation}

However, $\mathrm{LR}_g$ alone does not reveal how confident the compressed model is in the resulting incorrect answers. We evaluate confidence on newly incorrect answers using the length-normalized sequence probability:  
\cite{vashurin-etal-2025-benchmarking}:\begin{equation}\mathrm{Conf}(x_i)=p_M(\hat{y}_i\mid x_i)^{\frac{1}{T_i}},
\label{eq:generated_answer_confidence}
\end{equation}
where $p_M(\hat{y}_i\mid x_i) = \prod_{t=1}^{T_i} p_M(\hat{y}_{i,t}\mid x_i,\hat{y}_{i,<t})$.

Finally, while confidence captures the model's certainty on individual generated answers, it does not measure overall reliability. As a complementary metric, we compute the expected calibration error (ECE) to assess how well confidence aligns with empirical correctness across the group:
\begin{equation}
\mathrm{ECE}{g}=\sum_{k=1}^{K}\frac{|B_{k,g}|}{N_g}\left|\mathrm{Acc}(B_{k,g})-\mathrm{Conf}(B_{k,g})\right|,\label{eq:group_ece}\end{equation}where $\mathrm{Acc}(B_{k,g})$ and $\mathrm{Conf}(B_{k,g})$ denote the empirical accuracy and mean confidence of the examples in bin $k$, respectively.

\textbf{Mathematical Formulation.}
RQ2 examines the compressed model's confidence $\mathrm{Conf}(x_i)$ on lost knowledge ($i\in\mathcal{I}_{\mathrm{loss},g}$). High confidence here indicates certainty in newly incorrect answers. We compare this across popularity groups $g\in\mathcal{G}$, while also reporting $\mathrm{LR}_g$ (loss frequency) and $\mathrm{ECE}_g$ (calibration). Results are detailed in Section~\ref{subsec:RQ2_results}.

\subsection{RQ3: Compression-Induced Bias Change}
\label{subsec:rq3_framework}


Compression may alter stereotypical preferences even when aggregate utility and bias measures remain stable, particularly when opposing subgroup-level shifts cancel out.
We therefore ask:

\textbf{RQ3: How does model compression affect overall and subgroup-level
stereotypical preferences?}
Unlike the previous two research questions, where groups represent knowledge popularity levels, in RQ3 each $g\in\mathcal{G}$ represents a demographic group, defined by features such as age, gender, religion, or race. 
This grouping enables us to examine compression-induced bias both in aggregate across all examples and separately within each demographic group. 

\textbf{Measures.} 
For RQ3, we consider item-level stereotypical preference, overall bias change,
and subgroup-level bias change.

For each example $i$, let $s_i^{\mathrm{st}}(M)$ and
$s_i^{\mathrm{ref}}(M)$ denote the scores assigned by model $M$ to the
stereotypical and reference answers, respectively. To determine whether a model favors the stereotypical answer on each example,
we define
$z_i(M)
=
\mathbf{1}
\left[
s_i^{\mathrm{st}}(M)
>
s_i^{\mathrm{ref}}(M)
\right],
\label{eq:item_bias_indicator}$
where $z_i(M)=1$ indicates a preference for the stereotypical answer.
To summarize the prevalence of these item-level preferences across the entire dataset, we define the overall bias score as
$B(M)
=
\frac{1}{N}
\sum_{i=1}^{N}
z_i(M).
\label{eq:overall_bias_score}$

Finally, to isolate the specific impact of compression, we calculate the shift in bias relative to the original base model:
\begin{equation}
\Delta B(M_c)
=
B(M_c)-B(M_0).
\label{eq:overall_delta_bias}
\end{equation}
A positive $\Delta B(M_c)$ indicates increased stereotypical preference after
compression, whereas a negative value indicates decreased preference.
To quantify uncertainty, we report 95\% confidence intervals for $\Delta B(M_c)$ and $\Delta B_g(M_c)$.

The overall change may conceal heterogeneous effects across demographic groups.
We also compute the group-level bias score
$B_g(M)
=
({1}/{N_g})
\sum_{i\in I_g}
z_i(M),$
\label{eq:subgroup_bias_score}
and its compression-induced change
$\Delta B_g(M_c)
=
B_g(M_c)-B_g(M_0).$
\label{eq:subgroup_delta_bias}

\textbf{Mathematical Formulation.}
RQ3 examines the overall compression-induced bias
change $\Delta B(M_c)$ as well as the change within each demographic group,
$\Delta B_g(M_c)$.
Differences in $\Delta B_g(M_c)$ across $g\in\mathcal{G}$ indicate that
compression affects demographic groups heterogeneously, even when the overall
change $\Delta B(M_c)$ is small.
We report the results for RQ3 in Section~\ref{subsec:RQ3_results}.

\section{Experiments and Results}
\label{sec:results}

We discuss our results for each RQ separately after providing more details on our experimental pipeline, methods and datasets that we use.

\subsection{Experiment setup}

\textbf{Compression Methods.}
We consider four representative post-training weight-only quantization methods widely used for efficient LLM deployment:
GPTQ~\cite{frantar2023gptq}, AWQ~\cite{lin2026awq}, OmniQuant~\cite{shao2024omniquant}, and AQLM~\cite{pmlr-v235-egiazarian24a}. 
For pruning, we evaluate unstructured magnitude pruning, WANDA~\cite{sun2024wanda}, and SparseGPT~\cite{frantar2023sparsegpt} at 30\%, 50\%, and 70\% sparsity; semi-structured WANDA and SparseGPT with 4:8 and 2:4 sparsity patterns; and structured methods including ShortGPT~\cite{men2024shortgpt}, importance-based layer dropping~\cite{kim2024shortened,gromov2025unreasonable,song2024sleb,yang2024laco}.
More details are in Appendix \ref{app:setup_additional_details}.

\begin{figure*}[t!]
    \centering
    \includegraphics[width=0.9\textwidth]{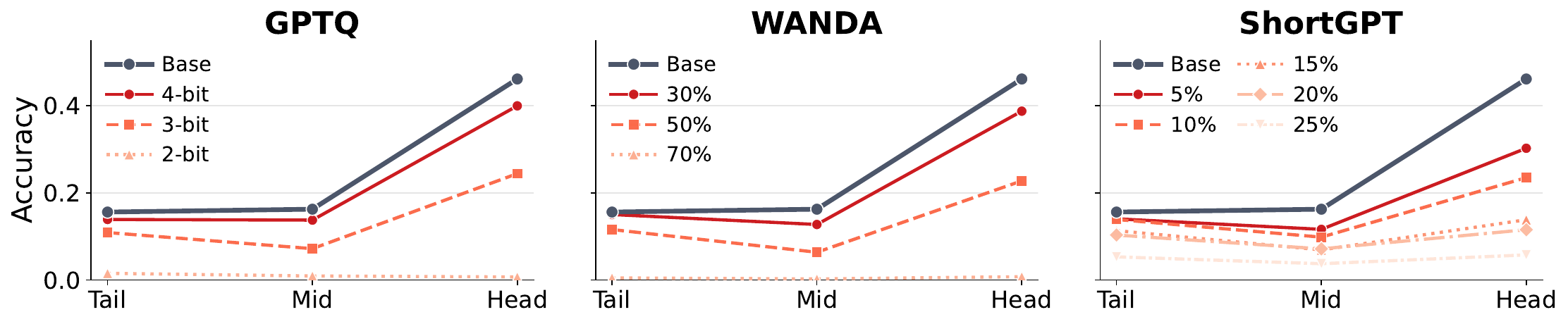}
   \caption{
Accuracy across different popularity groups on PopQA under different compression settings. Results for all methods and models are provided in
Figures~\ref{fig:rq1_popqa_bucket_accuracy_other_methods_llama},
\ref{fig:rq1_popqa_bucket_accuracy_other_methods_qwen}, and
\ref{fig:rq1_popqa_bucket_accuracy_other_methods_gemma}
in Appendix~\ref{app:rq1_additional_results}.
}
    \label{fig:rq1_popqa_bucket_accuracy}
\end{figure*}

\begin{figure*}[t!]
    \centering
    \includegraphics[width=0.9\textwidth]{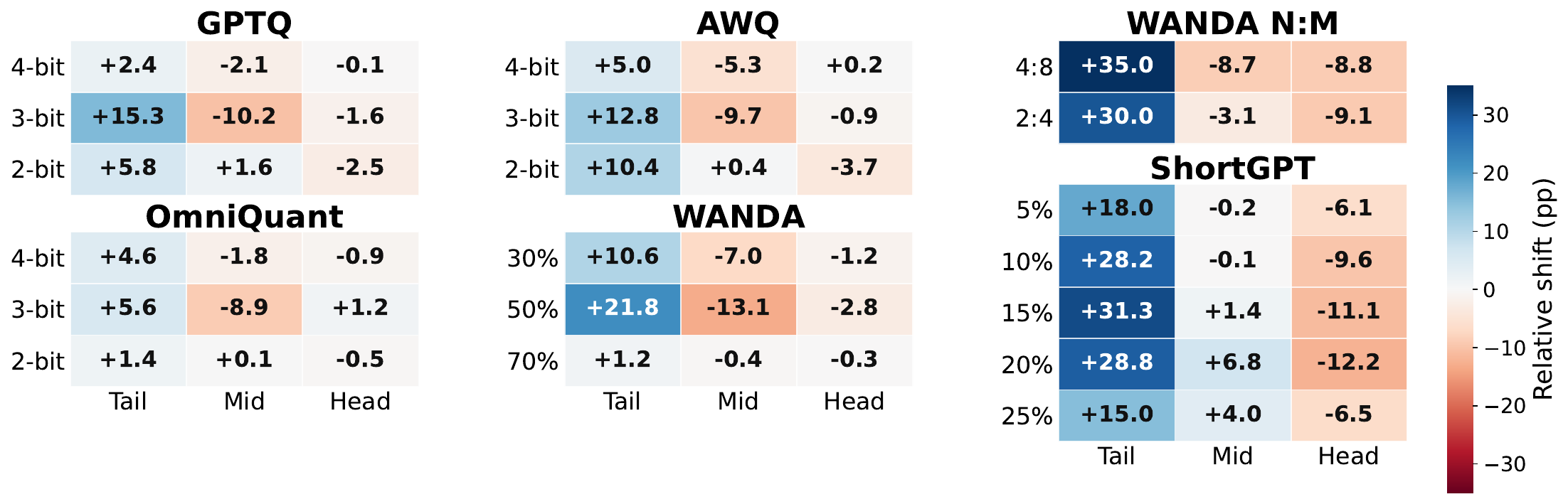}
    \caption{
Relative retention shift across PopQA popularity groups under different compression settings, reported in percentage points.
Negative and positive $\mathrm{RS}_g$ indicate lower and higher relative retention than the dataset overall, respectively.
Results for all methods and models are provided in
Figures~\ref{fig:rq1_popqa_relative_shift_other_methods_llama},
\ref{fig:rq1_popqa_relative_shift_other_methods_qwen}, and
\ref{fig:rq1_popqa_relative_shift_other_methods_gemma}
in Appendix~\ref{app:rq1_additional_results}.}
    \label{fig:rq1_popqa_relative_shift}
\end{figure*}


\textbf{Datasets.} 
We use PopQA~\cite{mallen-etal-2023-trust} and Head-to-Tail~\cite{sun-etal-2024-head} for RQ1 and RQ2 because they organize factual knowledge by popularity into head, middle, and tail groups.
For RQ3, we use WinoBias~\cite{zhao-etal-2018-gender} and BBQ~\cite{parrish-etal-2022-bbq} which provide the demographic and stereotype annotations needed for overall and subgroup-level bias analysis.

\textbf{Models.} We evaluate results on three open-weight models: 
Llama-3.1-8B-Instruct~\cite{grattafiori2024llama},
Qwen-3-8B~\cite{yang2025qwen3}, and
Gemma-2-9B-it~\cite{gemmateam2024gemma2} 
to span distinct
pretraining corpora and architectural choices. 
We present results for \texttt{Llama-3.1-8B-Instruct} in Section~\ref{sec:results}. Results for the other two models are in Appendix~\ref{app:result_additional_details}.


\textbf{Compression performance.}
We evaluate overall performance using accuracy for PopQA  and perplexity for WikiText-2  (Figure~\ref{fig:rq1_popqa_overall_accuracy} and  Table~\ref{tab:compression_ppl} in Appendix \ref{app:setup_additional_details} and \ref{app:rq1_additional_results}). Throughout our analysis, we define collapsed settings as those with near-zero accuracy and sharply increased perplexity, and non-collapsed settings as those retaining meaningful performance. For instance, 4-bit quantization and 30\% WANDA remain non-collapsed, whereas 2-bit GPTQ, AWQ, and OmniQuant, 70\% unstructured pruning (WANDA/SparseGPT), and 25\% ShortGPT collapse. Because SliceGPT suffers severe collapse across all tested levels for \texttt{Qwen-3-8B} and \texttt{Gemma-2-9B-it} (highlighted orange in Table~\ref{tab:compression_ppl}), we exclude it from subsequent evaluations.

\subsection{RQ1 results}
\label{subsec:RQ1_results}

Using group-level accuracy and relative retention shift (Equations~\ref{eq:group_accuracy} and \ref{eq:relative_retention_shift}), we reach two conclusions: In \emph{absolute} terms, the base model's popularity ordering (head $>$ tail) remains under mild compression. However, in \emph{relative} terms, this ordering reverses.

\textbf{Accuracy by Popularity.} The base model exhibits a large $\sim 30\%$ popularity accuracy gap between tail/middle and head knowledge (Figure~\ref{fig:rq1_popqa_bucket_accuracy}). Mild compression (4-bit quantization, 30\% WANDA, 5\% ShortGPT) largely preserves this ordering, while more aggressive compression degrades all three groups. Head knowledge thus stays the most accurate group throughout the non-collapsed regime. The \emph{magnitude} of degradation, however, is model- and method-dependent: at 25\% layer dropping, for example, Llama's overall
accuracy decreases from $26.0\%$ to $2.3\%$ while Qwen and Gemma retain $14.3\%$ and $13.5\%$, and AQLM, designed for extreme low-bit quantization, keeps $12.3\%$ accuracy at 2-bit versus $\leq 1.9\%$ for GPTQ, AWQ, and OmniQuant (Appendix~\ref{app:rq1_additional_results}). For near-collapse settings, under 2-bit quantization and the strongest WANDA, accuracy falls to near zero across all groups.

\textbf{Relative Retention Shift by Popularity.} Relative retention reverses the accuracy ordering. After normalizing each group's accuracy to its base-model value, tail knowledge is retained better than the model's overall accuracy, while head knowledge is worse. This asymmetry is consistent across almost all model and compression combinations on PopQA. Further, its magnitude grows from quantization to pruning (Figure~\ref{fig:rq1_popqa_relative_shift}, and Figures \ref{fig:rq1_popqa_relative_shift_other_methods_llama}, \ref{fig:rq1_popqa_relative_shift_other_methods_qwen}, \ref{fig:rq1_popqa_relative_shift_other_methods_gemma} in Appendix~\ref{app:rq1_additional_results}). Quantization produces smaller shifts, with 4-bit GPTQ, AWQ, and OmniQuant within 5.3 percentage points of overall retention, while N:M pruning is the most extreme with the tail retention 22.7 to 39.5 percentage points above overall and head retention 3.6 to 14.4 percentage points below it. At the strongest compression settings, such as 70\% WANDA and 25\% ShortGPT, the head/tail gap largely disappears because all groups lose almost all of their accuracy. One trend does not hold monotonically: under ShortGPT the tail and head gap is widest at a moderate 10\% to 15\% pruning ratio rather than at 25\%, reaching up to 42.4, 44.6, and 52.7 percentage points for Llama, Qwen, and Gemma before narrowing at the strongest setting (see Appendix \ref{app:rq1_additional_results}).

\begin{figure*}[ht]
    \centering
    \includegraphics[width=0.9\textwidth]
    {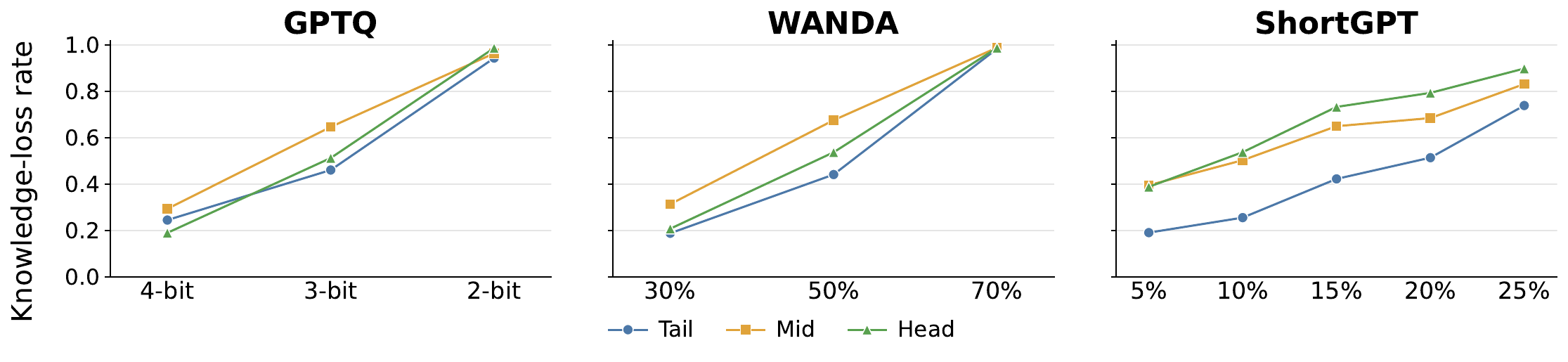}
    \caption{Knowledge-loss rate across popularity groups on PopQA under different compression settings. Higher values indicate that a
    larger proportion of the knowledge correctly answered by the base model is incorrect after compression. Results for all methods and models are provided in
Figures~\ref{fig:rq2_popqa_knowledge_loss_rate_other_methods_llama},
\ref{fig:rq2_popqa_knowledge_loss_rate_all_methods_qwen}, and
\ref{fig:rq2_popqa_knowledge_loss_rate_all_methods_gemma}
in Appendix~\ref{app:rq2_additional_results}.}
    \label{fig:rq2_popqa_knowledge_loss_rate}
\end{figure*}

\begin{figure*}[ht]
    \centering
    \includegraphics[width=0.9\textwidth]{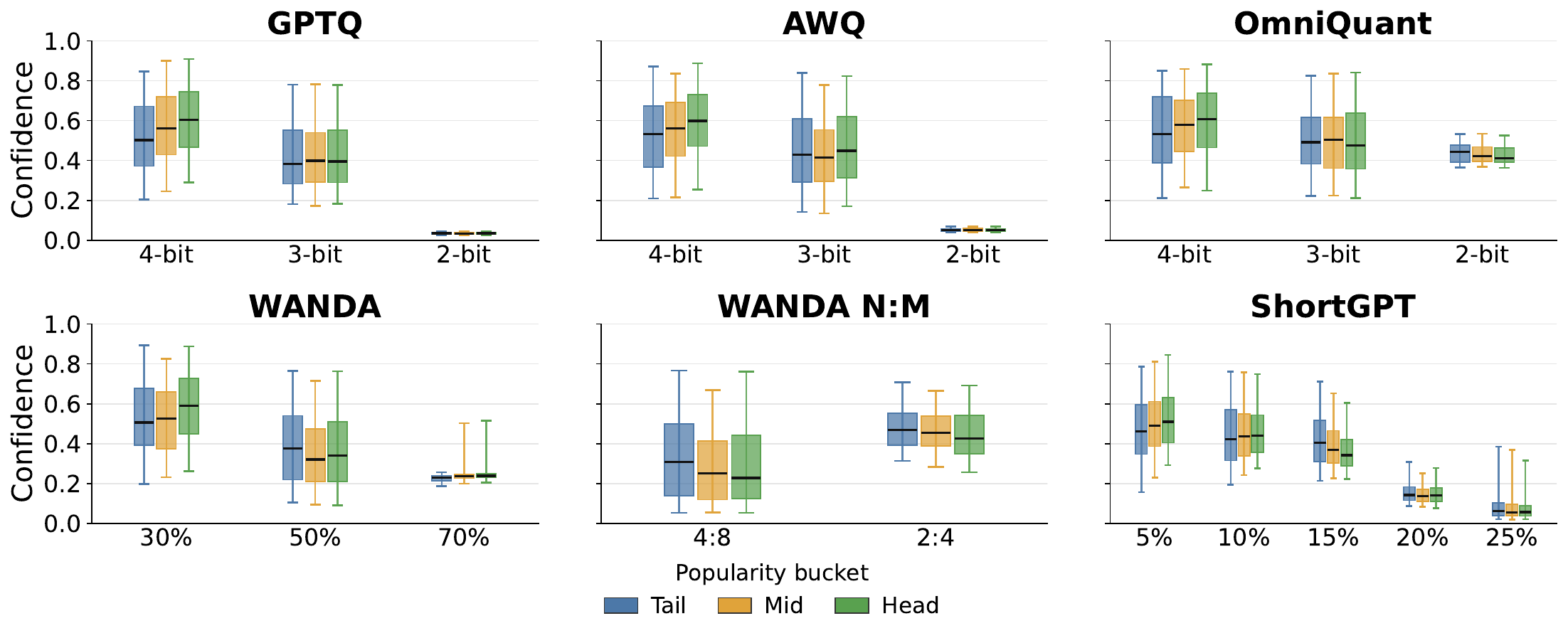}
    \caption{
Confidence on lost knowledge across popularity groups on PopQA under different compression settings. Each dot represents the median confidence; the thick vertical bar shows the 25th--75th percentile range, and the thin vertical bar shows the 5th--95th percentile range. Colors distinguish tail, middle, and head knowledge.
Results for all methods and models are provided in
Figures~\ref{fig:rq2_popqa_lost_confidence_other_methods_llama},
\ref{fig:rq2_popqa_lost_confidence_other_methods_qwen}, and
\ref{fig:rq2_popqa_lost_confidence_other_methods_gemma}
in Appendix~\ref{app:rq2_additional_results}.
}
    \label{fig:rq2_popqa_lost_confidence}
\end{figure*}

Taken together, compression does not preferentially erase tail knowledge. Although head knowledge remains the most accurate group, it is proportionally the least retained, an effect invisible to aggregate accuracy or perplexity.

\subsection{RQ2 Results}
\label{subsec:RQ2_results}

Because raw sequence probabilities are not necessarily calibrated to empirical correctness, we calibrate confidence scores for RQ2 using isotonic regression. For each dataset, we use a popularity-stratified 20/80 calibration--test split and fit the regression on the calibration split, using raw $\mathrm{Conf}(x_i)$ to predict the substring-based correctness indicator $\mathrm{Match}\bigl(M(x_i),y_i\bigr)$. All subsequent confidence and calibration metrics are computed on the held-out test split.

Using the knowledge-loss rate and confidence on lost knowledge (Equations~\ref{eq:knowledge_loss_rate} and~\ref{eq:generated_answer_confidence}, see Appendix~\ref{app:rq2_additional_results} for calibration error) we observe that compression causes substantial knowledge loss even at settings where compressed accuracy is stable and that models often remain confident in these incorrect answers.

\textbf{Knowledge-Loss Rate.} Knowledge loss rises sharply with compression strength. Even at mild settings, such as 4-bit GPTQ, 30\% WANDA, and 5\% ShortGPT, where standard metrics suggest only minimal degradation, 22\% to 35\% of previously correct PopQA answers become incorrect. This proportion rises to 85 to 99\% under the strongest compression settings (Figure~\ref{fig:rq2_popqa_knowledge_loss_rate}). Consistent with RQ1, it is unevenly distributed, and tail knowledge is the least affected, for example 19.1\% loss versus 38 to 40\% for middle and head knowledge under 5\% ShortGPT. 
However, this pattern does not fully generalize to Head-to-Tail,
where mild and moderate compression generally results in higher loss rates and a less consistent ordering across popularity groups
(Appendix~\ref{app:rq2_additional_results}).

\textbf{Confidence on Lost Knowledge.}
On the knowledge it loses, a compressed model often remains confident. Under mild to moderate compression the median confidence on now-incorrect answers stays near 0.4 to 0.6, and the model is more confident on lost head knowledge than on lost tail knowledge, by up to 1.20$\times$ under 4-bit GPTQ (Figure~\ref{fig:rq2_popqa_lost_confidence}). This confidence does not fall steadily as compression increases. 
It drops to near zero under 2-bit GPTQ and AWQ but stays around 0.4 under 2-bit OmniQuant, and it is higher under 2:4 than under 4:8 N:M pruning, so confidence declines sharply only in some collapsed settings and only after accuracy has already deteriorated. The head-over-tail pattern is also not universal across models, weakening or reversing in several Qwen settings (Appendix~\ref{app:rq2_additional_results}).

Standard calibration metrics might be misleading here. ECE tends to improve under compression, but its largest reductions align with collapsed settings where accuracy and confidence are both near zero, so a lower ECE does not necessarily indicate better reliability (Appendix~\ref{app:rq2_additional_results}).

\begin{figure*}[ht]
    \centering
    \includegraphics[width=0.9\textwidth]
    {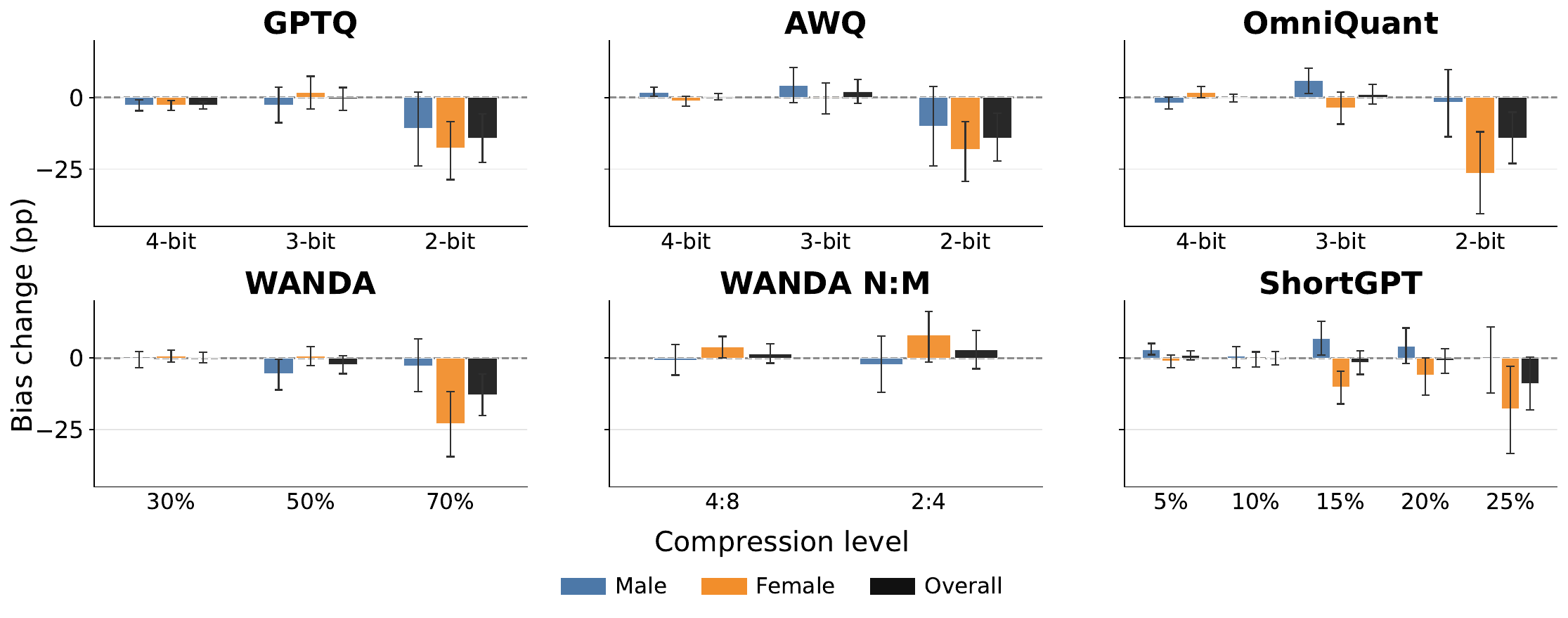}
    \caption{
Bias changes on WinoBias under different compression settings, reported in
percentage points. Blue, orange, and black bars denote changes for male,
female, and all examples, respectively, and error bars denote 95\% confidence
intervals. Positive values indicate increased stereotypical preference,
negative values indicate decreased preference, and the dashed line marks no
change from the base model. Results for all methods and models are provided in
Figures~\ref{fig:rq3_winobias_llama_subgroup_bias_change_other_methods},
\ref{fig:rq3_winobias_qwen_subgroup_bias_change}, and
\ref{fig:rq3_winobias_gemma_subgroup_bias_change} in
Appendix~\ref{app:rq3_additional_results}.
}
\label{fig:rq3_winobias_subgroup_bias_change}
\end{figure*}

\subsection{RQ3 Results} 
\label{subsec:RQ3_results} 
Using the bias change defined in Equation~\ref{eq:overall_delta_bias}, we reach two conclusions. First, a small overall change in stereotypical preference does not imply stable behavior, since it can hide large and opposing shifts within subgroups. Second, these shifts concentrate on a few specific subgroups, and which subgroup is affected depends on the model and the dataset. We focus here on non-collapsed configurations, because collapsed models produce unreliable and often extreme bias estimates.

\textbf{Overall and Subgroup Bias Change.} Figure~\ref{fig:rq3_winobias_subgroup_bias_change} shows that a small overall bias change can mask larger, opposing changes within gender subgroups. Under 3-bit OmniQuant, stereotypical preference increases by about 6 percentage points for male group but decreases by about 4 points for female group, with the overall change being near zero. A similar ``cancellation'' appears under 15\% ShortGPT, where the male and female shifts are about $+7$ and $-9$ percentage points. Such patterns occur across several quantization and pruning settings, so a small aggregate change does not indicate stable bias across subgroups. The effect is also benchmark dependent. On WinoBias the subgroup changes are large and sometimes opposing, whereas on BBQ the overall changes are small and mostly negative, ranging from about a 5-point decrease to a 4-point increase (see Appendix \ref{app:rq3_additional_results}). 

\textbf{Worst-Subgroup Effects.}
Tables~\ref{tab:rq3_winobias_largest_subgroup_changes} and
\ref{tab:rq3_bbq_largest_subgroup_changes} show that large subgroup shifts can coexist with much smaller aggregate changes. On WinoBias, Llama under SparseGPT 2:4 yields \(\Delta B_g=-53.1\) pp for \textit{secretary} (95\% CI \([-77.7,-14.1]\)), while the overall change is only \(\Delta B=-2.2\) pp (95\% CI \([-9.1,+4.2]\)).
This difference indicates that an aggregate score can conceal a large change concentrated within a single occupation. 
A similar pattern appears on BBQ: Qwen under 10\% ShortGPT yields
\(\Delta B_g=+37.5\) pp for \textit{Down's syndrome}
(95\% CI \([-17.3,+74.4]\)), compared with an overall change of only \(-1.2\) pp (95\% interval \([-2.1,-0.2]\)).
Although the subgroup interval is wide, the point estimate illustrates how a large identity-specific shift can be diluted when averaged across the full benchmark. 
Thus, aggregate scores can obscure both the magnitude and location of compression-induced bias shift (Appendix~\ref{app:rq3_additional_results}).
Taken together, overall bias score is not sufficient to certify that a compressed model is fair.


\section{Discussion and Conclusions}

Model compression is rarely a uniform scaling down of model quality. Across 11 compression methods, we find that standard aggregate metrics such as average accuracy, perplexity, and overall bias scores create blind spots for practitioners, because they mask asymmetric behavioral changes. Compression degrades common (head) knowledge proportionally more than rare (tail) knowledge, even though head knowledge remains the most accurate group. It leaves models moderately confident in answers that have become incorrect. It also produces large, subgroup-specific shifts in stereotypical preference that a small aggregate change conceals. These effects are most concerning in the mild-to-moderate regimes where aggregate metrics  appear reliable. Certifying a compressed model as intact from average utility alone is therefore insufficient and reliable deployment requires granular, subgroup-level evaluation.

\section*{Limitations}

Our study has following limitations. First, our experiments cover three 8--9B instruction-tuned models and selected post-training quantization and pruning methods. The findings may not generalize to substantially larger models, other architectures, or compression paradigms such as distillation.


Second, unlike the confidence analysis in RQ2, the stereotypical-preference analysis in RQ3 does not apply probability calibration. WinoBias and BBQ do not provide a natural correctness-based calibration target for the stereotypical-preference scores, so the calibration procedure used in RQ2 cannot be directly transferred to these benchmarks.

\bibliography{custom,anthology-1,anthology-2}
\clearpage

\appendix

\section{Appendix}

\subsection{Experiment Setup: Additional Details}
\label{app:setup_additional_details}

\subsubsection{LLM Compression Configurations}

\begin{table*}[!t]
\centering
\normalsize
\setlength{\tabcolsep}{6pt}
\renewcommand{\arraystretch}{1.05}

\begin{tabularx}{\textwidth}{
@{}
>{\raggedright\arraybackslash}p{0.34\textwidth}
>{\centering\arraybackslash}X
>{\centering\arraybackslash}X
>{\centering\arraybackslash}X
>{\centering\arraybackslash}X
@{}
}
\hline
\textbf{Parameter}
& \textbf{GPTQ}
& \textbf{AWQ}
& \textbf{OmniQuant}
& \textbf{AQLM} \\
\hline

Bit width
& 2, 3, 4
& 2, 3, 4
& 2, 3, 4
& 2, 3, 4 \\

Calibration dataset
& C4
& C4
& C4
& C4 \\

Calibration samples
& 128
& 128
& 128
& 128 \\

Calibration seed
& 42
& 42
& 42
& 42 \\

Calibration sequence length
& 512
& 512
& 512
& 512 \\

Group size
& 128
& 128
& 128
& -- \\

Calibration split
& validation
& validation
& --
& -- \\

Calibration source records
& 4096
& --
& --
& -- \\

Calibration batch size
& 1
& 1
& --
& -- \\

Symmetric quantization
& True
& False
& --
& -- \\

Activation ordering
& True
& --
& --
& -- \\

Sequential quantization
& True
& --
& --
& -- \\

Target modules
& --
& Linear
& --
& -- \\

Ignored modules
& --
& lm\_head
& --
& -- \\

Activation bit width
& --
& --
& 16
& -- \\

Optimization epochs
& --
& --
& 40 / 20 / 20
& -- \\

Learnable weight clipping
& --
& --
& True
& -- \\

Learnable equivalent transformation
& --
& --
& False
& -- \\

Input group size
& --
& --
& --
& 8 \\

Output group size
& --
& --
& --
& 1 \\

Relative MSE tolerance
& --
& --
& --
& 0.01 \\

Maximum fine-tuning epochs
& --
& --
& --
& 10 \\

Activation offloading
& --
& --
& --
& True \\

Resume enabled
& --
& --
& --
& True \\

\hline
\end{tabularx}

\caption{Quantization configurations.}
\label{tab:quantization_configuration}
\end{table*}

\paragraph{Quantization.}

Table~\ref{tab:quantization_configuration} summarizes the quantization configurations used for all our models. 
We evaluate all four
quantization methods at 2-, 3-, and 4-bit precision and use C4 for calibration.
The remaining parameters specify the
method-specific quantization settings.

The bit width specifies the precision of the quantized weights. We
evaluate GPTQ, AWQ, OmniQuant, and AQLM at 2-, 3-, and 4-bit precision. For all
methods, the calibration dataset, number of calibration samples, seed, and
sequence length define the data used during quantization. We use C4 with 128
calibration samples, a seed of 42, and a sequence length of 512.

GPTQ uses a group size of
128 and draws calibration samples from 4,096 records in the C4 validation split
with a batch size of 1. Symmetric quantization, activation ordering, and
sequential quantization are enabled. AWQ also uses a group size of 128, the C4
validation split, and a batch size of 1, but applies asymmetric quantization. It
targets linear modules while excluding lm\_head. OmniQuant retains
16-bit activations and performs 40, 20, and 20 optimization epochs for 2-, 3-,
and 4-bit quantization, respectively. Learnable weight clipping is enabled,
whereas learnable equivalent transformation is disabled. AQLM uses input and
output group sizes of 8 and 1, respectively, a relative MSE tolerance of 0.01,
and a maximum of 10 fine-tuning epochs. Activation offloading and resumption
from saved progress are enabled.

\paragraph{Pruning.}

Table~\ref{tab:pruning_configuration} summarizes the pruning configurations
used for all models. Magnitude, WANDA, and SparseGPT prune individual weights
at sparsity levels of 30\%, 50\%, and 70\%, whereas ShortGPT removes
transformer blocks at ratios from 5\% to 25\%. All calibration-based methods use 32 sequences of
length 512 from the C4 validation split. No method applies recovery
fine-tuning.

Magnitude pruning requires
no calibration data or activation statistics. WANDA and SparseGPT additionally
support semi-structured 4:8 and 2:4 sparsity patterns. Together with magnitude
pruning, they target linear projections in the attention and MLP modules while
excluding lm\_head. SparseGPT further uses Hessian-based weight
reconstruction. ShortGPT ranks decoder blocks by the mean per-token cosine
similarity between their inputs and outputs while protecting the first and last
blocks from removal. 

For layer dropping, in each decoder
layer \(\ell\), we estimate an importance score using both task activations and
weight magnitudes to measure how much each layer changes the hidden representation:
\[
a_{\ell}(x)
=
\frac{
\mathrm{RMS}\!\left(h_{\ell+1}(x)-h_{\ell}(x)\right)
}{
\mathrm{RMS}\!\left(h_{\ell}(x)\right)+\epsilon
}.
\]
We average this quantity across calibration prompts to obtain activation
importance \(A_{\ell}\). We also compute a weight-magnitude importance score
\(W_{\ell}\), defined as the average Root Mean Square magnitude of linear weights inside the
decoder block. After min-max normalizing both quantities across layers, the
combined layer importance is
\[
I_{\ell}=0.7\,\widetilde{A}_{\ell}+0.3\,\widetilde{W}_{\ell}.
\]
We protect the first and last decoder layers. Among the remaining interior layers, we add a
position prior
\[
P_{\ell}=1-\left|2\ell/(L-1)-1\right|,
\]
which is largest near the middle of the network. The final drop score is
\[
D_{\ell}=0.65(1-I_{\ell})+0.35P_{\ell}.
\]
Layers with the largest \(D_{\ell}\) are dropped. Thus, the policy favors
layers that are both low-importance and safely located in the interior of the
model. Block importance is estimated once per base model from C4 calibration data.

\begin{table*}[!t]
\centering
\setlength{\tabcolsep}{3pt}
\renewcommand{\arraystretch}{1.08}

\begin{tabularx}{\textwidth}{
@{}
>{\raggedright\arraybackslash}p{0.245\textwidth}
>{\centering\arraybackslash}X
>{\centering\arraybackslash}X
>{\centering\arraybackslash}X
>{\centering\arraybackslash}X
>{\centering\arraybackslash}X
@{}
}
\hline
\textbf{Parameter}
& \textbf{Magnitude}
& \textbf{WANDA}
& \textbf{SparseGPT}
& \textbf{ShortGPT}
& \textbf{Layer Dropping} \\
\hline

Compression granularity
& weights
& weights
& weights
& decoder blocks
& decoder blocks \\

Compression levels
& 30/50/70\%
& 30/50/70\%
& 30/50/70\%
& 5/10/15/20/25\%
& 5/10/15/20/25\% \\

Semi-structured patterns
& --
& 4:8, 2:4
& 4:8, 2:4
& --
& -- \\

Calibration dataset
& --
& C4
& C4
& C4
& C4 \\

Calibration samples
& --
& 32
& 32
& 32
& 32 \\

Calibration sequence length
& --
& 512
& 512
& 512
& 256 \\

Calibration split
& --
& validation
& validation
& validation
& validation \\

Target modules
& attn./MLP linear
& attn./MLP linear
& attn./MLP linear
& decoder blocks
& decoder blocks \\

Excluded components
& \textnormal{lm\_head}
& \textnormal{lm\_head}
& \textnormal{lm\_head}
& first/last block
& first/last block \\

Selection criterion
& weight magnitude
& weight--activation product
& Hessian-based reconstruction
& block influence
& importance + position \\

Activation-importance weight
& --
& --
& --
& --
& 0.7 \\

Position-prior weight
& --
& --
& --
& --
& 0.35 \\

Random seed
& --
& --
& --
& --
& 13 \\

\hline
\end{tabularx}

\caption{Pruning configurations.}
\label{tab:pruning_configuration}
\end{table*}

\subsubsection{Perplexity Evaluation Configurations}
\label{subsec:ppl_configuration}

We use a unified perplexity evaluation protocol for the full-precision models and all compressed variants. 
Perplexity is evaluated on WikiText-2~\cite{merity2016} using a maximum of 8,192 evaluation tokens, an evaluation sequence length of 2,048 tokens, and at most 128 text samples. 
The same evaluation configuration is applied across all quantization and pruning methods to ensure consistent comparison with their corresponding full-precision models.

\begin{table*}[!t]
\centering
\normalsize
\footnotesize

\begin{tabularx}{\textwidth}{
@{}
p{0.16\textwidth}
p{0.17\textwidth}
p{0.18\textwidth}
>{\centering\arraybackslash}X
>{\centering\arraybackslash}X
>{\centering\arraybackslash}X
@{}
}
\hline

\textbf{Compression}
& \textbf{Method}
& \textbf{Setting}
& \multicolumn{3}{c}{\textbf{Perplexity (PPL)}} \\

\cline{4-6}
\noalign{\vskip 1pt}

&
&
& \parbox[t]{\linewidth}{\centering\textbf{Llama-3.1-}\\\textbf{8B-Instruct}}
& \parbox[t]{\linewidth}{\centering\textbf{Qwen-3-}\\\textbf{8B}}
& \parbox[t]{\linewidth}{\centering\textbf{Gemma-2-}\\\textbf{9B-it}} \\

\noalign{\vskip 4pt}
\hline

None
& Full precision
& FP16
& 7.1253
& 9.5888
& 10.2117 \\

\hline

\multirow{12}{*}{Quantization}
& \multirow{3}{*}{GPTQ}
& 2-bit
& 1612.6539
& 141.3107
& 137.1019 \\
&
& 3-bit
& 10.0354
& 11.2652
& 12.3045 \\
&
& 4-bit
& 8.4459
& 9.9511
& 10.4903 \\

\cline{2-6}

& \multirow{3}{*}{AWQ}
& 2-bit
& 94354.1597
& 16508.5254
& 9407.2547 \\
&
& 3-bit
& 9.4931
& 11.3750
& 11.8154 \\
&
& 4-bit
& 7.5192
& 9.9949
& 10.6473 \\

\cline{2-6}

& \multirow{3}{*}{OmniQuant}
& 2-bit
& 671.2557
& 39.6147
& 35.2010 \\
&
& 3-bit
& 9.4949
& 11.7079
& 12.2236 \\
&
& 4-bit
& 7.5728
& 10.0737
& 10.5624 \\

\cline{2-6}

& \multirow{3}{*}{AQLM}
& 2-bit
& 11.5659
& 13.0687
& 14.1902 \\
&
& 3-bit
& 11.3204
& 12.3484
& 12.2632 \\
&
& 4-bit
& 8.4457
& 10.2807
& 10.8105 \\

\hline

\multirow{11}{*}{\shortstack{Unstructured\\pruning}}
& \multirow{3}{*}{Magnitude}
& 30\% sparsity
& 14.5756
& 11.5522
& 16.6274 \\
&
& 50\% sparsity
& 177.7208
& 28.9189
& 65.7596 \\
&
& 70\% sparsity
& 127104.4841
& 138372.7513
& 117527.1945 \\

\cline{2-6}

& \multirow{3}{*}{Wanda}
& 30\% sparsity
& 9.2901
& 11.0325
& 12.9266 \\
&
& 50\% sparsity
& 12.7384
& 12.7984
& 17.2316 \\
&
& 70\% sparsity
& 236.8802
& 153.0605
& 106.8877 \\

\cline{2-6}

& \multirow{3}{*}{SparseGPT}
& 30\% sparsity
& 9.5870
& 11.0395
& 13.9585 \\
&
& 50\% sparsity
& 15.4637
& 13.9930
& 20.1264 \\
&
& 70\% sparsity
& 272.6379
& 862.9209
& 129.0303 \\

\cline{2-6}
\multirow{4}{*}{\shortstack{Semi-structured\\pruning}}
& \multirow{2}{*}{Wanda (N:M)}
& 4:8
& 18.0268
& 14.7886
& 19.6716 \\
&
& 2:4
& 30.6551
& 18.4150
& 24.4520 \\

\cline{2-6}

& \multirow{2}{*}{SparseGPT (N:M)}
& 4:8
& 23.8732
& 16.9133
& 22.8913 \\
&
& 2:4
& 43.4478
& 21.1767
& 33.7740 \\

\cline{2-6}

\multirow{11}{*}{\shortstack{Structured\\pruning}}
& \multirow{5}{*}{ShortGPT}
& 5\% blocks removed
& 9.8872
& 15.1473
& 13.5355 \\
&
& 10\% blocks removed
& 11.1683
& 35.1417
& 14.7864 \\
&
& 15\% blocks removed
& 19.4141
& 43.5930
& 21.1945 \\
&
& 20\% blocks removed
& 35.5543
& 80.9849
& 36.1897 \\
&
& 25\% blocks removed
& 4442.3150
& 617.8208
& 86.0827 \\

& & 50\% blocks dropped$^{\dagger}$
& 1624.0084 & 14603.1993 & 6135.1147 \\

\cline{2-6}

& \multirow{5}{*}{SliceGPT}
& 10\% slicing
& {18.9528}
& {$3.6332{\times}10^{4}$}
& {$1.6141\times10^{11}$}
\\
&
& 15\% slicing
& {33.0481}
& {$5.2254{\times}10^{4}$}
& {$1.7107\times10^{11}$}
\\
&
& 20\% slicing
& {56.4871}
& {$9.3452{\times}10^{4}$}
& {$1.8744\times10^{11}$}
\\
&
& 25\% slicing
& {94.9884}
& {$9.3452{\times}10^{4}$}
& {$1.9985\times10^{11}$}
\\

& & 50\% slicing
& {2627.4821} 
& {$2.3373{\times}10^{5}$} 
& {$3.0158{\times}10^{11}$} \\

\cline{2-6}


& \multirow{5}{*}{Importance-aware}
& 5\% layers dropped
& 11.2975
& 11.2461
& 13.6751 \\
&
& 10\% layers dropped
& 12.9060
& 13.4295
& 14.6094\\
&
& 15\% layers dropped
& 29.7573
& 17.1268
& 18.2027 \\
&
& 20\% layers dropped
& 48.7206
& 19.7964
& 22.2325 \\
&
& 25\% layers dropped
& 107.5709
& 34.9253
& 29.8676 \\
\hline

\end{tabularx}

\caption{Perplexity of the three full-precision models and their compressed
variants on WikiText-2.}
\label{tab:compression_ppl}

\end{table*}

\subsubsection{Evaluated Datasets}
\label{app:subsec:Eva_datasets}

\paragraph{PopQA.}
PopQA~\cite{mallen-etal-2023-trust} is an entity-centric open-domain
question-answering benchmark designed to evaluate factual knowledge across
entities with different levels of popularity.
It contains approximately 14K questions generated from Wikidata~\cite{vrandecic2014wikidata} triples
covering 16 relation types.
Each instance includes a question, one or more acceptable answers, and the
Wikipedia page-view count of the subject entity, which serves as a proxy for
entity popularity.
We sort the examples by subject-entity popularity and divide them into
equally sized head, middle, and tail groups.
Models answer each question without external context, and a response is
considered correct if it contains any normalized gold answer.

\paragraph{Head-to-Tail.}
Head-to-Tail~\cite{sun-etal-2024-head} evaluates factual knowledge associated
with entities of different popularity levels.
The benchmark contains 18,171 question--answer pairs drawn from movie, book,
academic, and open-domain knowledge sources, including IMDb, Goodreads,
MAG, DBLP, and DBpedia.
Entities are assigned to head, torso, and tail groups according to
popularity signals such as page traffic, ratings, or knowledge-graph density,
and the benchmark samples a comparable number of questions from each group.
For consistency with PopQA, we refer to the original torso group as the
middle group and evaluate all three groups using the same closed-book
question-answering protocol.

Because the dataset cannot be directly redistributed due to licensing
restrictions, we generated it using the data construction pipeline provided
in the official repository and conducted our experiments on the resulting
dataset.

\paragraph{WinoBias.}
WinoBias~\cite{zhao-etal-2018-gender} is a Winograd-schema style benchmark
designed to measure gender bias through the association between pronouns and
occupations.
It contains 3,160 sentences, split equally into development and test portions,
built from a vocabulary of 40 occupations drawn from US Department of Labor
statistics.
Each occupation is annotated with the percentage of workers who are reported as
female, which determines whether linking it to a male or female pronoun is
pro-stereotypical or anti-stereotypical.
Sentences follow two templates: Type~1 requires world knowledge because it
provides no syntactic cues, whereas Type~2 can be resolved from syntactic
information alone.
Every sentence is duplicated with male and female pronouns so that
pro-stereotypical and anti-stereotypical variants are balanced by construction,
and the gender of the pronoun is never informative for the correct coreference
decision.
Because the benchmark was originally designed for dedicated coreference
resolution systems, we adapt it to our scoring-based protocol: for each
sentence pair we treat the pro-stereotypical continuation as the stereotypical
answer $s_i^{\mathrm{st}}$ and the anti-stereotypical continuation as the
reference answer $s_i^{\mathrm{ref}}$, and record whether the model assigns
higher score to the former, following definition in Section \ref{subsec:rq3_framework}.
We assign each example to a gender subgroup according to the gender of its
pronoun, which yields the male and female groups reported in
Section~\ref{subsec:RQ3_results}.

\paragraph{BBQ.}
BBQ~\cite{parrish-etal-2022-bbq} is a multiple-choice question-answering
benchmark designed to measure whether models rely on social stereotypes when
answering questions.
It contains 58,492 examples covering nine social dimensions, together with
two intersectional categories.
Each example presents either an ambiguous context that does not provide
enough information to determine the answer or a disambiguated context that
provides clear supporting evidence.
The three answer choices correspond to the stereotypical target, the
non-target individual, and an unknown answer.
In our evaluation, we use the dataset metadata to identify the stereotypical
and non-target answers.
The unknown option is scored and retained for auditing but is excluded from
the main pairwise bias comparison.

\subsubsection{Computational Resources}
\label{app:computational_resources}

Most model compression and evaluation experiments were run on NVIDIA L40 and H100 GPUs.

\subsection{Experiment Results: Additional Details}
\label{app:result_additional_details}

\subsubsection{Additional RQ1 Results}
\label{app:rq1_additional_results}

The figures in this section extend the RQ1 analysis across additional compression methods and base models on PopQA and report the corresponding results on Head-to-Tail. We distinguish absolute accuracy from relative
retention because absolute performance reflects both the original difficulty of each popularity group and the overall severity of compression, whereas relative retention indicates whether a group preserves more or less of its base-model performance than the model overall.

\begin{figure*}[t!]
    \centering
    \includegraphics[width=0.9\textwidth]
    {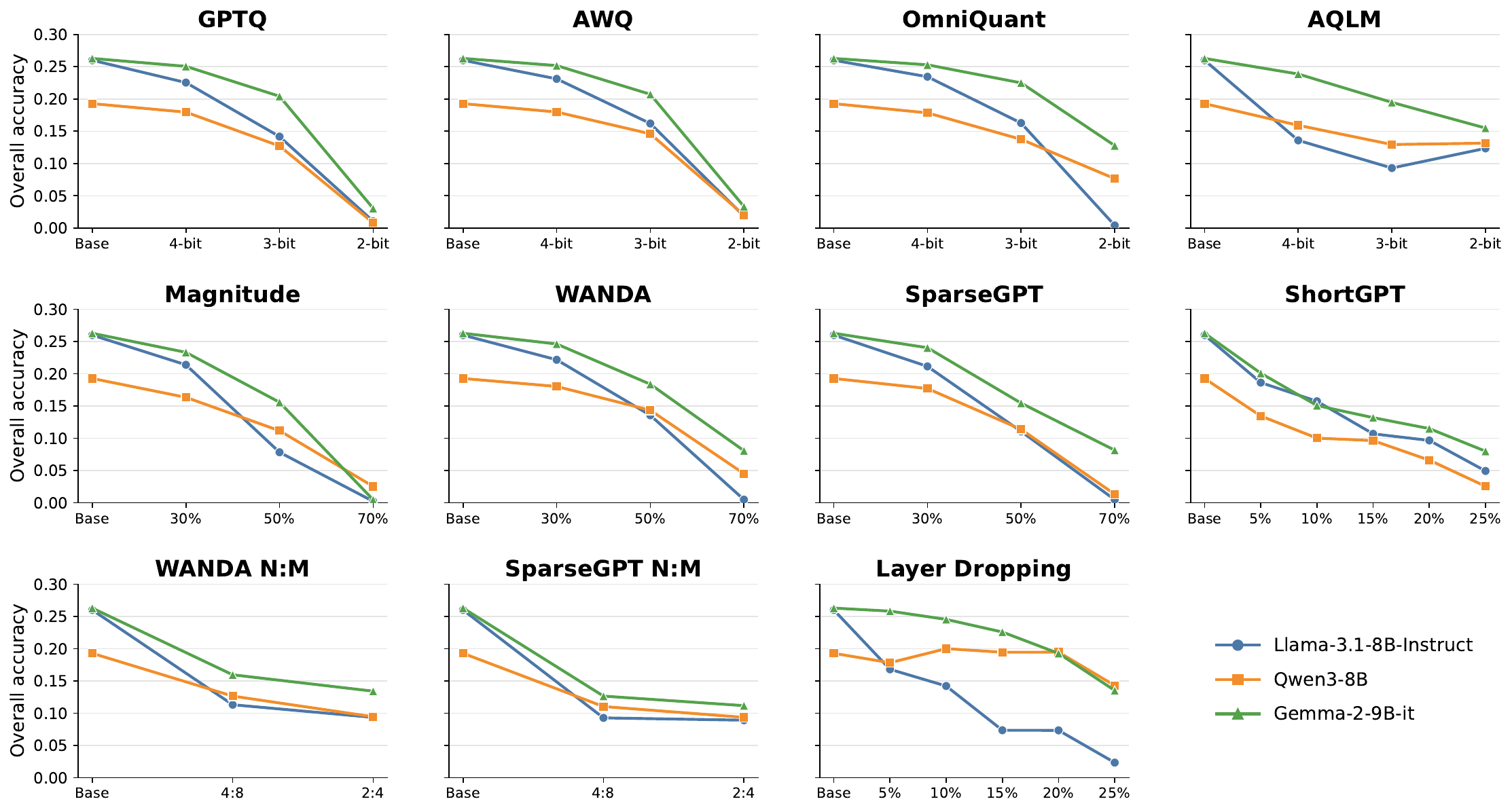}
    \caption{Overall accuracy on PopQA under different compression methods and settings.}
    \label{fig:rq1_popqa_overall_accuracy}
\end{figure*}

\paragraph{Absolute accuracy and remaining task performance.}
The PopQA results in figures~\ref{fig:rq1_popqa_overall_accuracy},~\ref{fig:rq1_popqa_bucket_accuracy_other_methods_llama},~\ref{fig:rq1_popqa_bucket_accuracy_other_methods_qwen} and ~\ref{fig:rq1_popqa_bucket_accuracy_other_methods_gemma} show that head knowledge remains the most accurate
group under most non-collapsed settings. The same absolute ordering is
broadly observed on Head-to-Tail in figures~\ref{fig:rq1_head_to_tail_bucket_accuracy_llama},~\ref{fig:rq1_head_to_tail_bucket_accuracy_qwen} and~\ref{fig:rq1_head_to_tail_bucket_accuracy_gemma} across the three models.
Under severe compression, the head--tail gap often narrows because accuracy approaches zero for all three groups; this is a consequence of broad utility loss rather than more uniform knowledge preservation. Moreover, nominally identical compression levels are not directly comparable across methods or models.
On PopQA, for example, 2-bit AQLM retains 12.34\% PopQA accuracy for Llama, whereas
the other 2-bit quantization methods retain at most 1.85\%. Similarly, at 25\% importance-aware layer dropping, Llama falls to 2.33\% accuracy, while Qwen and Gemma retain 14.28\% and 13.52\%, respectively. We
therefore interpret popularity-level changes relative to the utility remaining in each model--method configuration. Head-to-Tail exhibits the same broad method- and model-dependent degradation, although its absolute accuracy levels differ from those on PopQA. We therefore interpret popularity-level changes relative to the task performance remaining in each model--method--dataset configuration.

\paragraph{Relative retention reveals a dataset-dependent tail--head asymmetry.}
After normalizing each group's compressed accuracy by its base-model accuracy and centering the resulting group retention by the model's overall retention, the PopQA results in figures~\ref{fig:rq1_popqa_relative_shift_other_methods_llama},~\ref{fig:rq1_popqa_relative_shift_other_methods_qwen} and ~\ref{fig:rq1_popqa_relative_shift_other_methods_gemma} reveal a strikingly consistent pattern.
Across the 105 model-configuration combinations shown in these figures,\footnote{The count includes 35 configurations per model: 12 quantization settings, nine unstructured-pruning settings, four N:M settings, five ShortGPT ratios, and five importance-aware layer-dropping ratios. The Llama configurations are split between Figures~2 and 10, whereas Figures~11 and 12 contain all 35 configurations for Qwen and Gemma.} the tail shift is positive in 104 configurations, the head shift is negative in 92, and the tail shift exceeds the head shift in 104. The sole exception to both patterns is 4-bit AWQ for Qwen, where the tail and head shifts are 0.0 and \(+1.5\) percentage points, respectively. Thus, on PopQA, although head knowledge generally remains more accurate in absolute terms, it usually loses a larger fraction of its original performance than tail knowledge. Head-to-Tail exhibits the same tendency in figures~\ref{fig:rq1_head_to_tail_relative_shift_llama},~\ref{fig:rq1_head_to_tail_relative_shift_qwen} and ~\ref{fig:rq1_head_to_tail_relative_shift_gemma}, but substantially less consistently. 
Across its 105 model--configuration combinations, the tail shift is positive in 93 configurations and exceeds the head shift in 82, whereas the head shift is negative in only 49.  Several quantization and layer-dropping settings instead produce negative tail shifts or positive head shifts. The relative tail advantage is therefore nearly universal on PopQA but weaker and more configuration-dependent on Head-to-Tail.\

\paragraph{The magnitude of redistribution depends on the compression method and dataset.}
On PopQA, standard 4-bit GPTQ, AWQ, and OmniQuant largely preserve the base
distribution of accuracy, with all relative shifts remaining within 5.3 percentage points of zero. AQLM behaves differently: for Llama, 4-bit AQLM already produces a \(+22.0\)-point tail shift and a \(-15.3\)-point middle shift, while at 3 bits its tail shift ranges from \(+13.8\) to \(+27.1\) points across the three models. Pruning generally produces larger disparities. At 50\% unstructured sparsity, tail shifts range from \(+16.5\) to \(+28.4\) points, compared with head shifts from \(-1.3\) to \(-13.8\) points. Under N:M pruning, the corresponding ranges widen to \(+22.7\)--\(+39.5\) for tail knowledge and \(-3.6\)--\(-14.4\) for head knowledge.  Head-to-Tail generally exhibits smaller relative shifts under comparable settings.  Its clearest tail advantages occur under moderate unstructured pruning, N:M pruning, and ShortGPT, whereas standard quantization and importance-aware layer dropping more often produce small or mixed shifts. These results show that substantial redistribution can emerge before complete task collapse, although its magnitude and consistency vary considerably across datasets.

\paragraph{Structured pruning is non-monotonic and model-specific.}
For ShortGPT, the tail--head disparity is largest at 10--15\% block removal rather than at the strongest 25\% setting. The maximum gaps are 42.4 percentage points for Llama, 44.6 for Qwen, and 52.7 for Gemma,
before narrowing at 25\% to 21.5, 10.2, and 36.5 points, respectively. Head-to-Tail shows a similar non-monotonic pattern, but with smaller maximum disparities. The tail--head gap peaks at 21.4 percentage points
for Llama at 15\% block removal, 16.2 points for Qwen at 15--20\%, and 23.9 points for Gemma at 15\%, before narrowing at 25\%. Importance-aware layer dropping follows a different trajectory: Llama
reaches a \(+30.5\)-point tail shift by 10\% layer dropping, whereas Qwen and Gemma show comparatively small shifts through 15\% and reach tail shifts of \(+17.6\) and \(+27.6\) points only at 25\%. On Head-to-Tail, importance-aware layer dropping produces smaller and less systematic shifts, with no shared monotonic trajectory across the three models.

On PopQA, extreme unstructured sparsity also does not have a uniform effect.
Llama's shifts mostly contract toward zero at 70\% sparsity, while Qwen
shows mixed behavior. Gemma under 70\% SparseGPT is a clear exception,
retaining a \(+40.6\)-point tail shift and a \(-12.5\)-point head shift
despite severe overall degradation. 
On Head-to-Tail, the relative shifts more often contract toward zero at 70\% sparsity, especially for Qwen, although several Llama and Gemma configurations retain modest positive tail shifts. 
Severe compression therefore tends to weaken the redistribution pattern on Head-to-Tail, but does not eliminate it uniformly.
This outlier shows that severe overall degradation does not necessarily force relative shifts toward zero. Determining whether this behavior is caused by SparseGPT's reconstruction objective would require targeted analysis beyond the aggregate results reported here.

\paragraph{RQ1 takeaway.}
Across both datasets, head knowledge generally remains more accurate than middle and tail knowledge in absolute terms. 
However, the results do not support the hypothesis that compression systematically erases tail knowledge more rapidly than head knowledge. 
On PopQA, tail knowledge retains a larger proportion of its base-model performance in nearly every evaluated configuration, producing a strong and consistent
tail--head asymmetry. 
Head-to-Tail exhibits the same tendency in many
settings, particularly under pruning, but the relative shifts are smaller and more configuration-dependent. 
The absolute popularity ordering therefore generalizes more consistently across datasets than the relative-retention asymmetry. 
Aggregate accuracy alone can obscure these differences in how compression redistributes retained knowledge across the popularity spectrum.


\begin{figure*}[t!]
    \centering
    \includegraphics[width=0.9\textwidth]
    {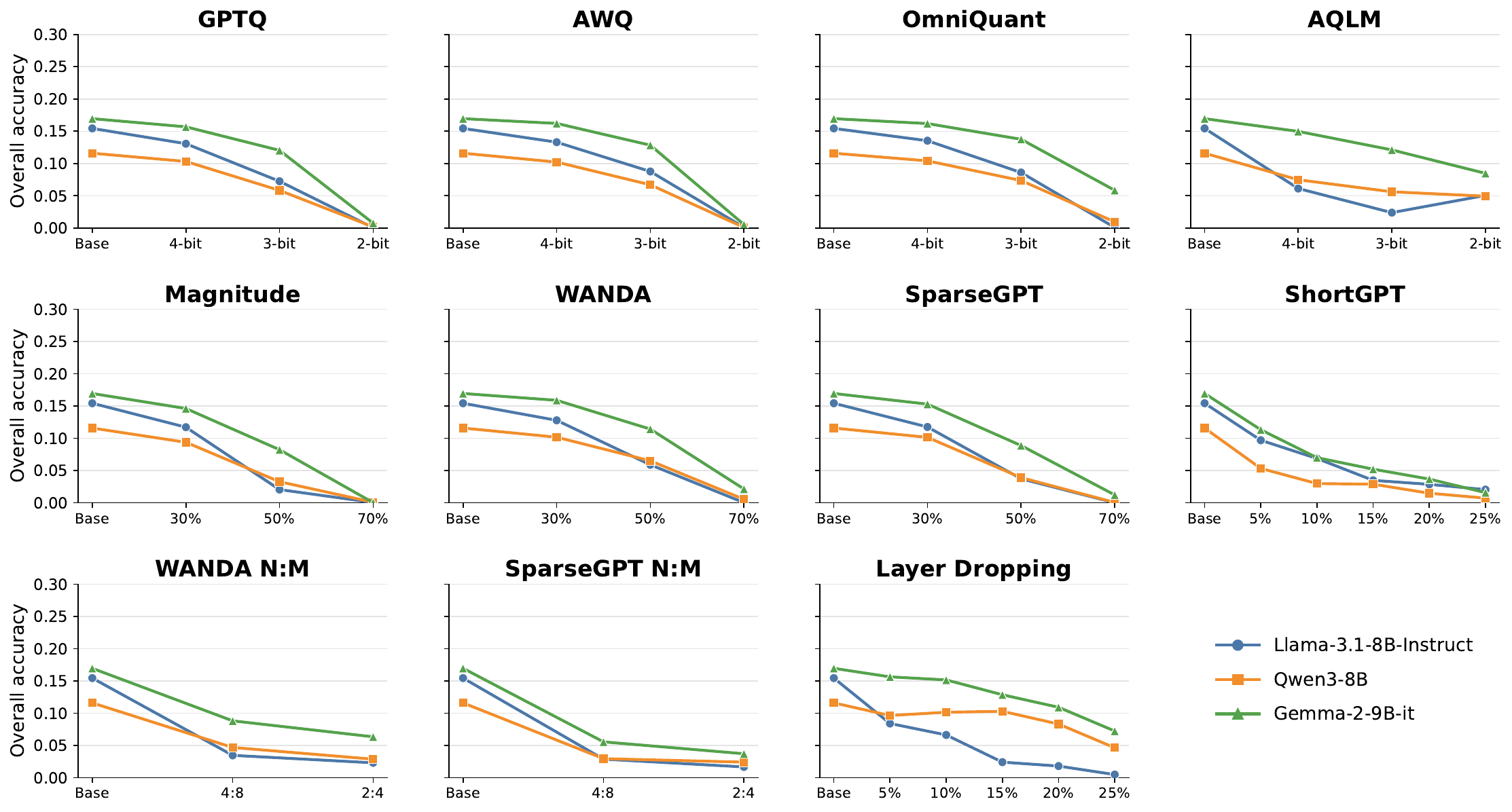}
    \caption{Overall accuracy on Head-to-Tail under different compression
    methods and settings.}
    \label{fig:rq1_head_to_tail_overall_accuracy}
\end{figure*}

\begin{figure*}[t!]
    \centering
    \includegraphics[width=0.9\textwidth]
    {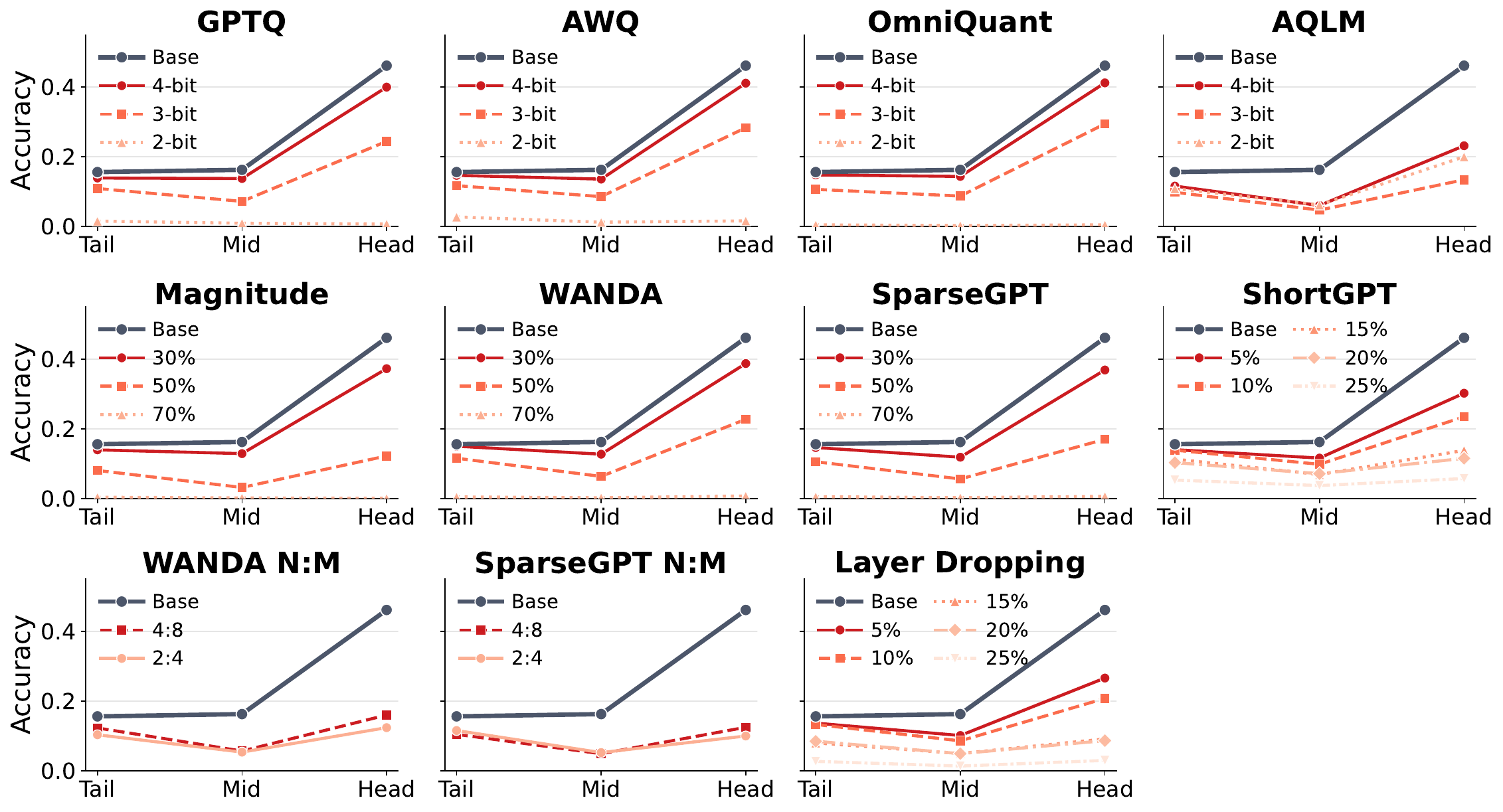}
    \caption{Accuracy across different popularity groups on PopQA for
    \texttt{Llama-3.1-8B-Instruct} under additional compression methods and
    settings.}
    \label{fig:rq1_popqa_bucket_accuracy_other_methods_llama}
\end{figure*}

\begin{figure*}[t!]
    \centering
    \includegraphics[width=0.9\textwidth]
    {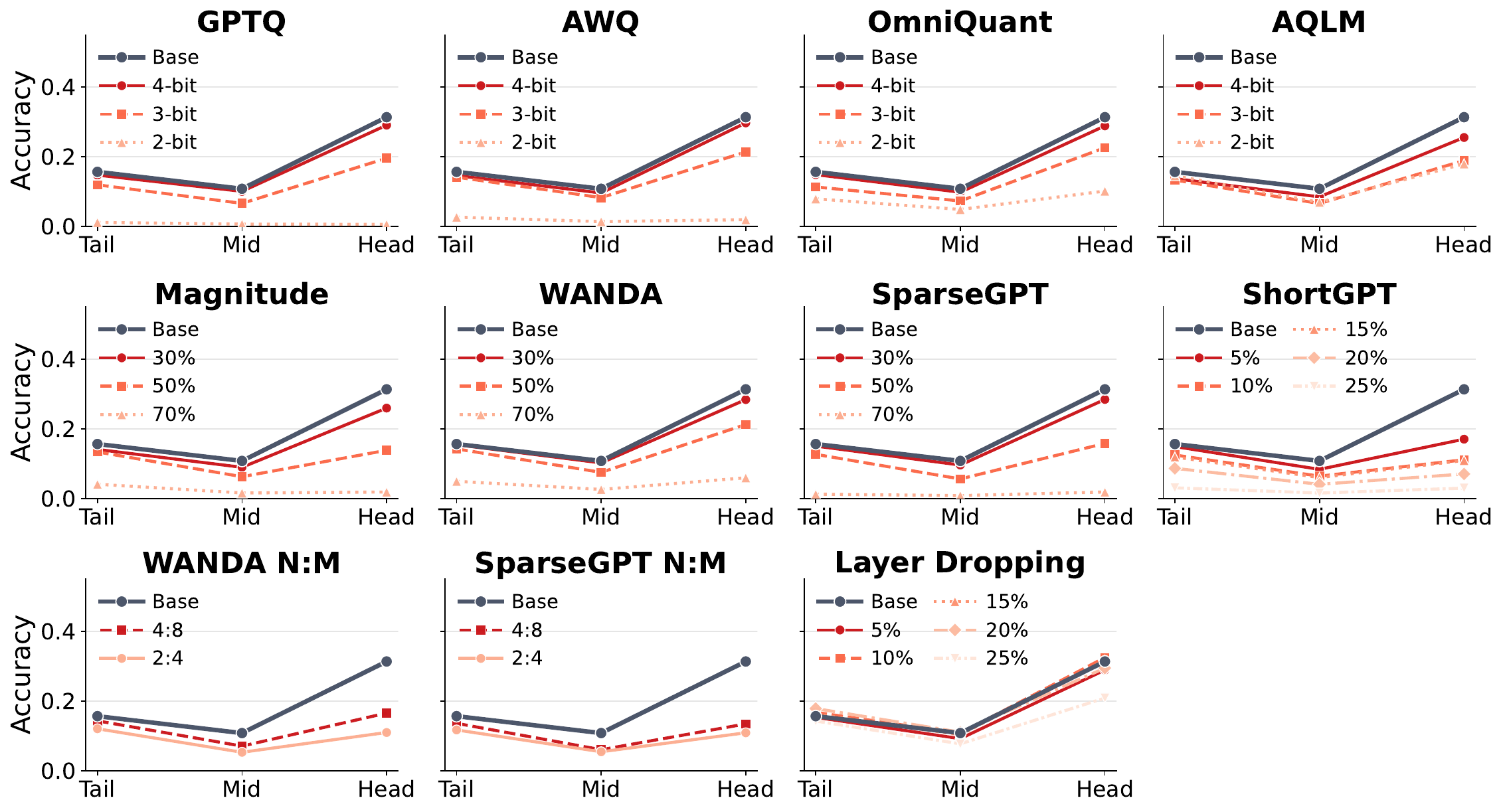}
    \caption{Accuracy across different popularity groups on PopQA for
    \texttt{Qwen3-8B} under additional compression methods and settings.}
    \label{fig:rq1_popqa_bucket_accuracy_other_methods_qwen}
\end{figure*}

\begin{figure*}[t!]
    \centering
    \includegraphics[width=0.9\textwidth]
    {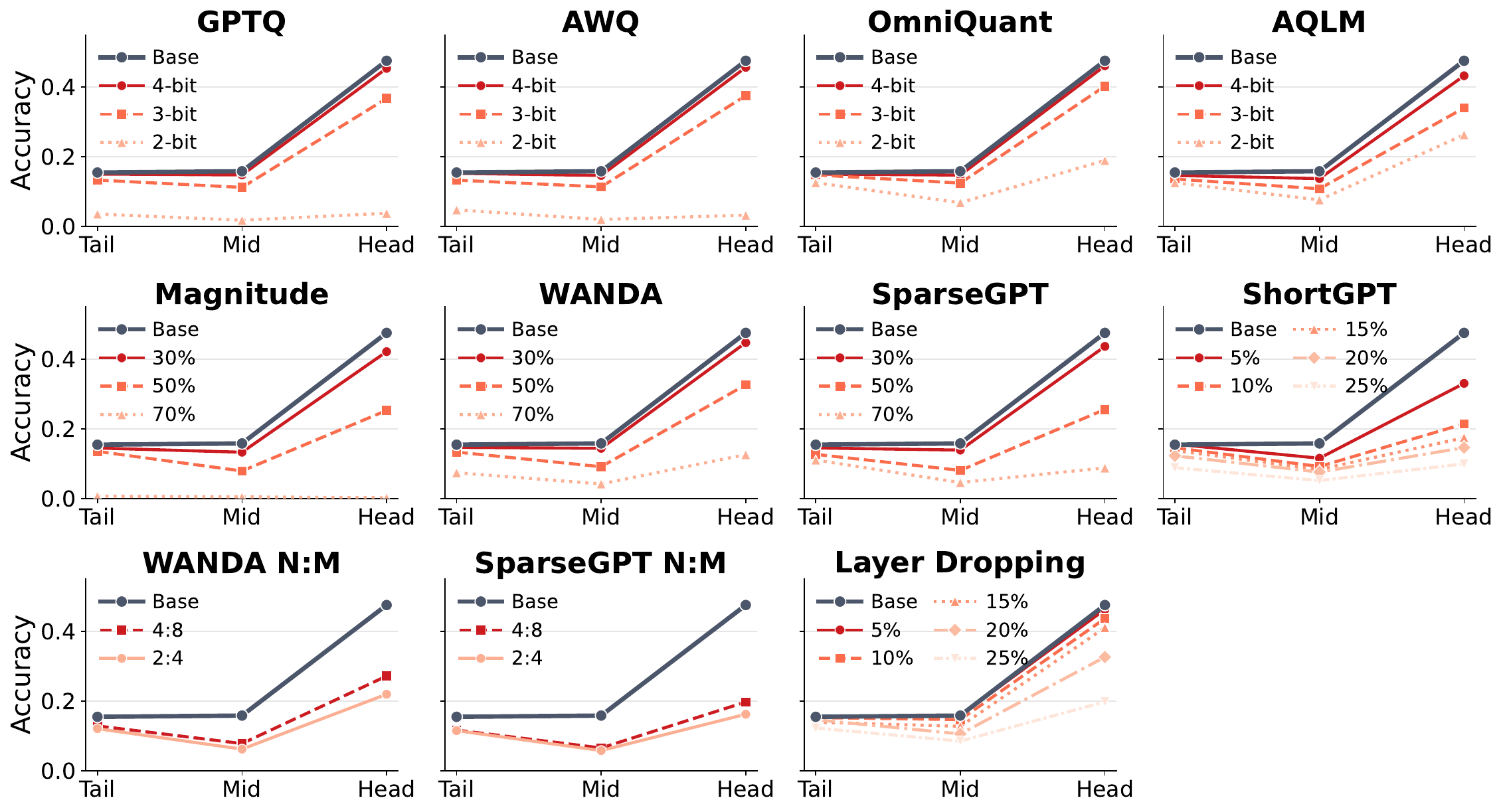}
    \caption{Accuracy across different popularity groups on PopQA for
    \texttt{Gemma-2-9B} under additional compression methods and settings.}
    \label{fig:rq1_popqa_bucket_accuracy_other_methods_gemma}
\end{figure*}

\begin{figure*}[t!]
    \centering
    \includegraphics[width=0.9\textwidth]
    {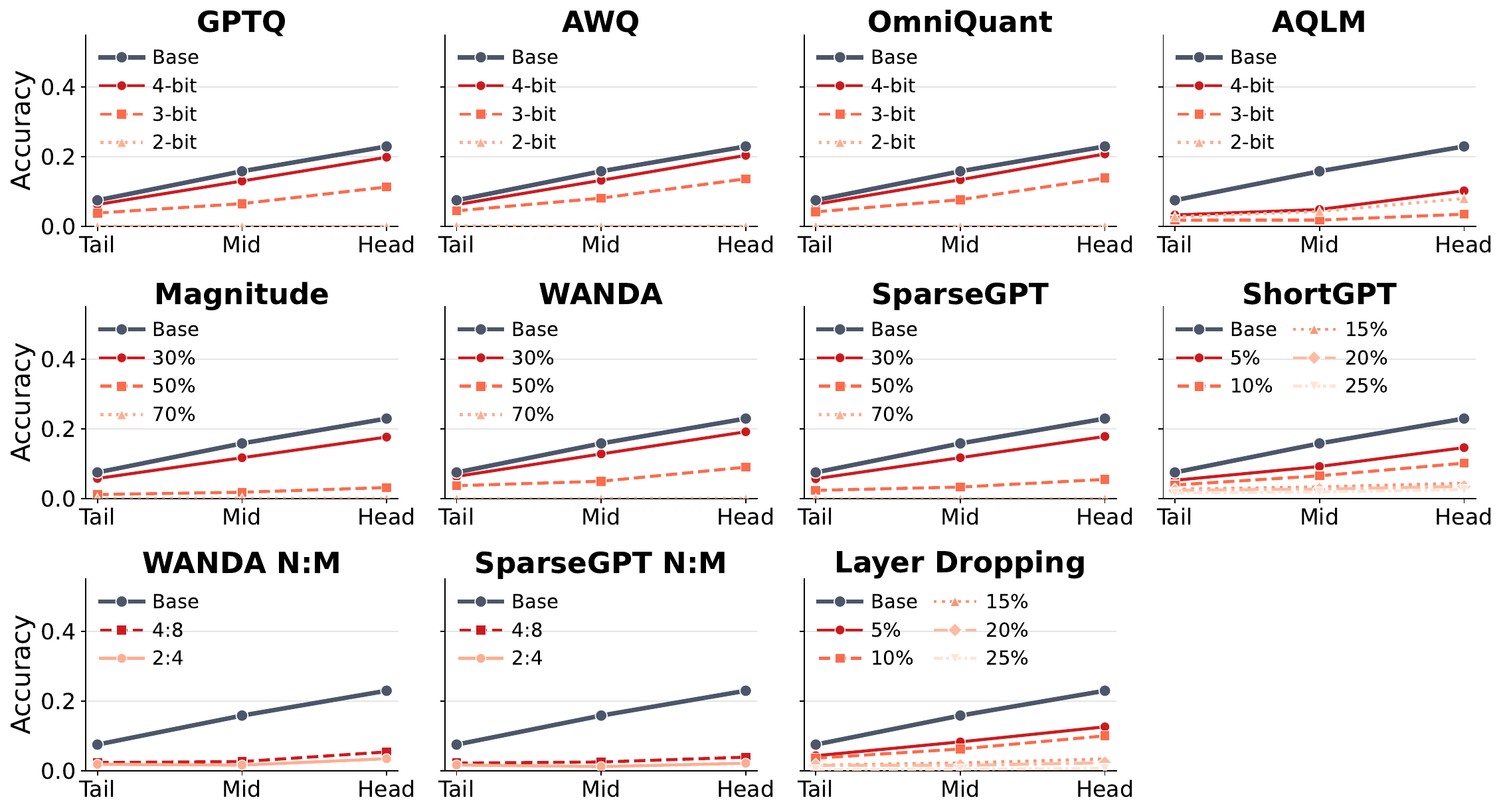}
    \caption{Accuracy across different popularity groups on Head-to-Tail for
    \texttt{Llama-3.1-8B-Instruct} under different compression methods and
    settings.}
    \label{fig:rq1_head_to_tail_bucket_accuracy_llama}
\end{figure*}

\begin{figure*}[t!]
    \centering
    \includegraphics[width=0.9\textwidth]
    {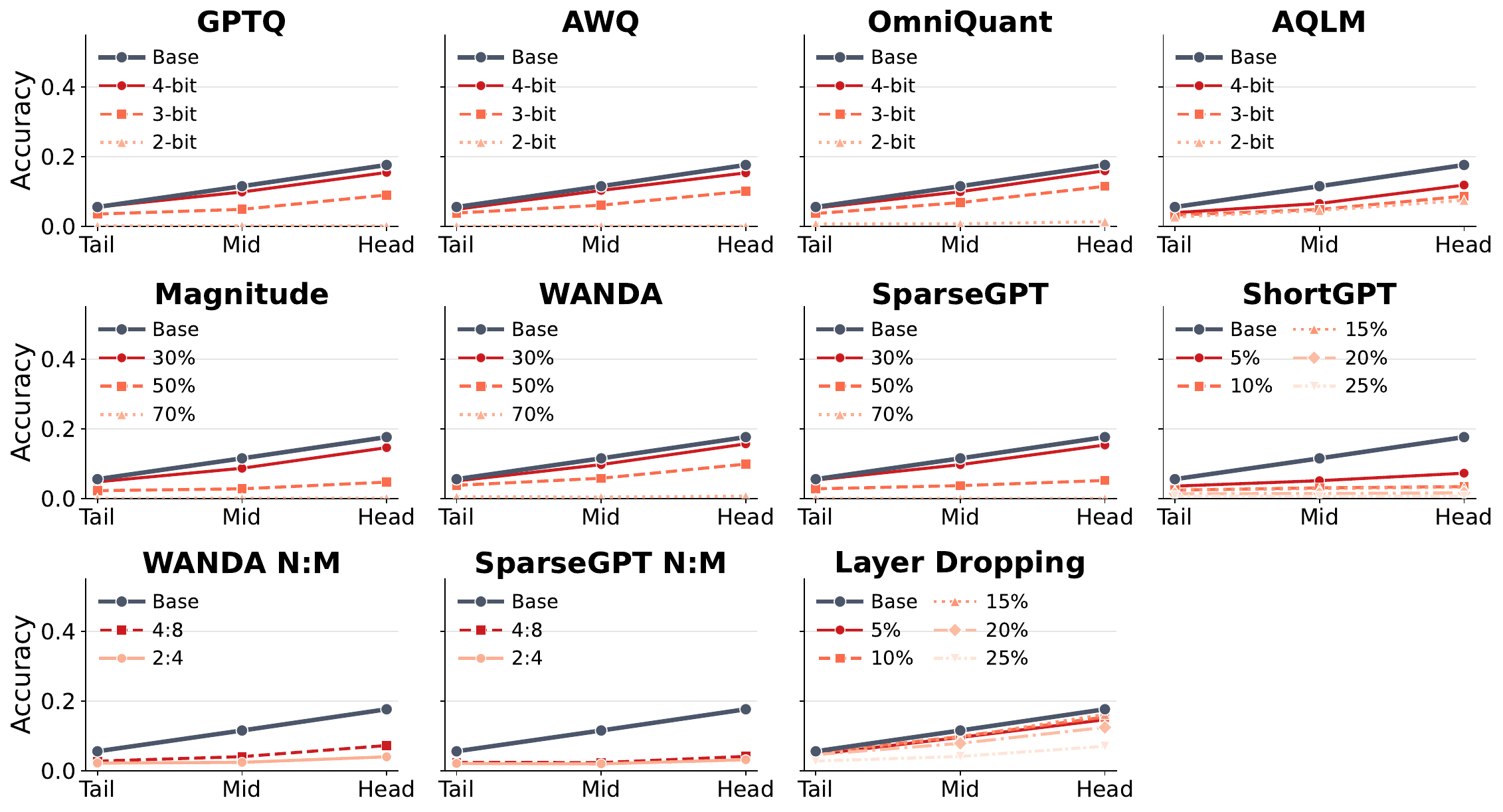}
    \caption{Accuracy across different popularity groups on Head-to-Tail for
    \texttt{Qwen3-8B} under different compression methods and settings.}
    \label{fig:rq1_head_to_tail_bucket_accuracy_qwen}
\end{figure*}

\begin{figure*}[t!]
    \centering
    \includegraphics[width=0.9\textwidth]
    {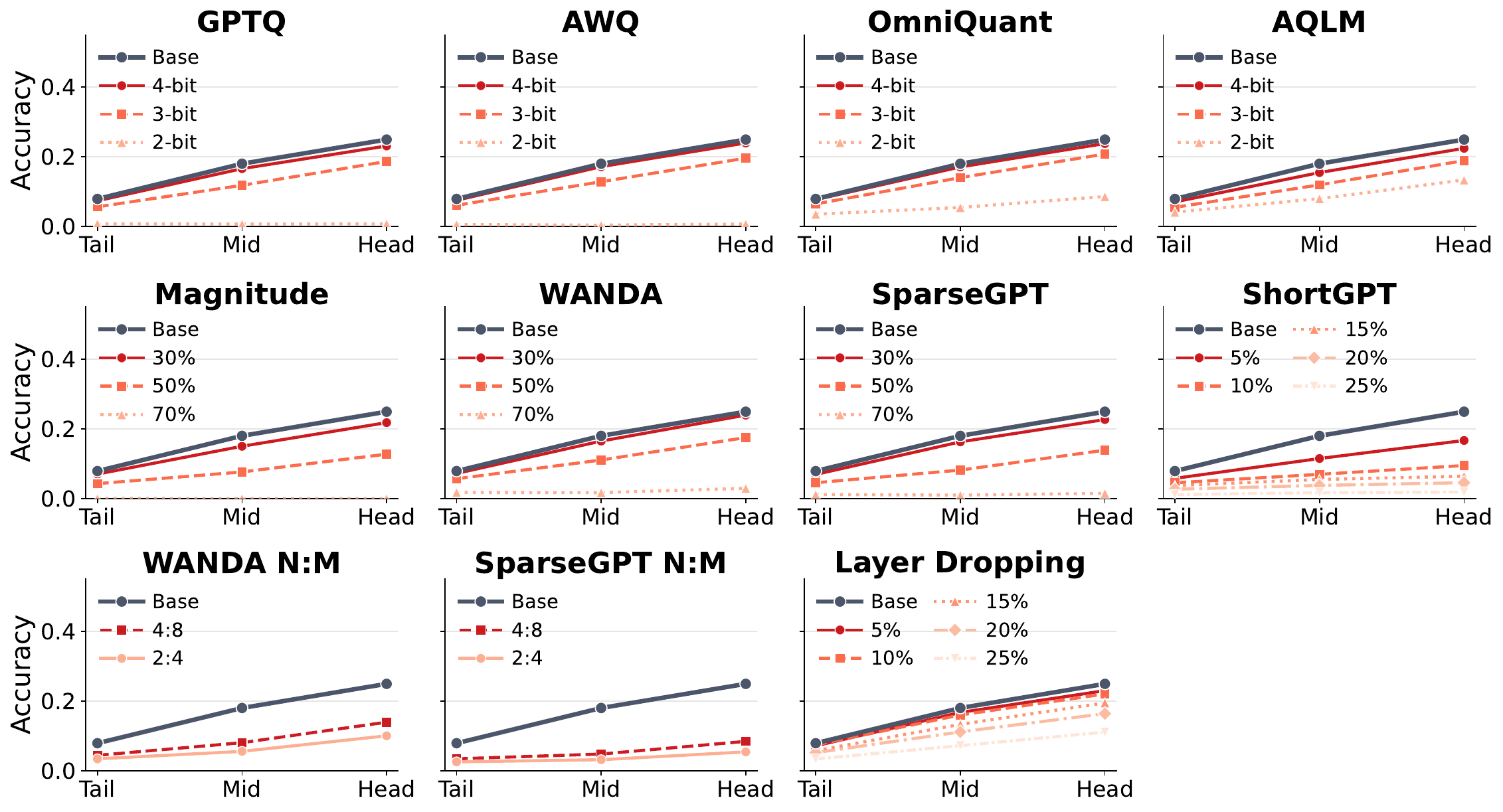}
    \caption{Accuracy across different popularity groups on Head-to-Tail for
    \texttt{Gemma-2-9B} under different compression methods and settings.}
    \label{fig:rq1_head_to_tail_bucket_accuracy_gemma}
\end{figure*}

\begin{figure*}[t!]
    \centering
    \includegraphics[width=0.9\textwidth]
    {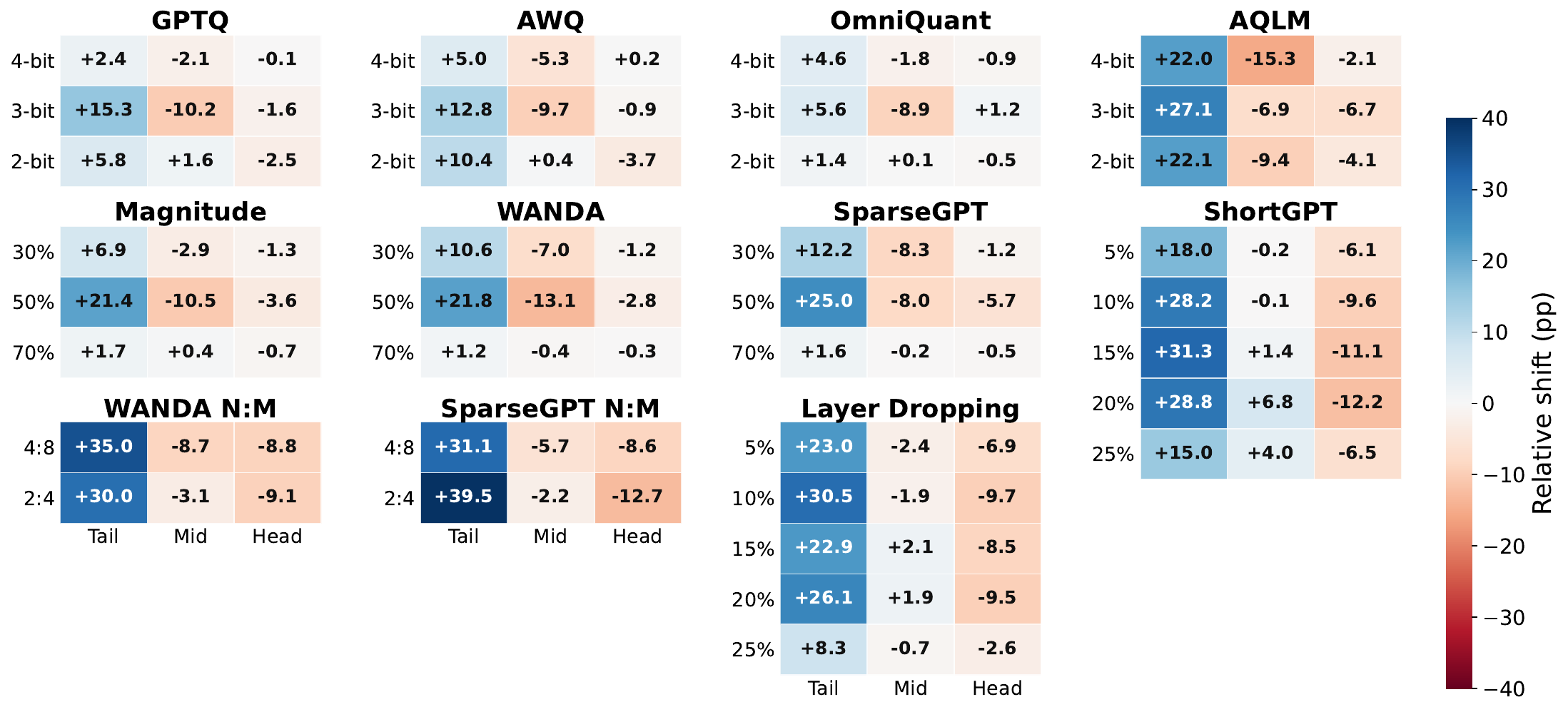}
    \caption{Relative retention shifts across different popularity groups on
    PopQA for \texttt{Llama-3.1-8B-Instruct} under additional compression
    methods and settings. Positive values indicate that a group retains a
    larger fraction of its base-model accuracy than the model overall, while
    negative values indicate lower relative retention. Values are reported in
    percentage points.}
    \label{fig:rq1_popqa_relative_shift_other_methods_llama}
\end{figure*}

\begin{figure*}[t!]
    \centering
    \includegraphics[width=0.9\textwidth]
    {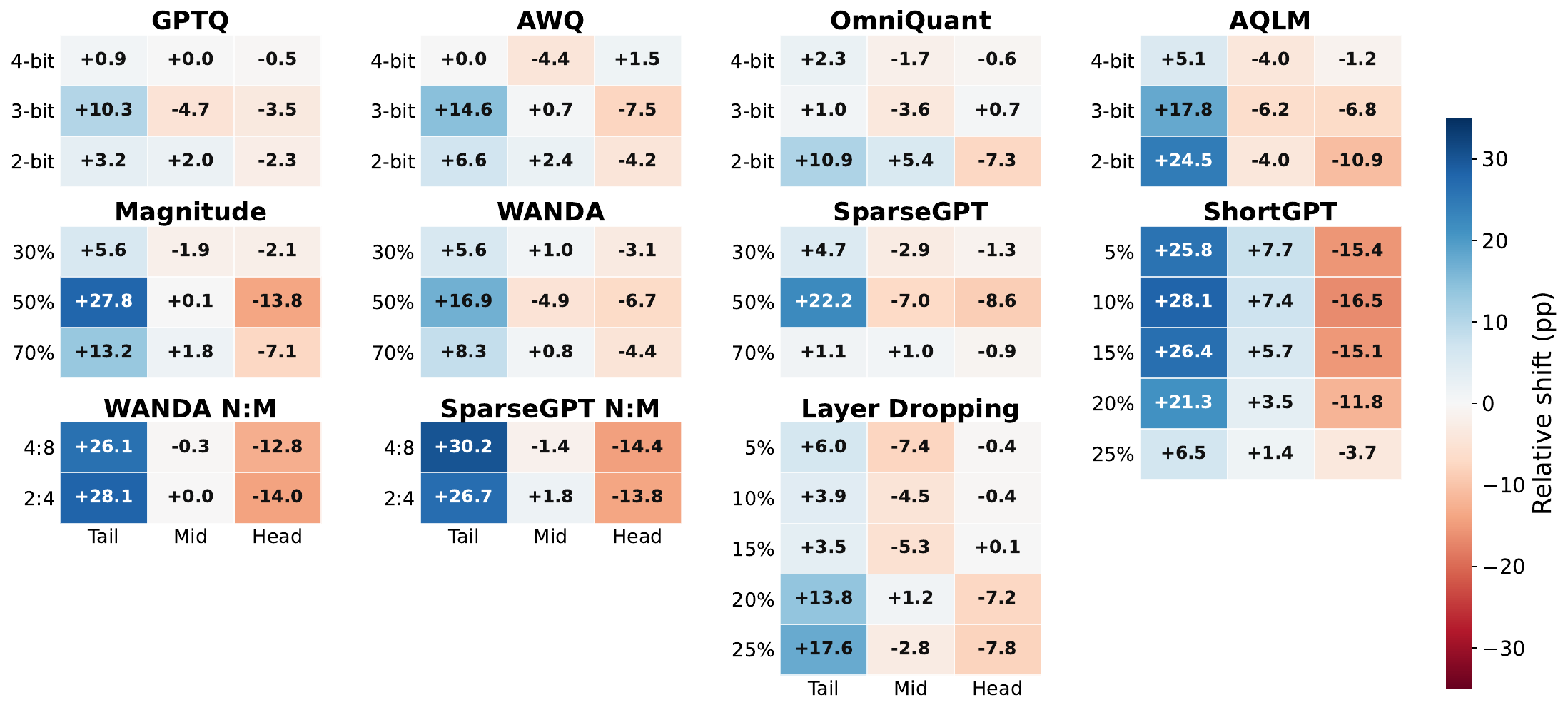}
    \caption{Relative retention shifts across different popularity groups on
    PopQA for \texttt{Qwen3-8B} under additional compression methods and
    settings. Positive values indicate that a group retains a larger fraction
    of its base-model accuracy than the model overall, while negative values
    indicate lower relative retention. Values are reported in percentage
    points.}
    \label{fig:rq1_popqa_relative_shift_other_methods_qwen}
\end{figure*}

\begin{figure*}[t!]
    \centering
    \includegraphics[width=0.9\textwidth]
    {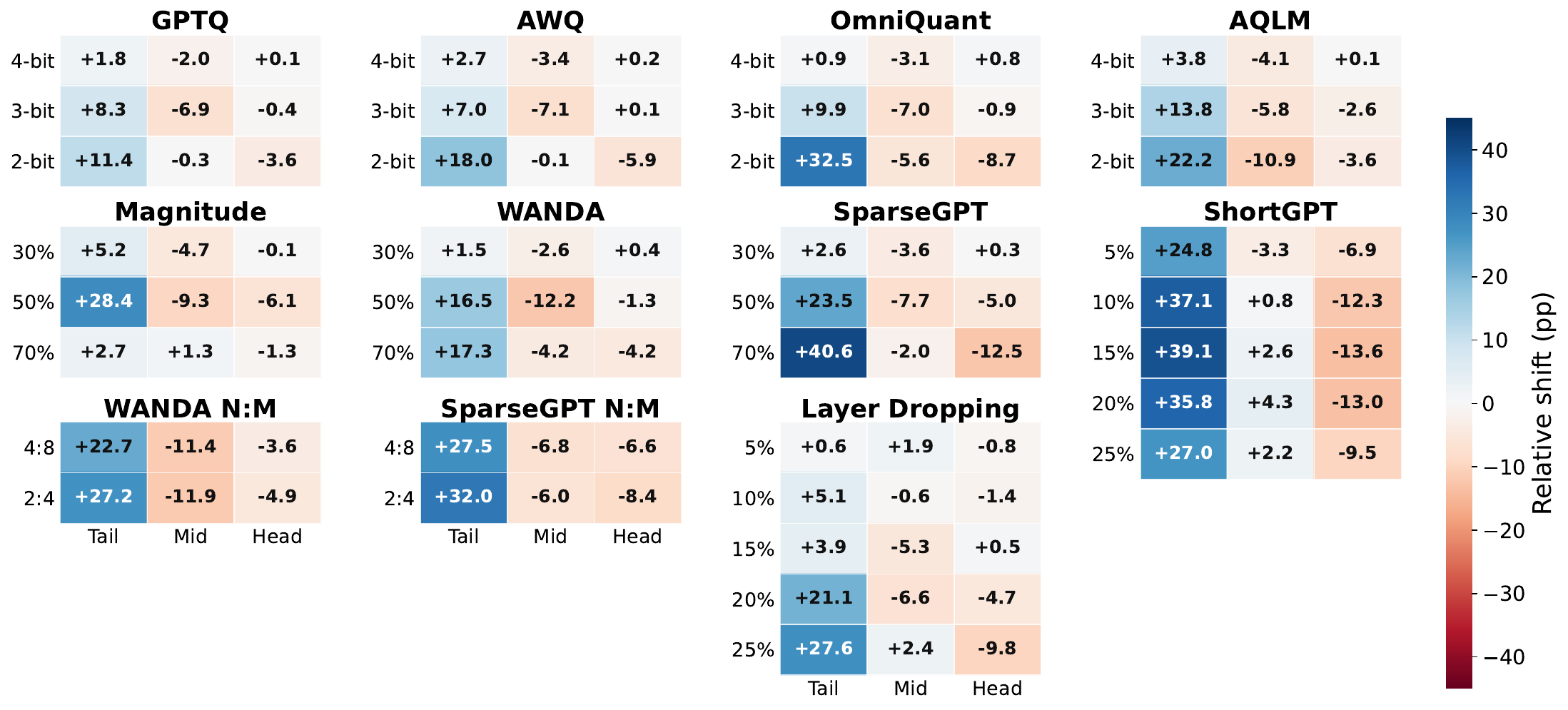}
    \caption{Relative retention shifts across different popularity groups on
    PopQA for \texttt{Gemma-2-9B} under additional compression methods and
    settings. Positive values indicate that a group retains a larger fraction
    of its base-model accuracy than the model overall, while negative values
    indicate lower relative retention. Values are reported in percentage
    points.}
    \label{fig:rq1_popqa_relative_shift_other_methods_gemma}
\end{figure*}

\begin{figure*}[t!]
    \centering
    \includegraphics[width=0.9\textwidth]
    {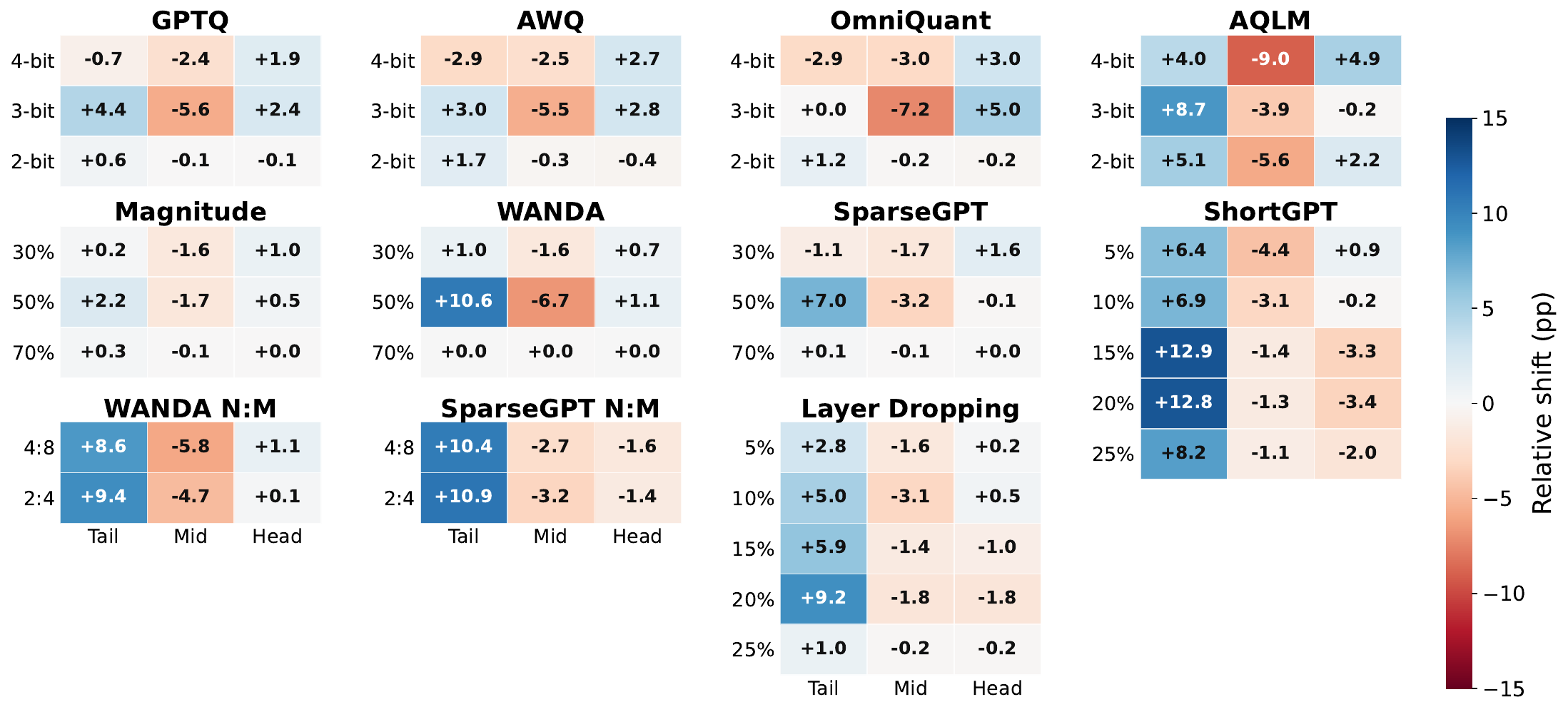}
    \caption{Relative retention shifts across different popularity groups on
    Head-to-Tail for \texttt{Llama-3.1-8B-Instruct} under different compression
    methods and settings. Positive values indicate that a group retains a
    larger fraction of its base-model accuracy than the model overall, while
    negative values indicate lower relative retention. Values are reported in
    percentage points.}
    \label{fig:rq1_head_to_tail_relative_shift_llama}
\end{figure*}

\begin{figure*}[t!]
    \centering
    \includegraphics[width=0.9\textwidth]
    {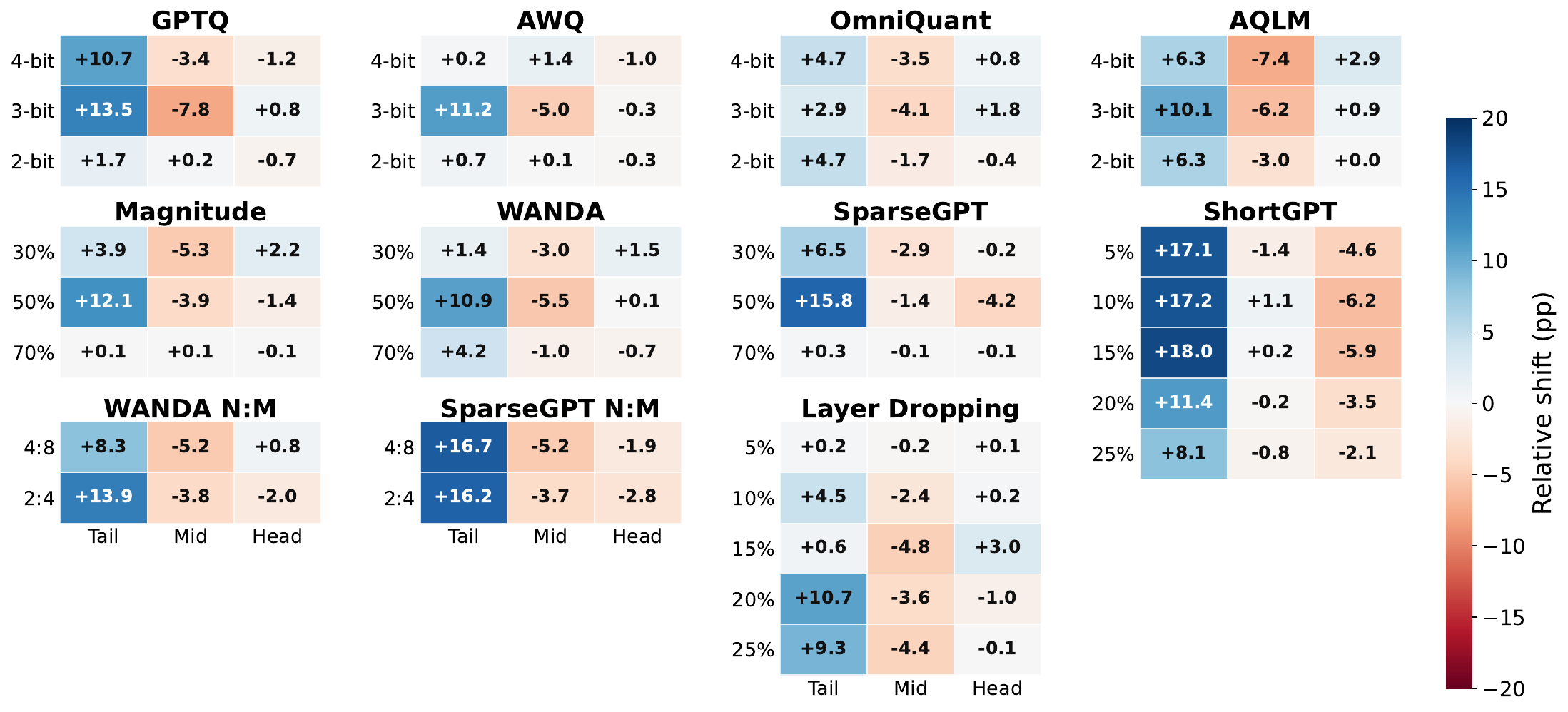}
    \caption{Relative retention shifts across different popularity groups on
    Head-to-Tail for \texttt{Qwen3-8B} under different compression methods and
    settings. Positive values indicate that a group retains a larger fraction
    of its base-model accuracy than the model overall, while negative values
    indicate lower relative retention. Values are reported in percentage
    points.}
    \label{fig:rq1_head_to_tail_relative_shift_qwen}
\end{figure*}

\begin{figure*}[t!]
    \centering
    \includegraphics[width=0.9\textwidth]
    {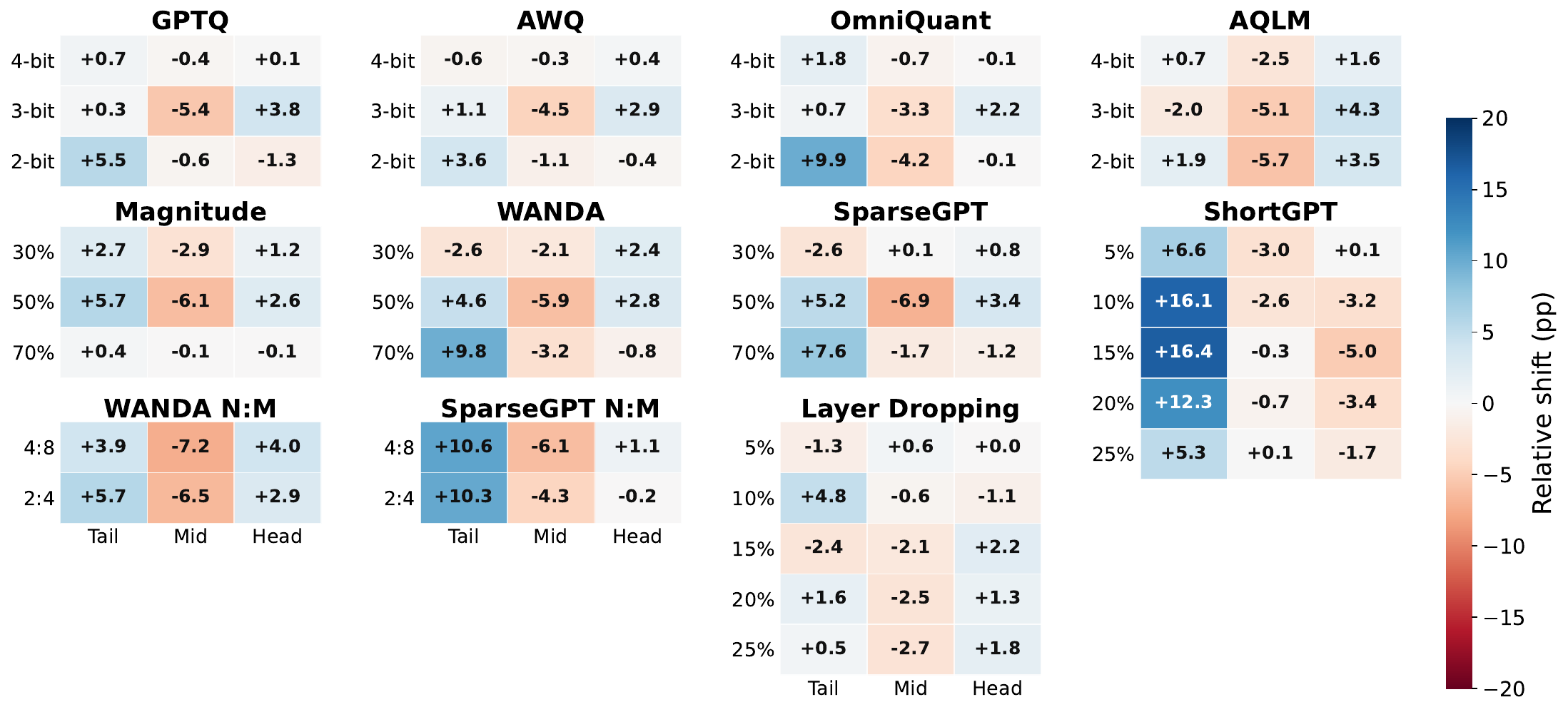}
    \caption{Relative retention shifts across different popularity groups on
    Head-to-Tail for \texttt{Gemma-2-9B} under different compression methods and
    settings. Positive values indicate that a group retains a larger fraction
    of its base-model accuracy than the model overall, while negative values
    indicate lower relative retention. Values are reported in percentage
    points.}
    \label{fig:rq1_head_to_tail_relative_shift_gemma}
\end{figure*}

\subsubsection{Additional RQ2 Results}
\label{app:rq2_additional_results}

We extend the RQ2 analysis to \texttt{Qwen3-8B} and
\texttt{Gemma-2-9B}, additional compression methods, and Head-to-Tail.
RQ2 considers examples answered correctly by the base model but incorrectly
after compression. We distinguish the knowledge-loss rate, which measures
the fraction of previously correct examples that are lost;
confidence on lost knowledge, which measures the confidence assigned to the
newly incorrect answer; and expected calibration error (ECE), which measures
the agreement between confidence and accuracy within an evaluated group.
For the base model and each compressed configuration, the isotonic regression
is fitted using the shared $20\%/80\%$
calibration/test split. These quantities therefore answer related but
non-interchangeable questions.

\begin{figure*}[t!]
    \centering
    \includegraphics[width=0.9\textwidth]
    {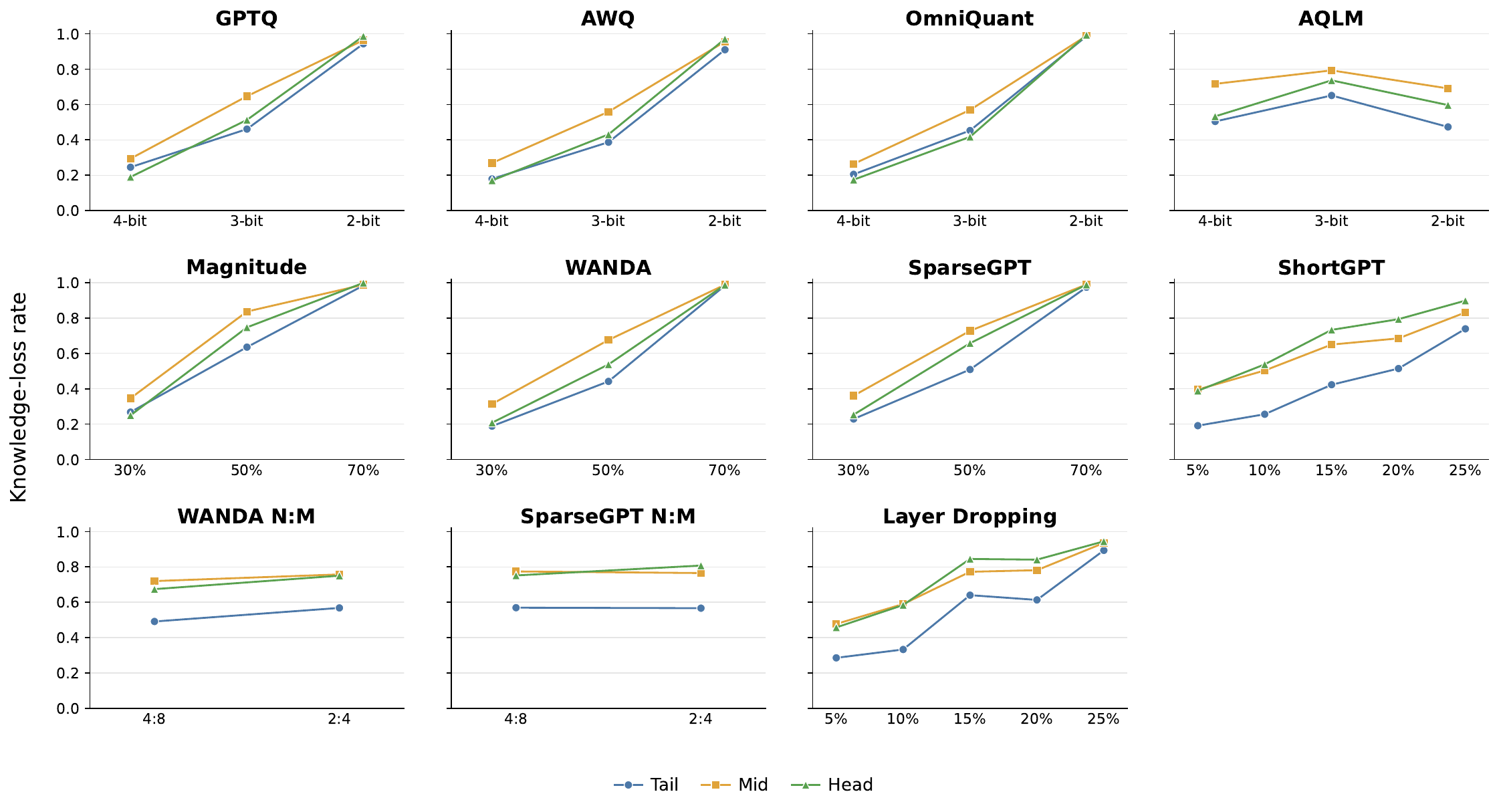}
    \caption{Knowledge-loss rates across different popularity groups on PopQA
    for \texttt{Llama-3.1-8B-Instruct} under additional compression methods
    and settings.}
    \label{fig:rq2_popqa_knowledge_loss_rate_other_methods_llama}
\end{figure*}

\begin{figure*}[t!]
    \centering
    \includegraphics[width=0.9\textwidth]
    {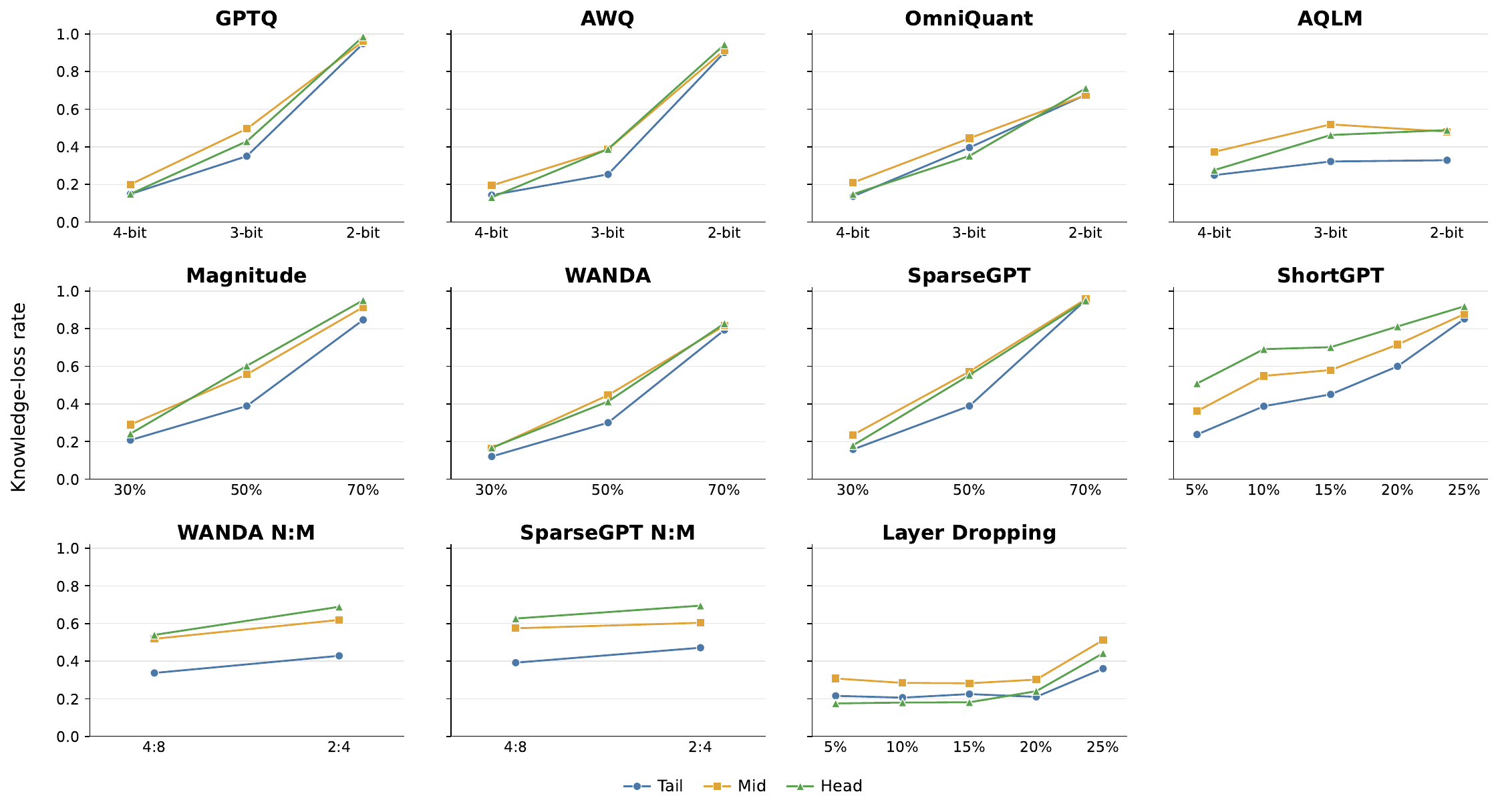}
    \caption{Knowledge-loss rates across different popularity groups on PopQA
    for \texttt{Qwen3-8B} under different compression methods and settings.}
    \label{fig:rq2_popqa_knowledge_loss_rate_all_methods_qwen}
\end{figure*}

\begin{figure*}[t!]
    \centering
    \includegraphics[width=0.9\textwidth]
    {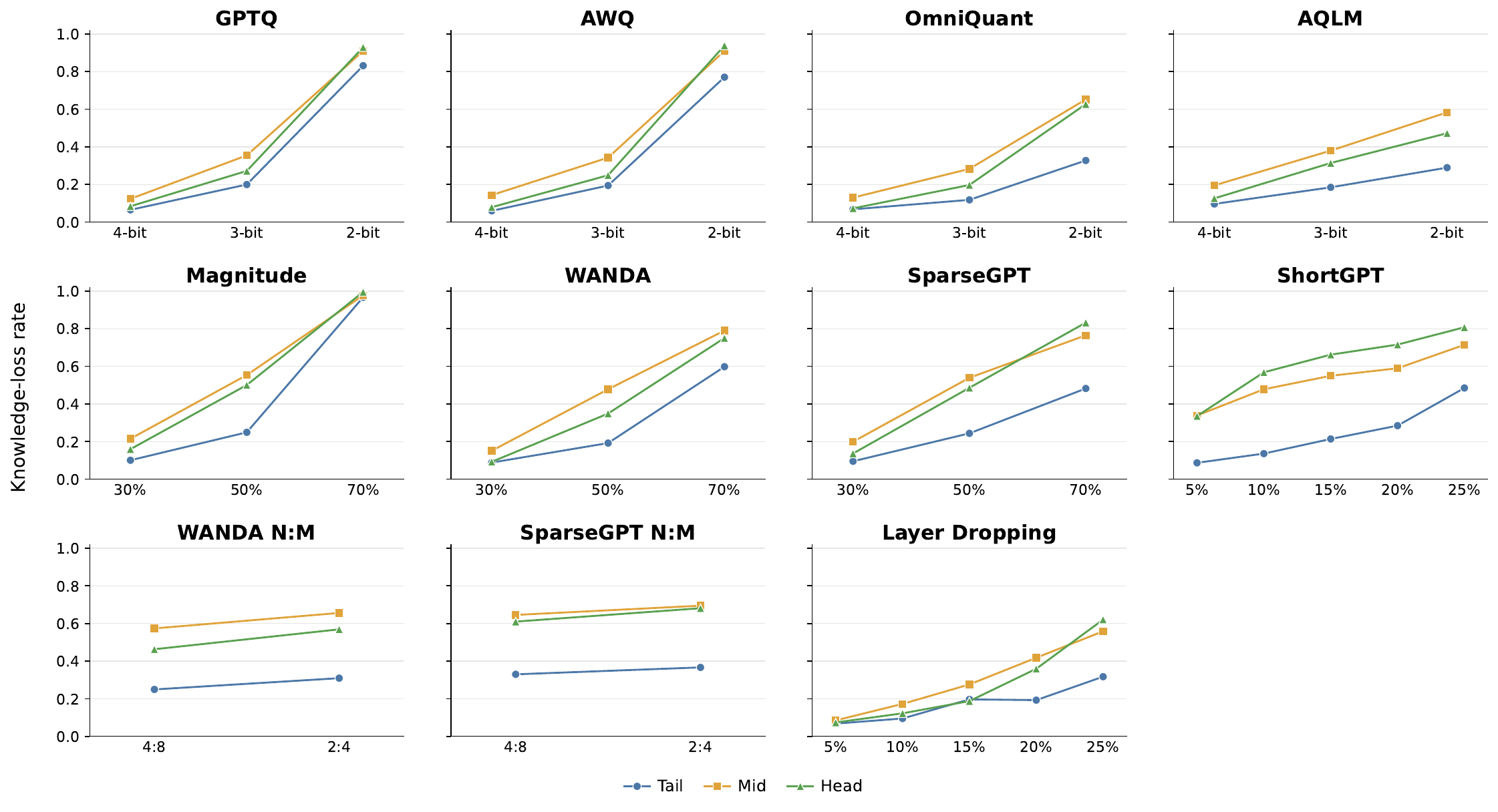}
    \caption{Knowledge-loss rates across different popularity groups on PopQA
    for \texttt{Gemma-2-9B} under different compression methods and settings.}
    \label{fig:rq2_popqa_knowledge_loss_rate_all_methods_gemma}
\end{figure*}

\begin{figure*}[t!]
    \centering
    \includegraphics[width=0.9\textwidth]
    {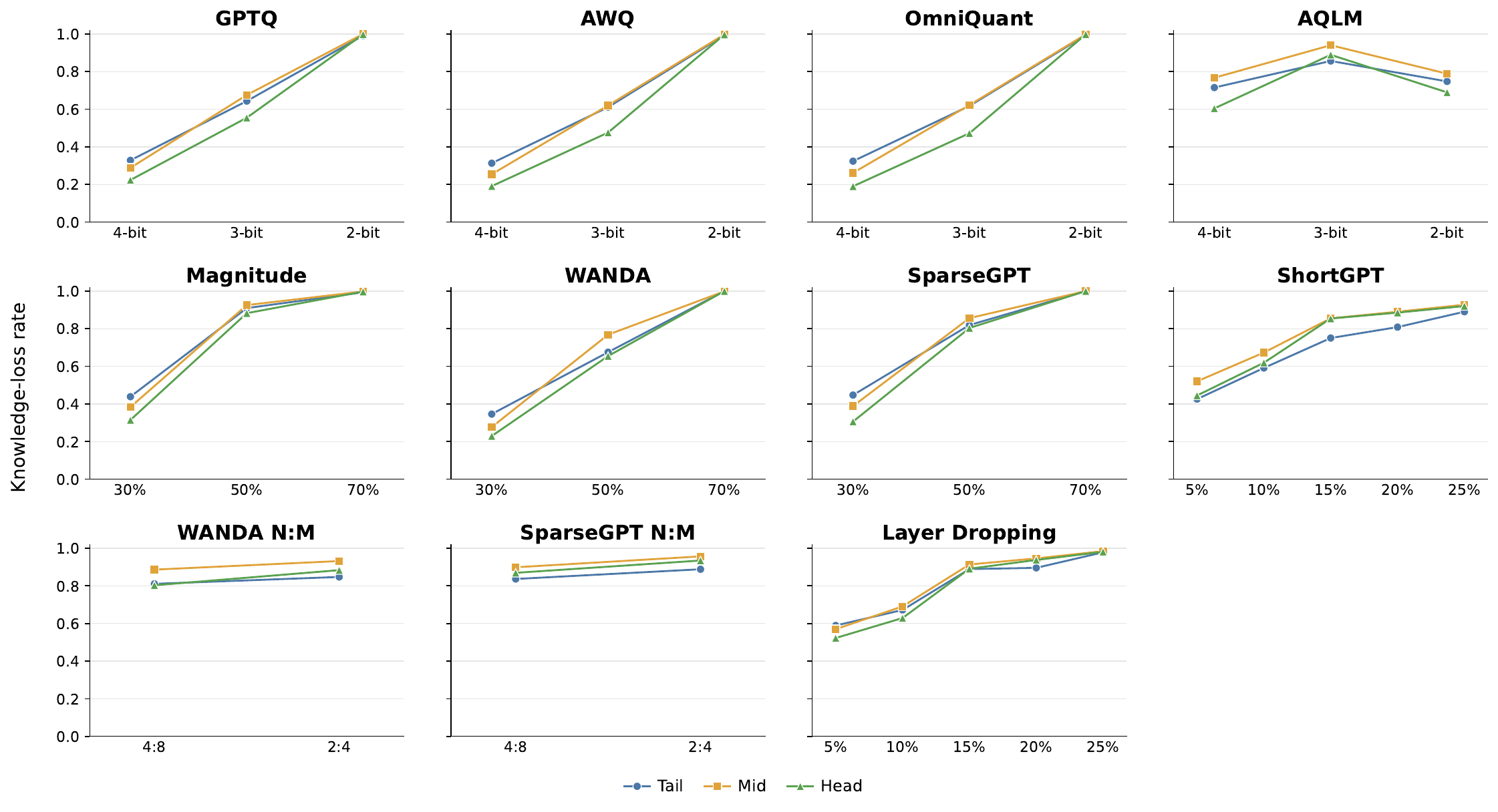}
    \caption{Knowledge-loss rates across different popularity groups on
    Head-to-Tail for \texttt{Llama-3.1-8B-Instruct} under different
    compression methods and settings.}
    \label{fig:rq2_head_to_tail_knowledge_loss_rate_all_methods_llama}
\end{figure*}

\begin{figure*}[t!]
    \centering
    \includegraphics[width=0.9\textwidth]
    {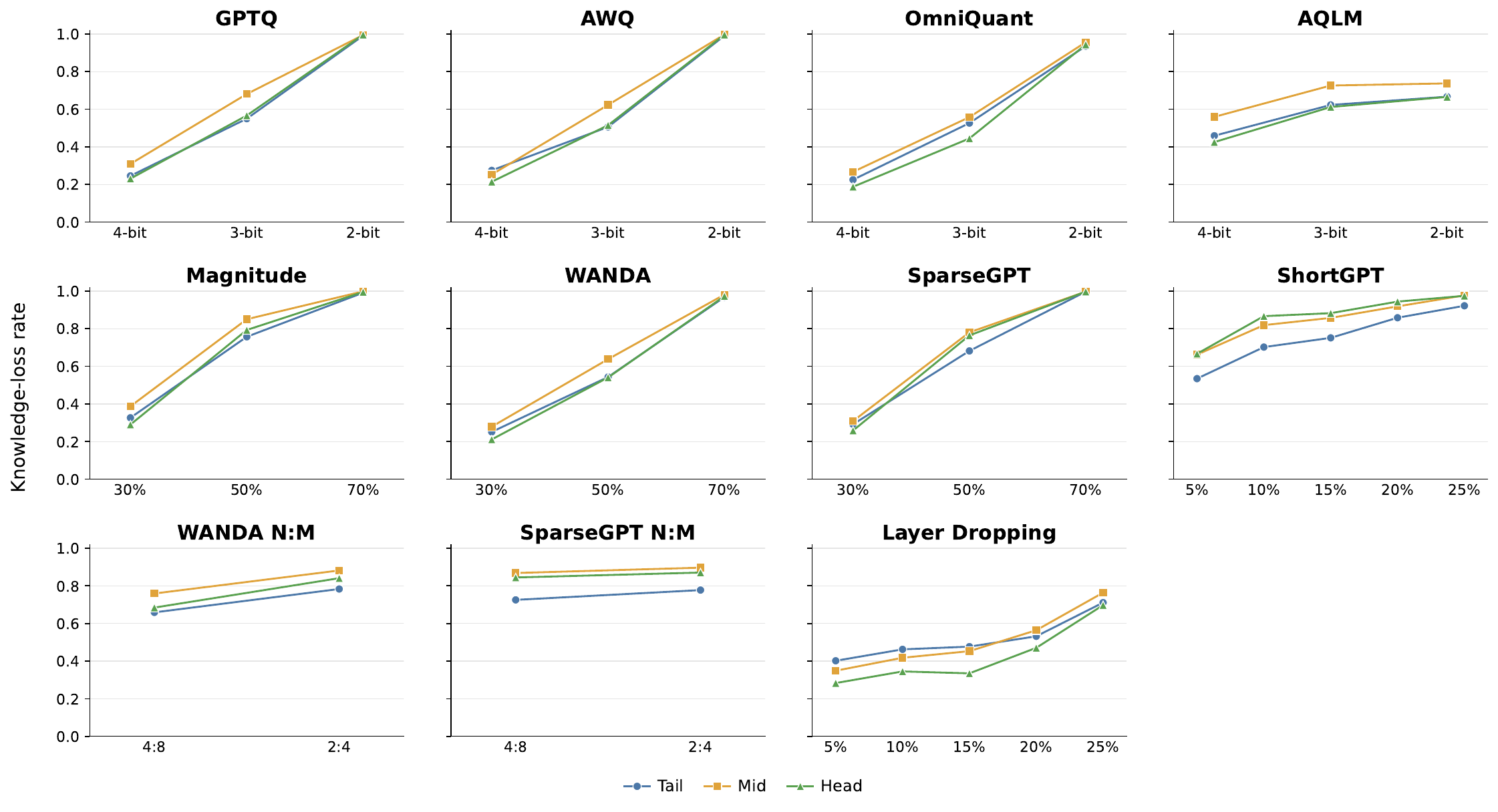}
    \caption{Knowledge-loss rates across different popularity groups on
    Head-to-Tail for \texttt{Qwen3-8B} under different compression methods
    and settings.}
    \label{fig:rq2_head_to_tail_knowledge_loss_rate_all_methods_qwen}
\end{figure*}

\begin{figure*}[t!]
    \centering
    \includegraphics[width=0.9\textwidth]
    {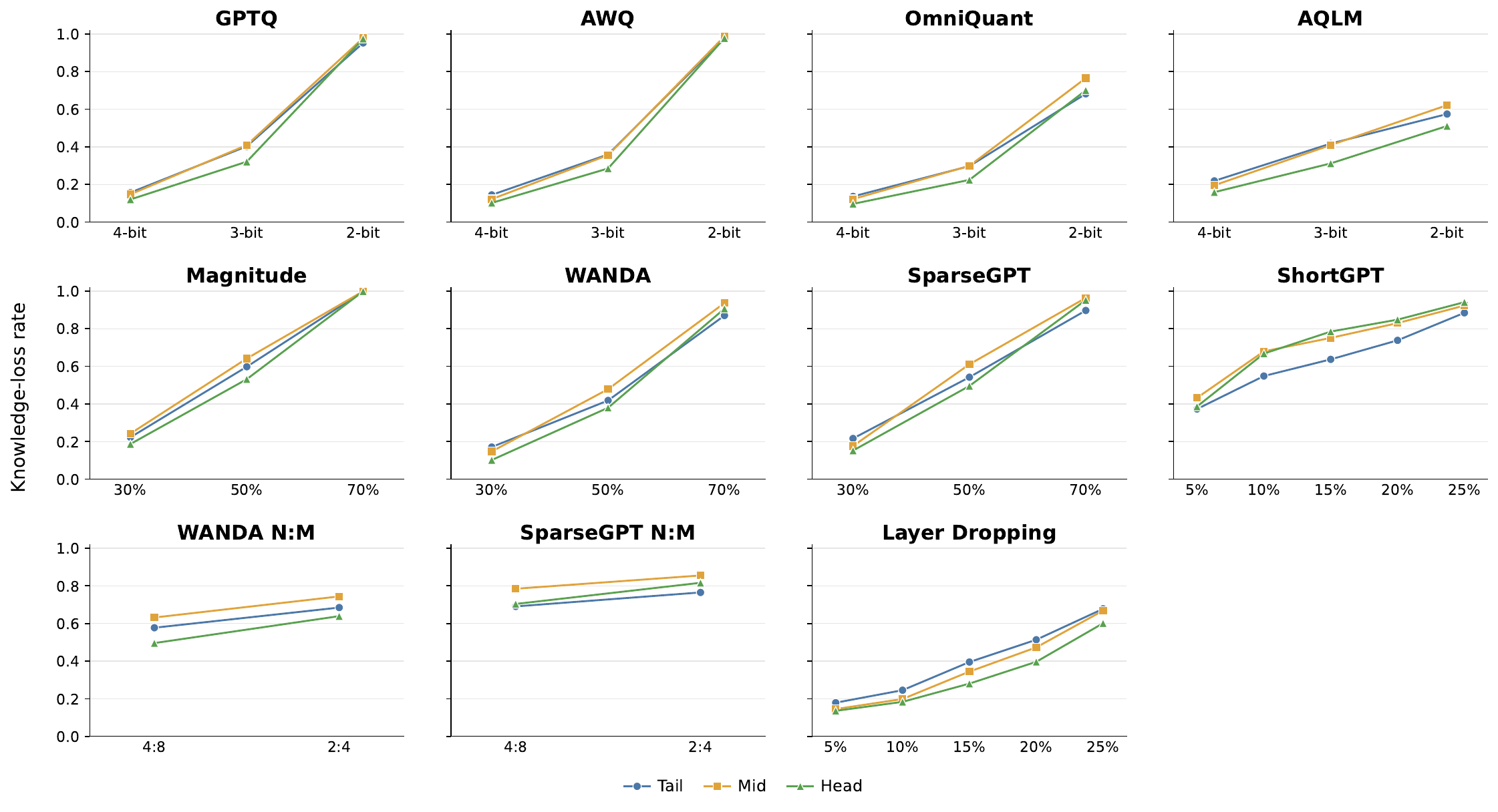}
    \caption{Knowledge-loss rates across different popularity groups on
    Head-to-Tail for \texttt{Gemma-2-9B} under different compression methods
    and settings.}
    \label{fig:rq2_head_to_tail_knowledge_loss_rate_all_methods_gemma}
\end{figure*}

\paragraph{Knowledge loss increases with compression severity.}
Across the 105 model--configuration combinations evaluated per dataset, the
least aggressive setting of each method already spans a wide range of overall
loss rates: $7.5$--$56.5\%$ on PopQA and $11.2$--$67.8\%$ on
Head-to-Tail. The most aggressive settings span $24.3$--$99.4\%$ and
$44.2$--$100.0\%$, respectively. This broad increase with compression
severity appears across all three models in figures~\ref{fig:rq2_popqa_knowledge_loss_rate_other_methods_llama},
\ref{fig:rq2_popqa_knowledge_loss_rate_all_methods_qwen}, and
\ref{fig:rq2_popqa_knowledge_loss_rate_all_methods_gemma} for PopQA, figures~\ref{fig:rq2_head_to_tail_knowledge_loss_rate_all_methods_llama},
\ref{fig:rq2_head_to_tail_knowledge_loss_rate_all_methods_qwen}, and
\ref{fig:rq2_head_to_tail_knowledge_loss_rate_all_methods_gemma} for
Head-to-Tail.

For \texttt{Llama-3.1-8B-Instruct}, PopQA loss is $22.3\%$ under
4-bit GPTQ, $22.6\%$ under WANDA at $30\%$ sparsity, and $35.0\%$
under ShortGPT at a $5\%$ pruning ratio in Figure~\ref{fig:rq2_popqa_knowledge_loss_rate_other_methods_llama}.
The corresponding Head-to-Tail rates are higher at $26.3\%$, $26.5\%$,
and $46.7\%$ in Figure~\ref{fig:rq2_head_to_tail_knowledge_loss_rate_all_methods_llama}.
At the other extreme, Llama Magnitude pruning at $70\%$ sparsity loses
$99.4\%$ of previously correct PopQA examples while retaining only
$0.23\%$ overall accuracy. Llama SparseGPT at $70\%$ sparsity loses all
previously correct Head-to-Tail examples and retains only $0.01\%$
accuracy. These configurations represent task collapse rather than useful
high-compression operating points.

\paragraph{Overconfidence persists under mild compression.}
For the Llama PopQA settings emphasized in the main text, median calibrated
confidence on lost tail, middle, and head knowledge is
$0.50/0.56/0.60$ under 4-bit GPTQ, $0.51/0.53/0.59$ under WANDA at
$30\%$ sparsity, and $0.46/0.49/0.51$ under ShortGPT at a $5\%$
pruning ratio.
Thus, even these mild settings assign moderate confidence to answers that
became incorrect.

Head-to-Tail shows the same qualitative behavior for these configurations,
with medians of $0.63/0.65/0.67$, $0.57/0.61/0.64$, and
$0.51/0.55/0.59$, respectively in figure~\ref{fig:rq2_head_to_tail_lost_confidence_all_methods_llama}.
Moderate confidence on lost knowledge also appears across multiple settings
for Qwen and Gemma, although its magnitude varies more substantially across
methods and compression levels in figures~\ref{fig:rq2_popqa_lost_confidence_other_methods_qwen}, 
\ref{fig:rq2_popqa_lost_confidence_other_methods_gemma}, ~\ref{fig:rq2_head_to_tail_lost_confidence_all_methods_qwen} and 
\ref{fig:rq2_head_to_tail_lost_confidence_all_methods_gemma}.
These values describe the confidence assigned to newly incorrect answers;
they do not show that a model detects or recognizes its own knowledge loss.

\begin{figure*}[t!]
    \centering
    \includegraphics[width=0.9\textwidth]
    {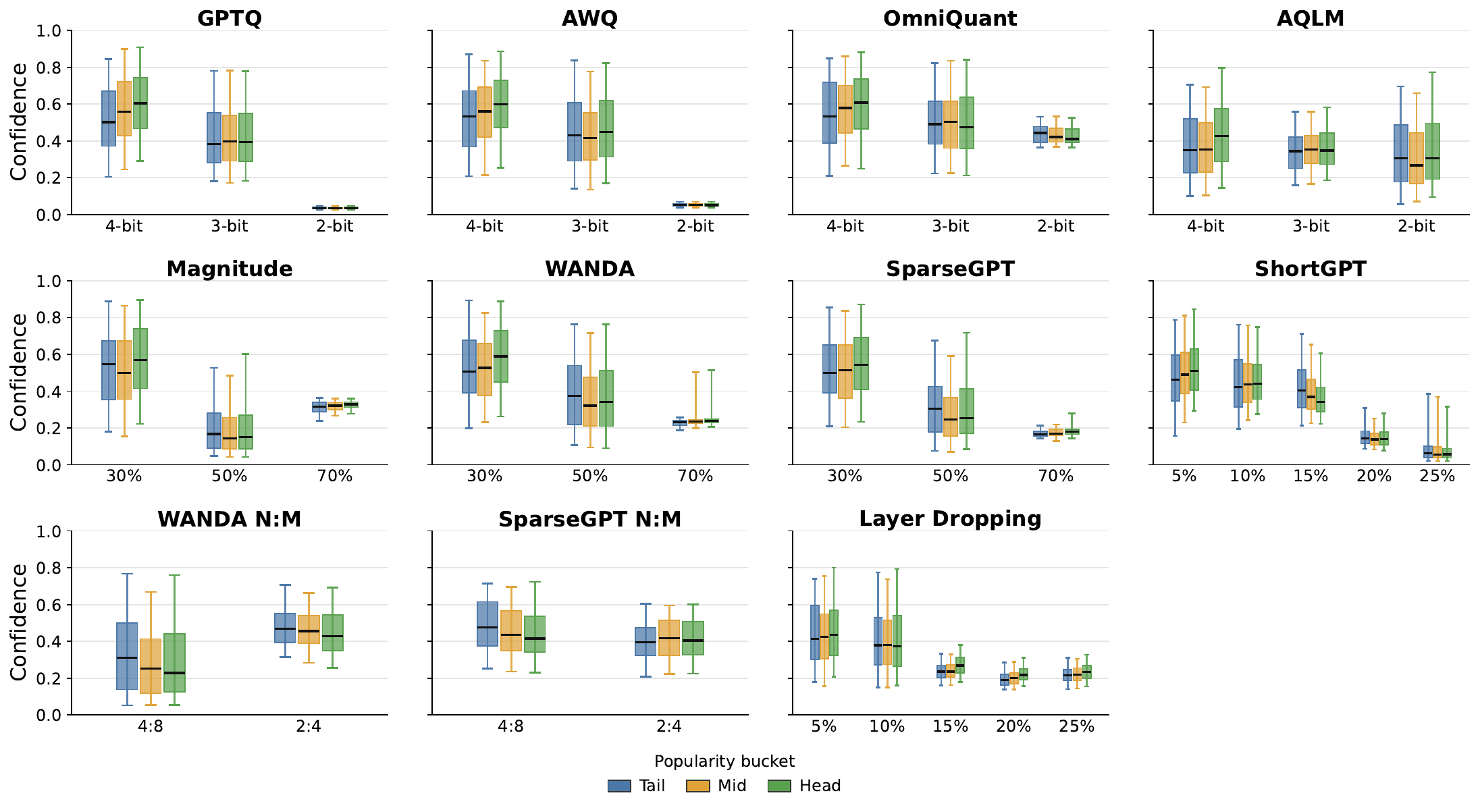}
    \caption{Calibrated confidence assigned to lost knowledge across different
    popularity groups on PopQA for \texttt{Llama-3.1-8B-Instruct} under
    additional compression methods and settings.}
    \label{fig:rq2_popqa_lost_confidence_other_methods_llama}
\end{figure*}

\begin{figure*}[t!]
    \centering
    \includegraphics[width=0.9\textwidth]
    {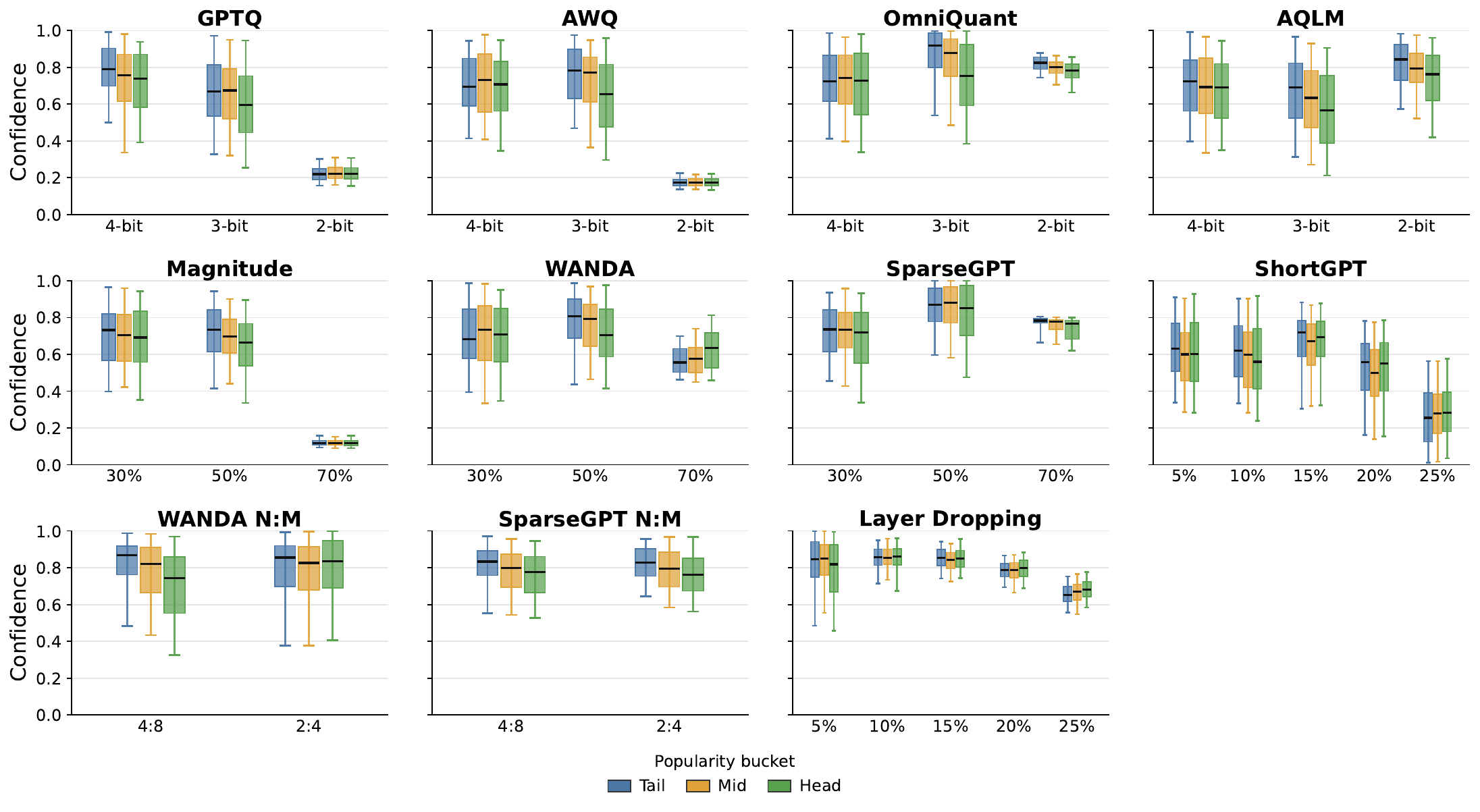}
    \caption{Calibrated confidence assigned to lost knowledge across different
    popularity groups on PopQA for \texttt{Qwen3-8B} under additional
    compression methods and settings.}
    \label{fig:rq2_popqa_lost_confidence_other_methods_qwen}
\end{figure*}

\begin{figure*}[t!]
    \centering
    \includegraphics[width=0.9\textwidth]
    {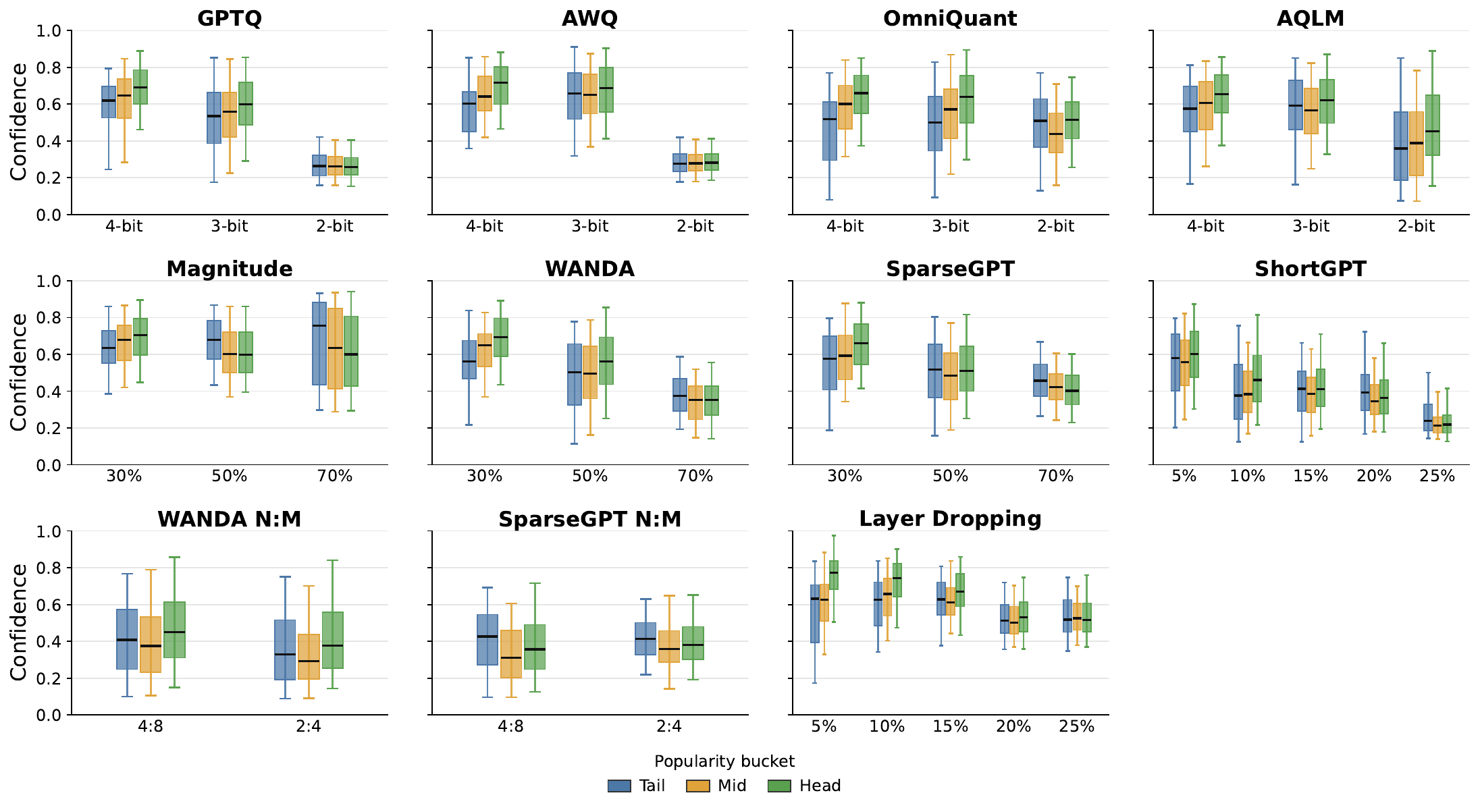}
    \caption{Calibrated confidence assigned to lost knowledge across different
    popularity groups on PopQA for \texttt{Gemma-2-9B} under additional
    compression methods and settings.}
    \label{fig:rq2_popqa_lost_confidence_other_methods_gemma}
\end{figure*}

\begin{figure*}[t!]
    \centering
    \includegraphics[width=0.9\textwidth]
    {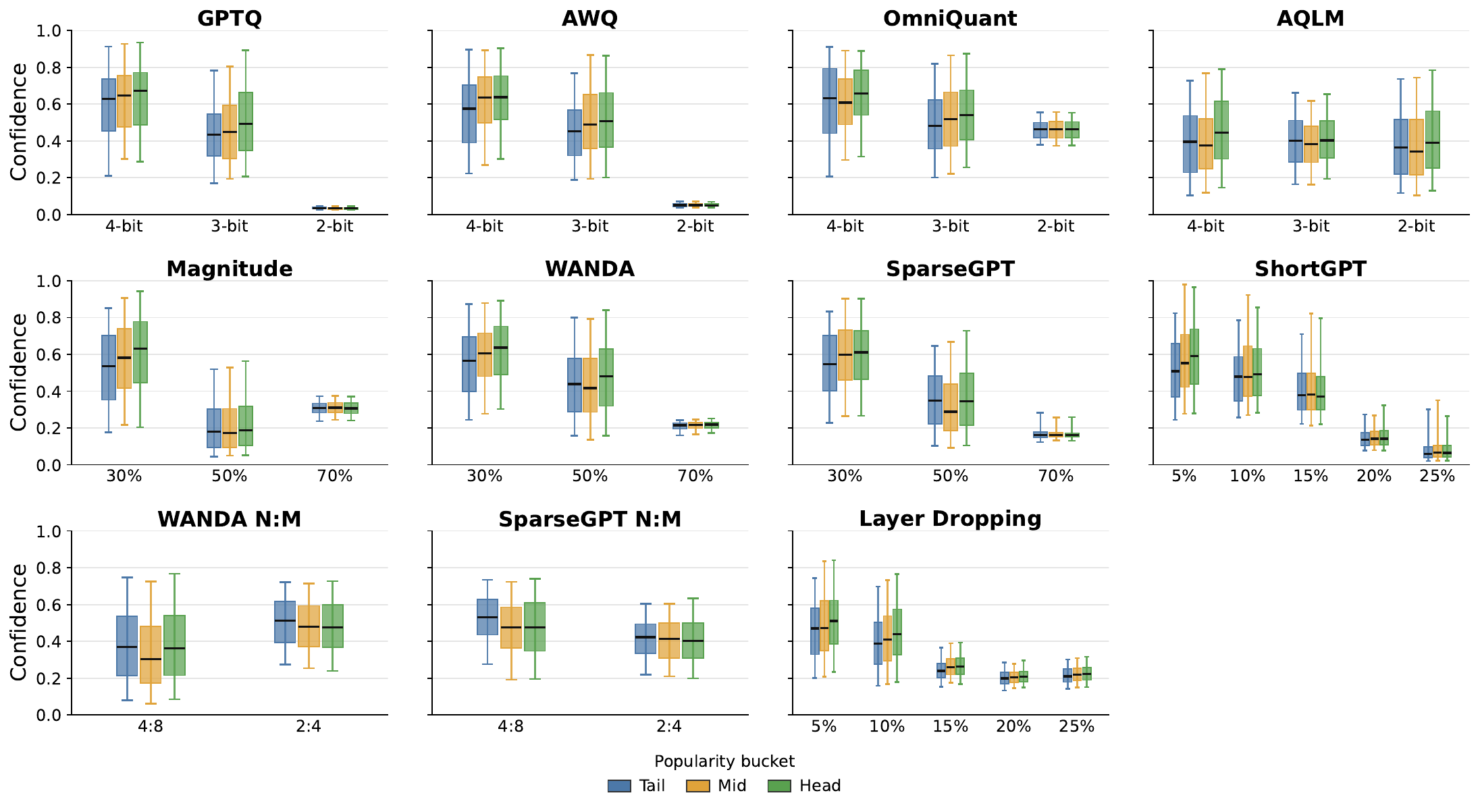}
    \caption{Calibrated confidence assigned to lost knowledge across different
    popularity groups on Head-to-Tail for
    \texttt{Llama-3.1-8B-Instruct} under all compression methods and settings.}
    \label{fig:rq2_head_to_tail_lost_confidence_all_methods_llama}
\end{figure*}

\begin{figure*}[t!]
    \centering
    \includegraphics[width=0.9\textwidth]
    {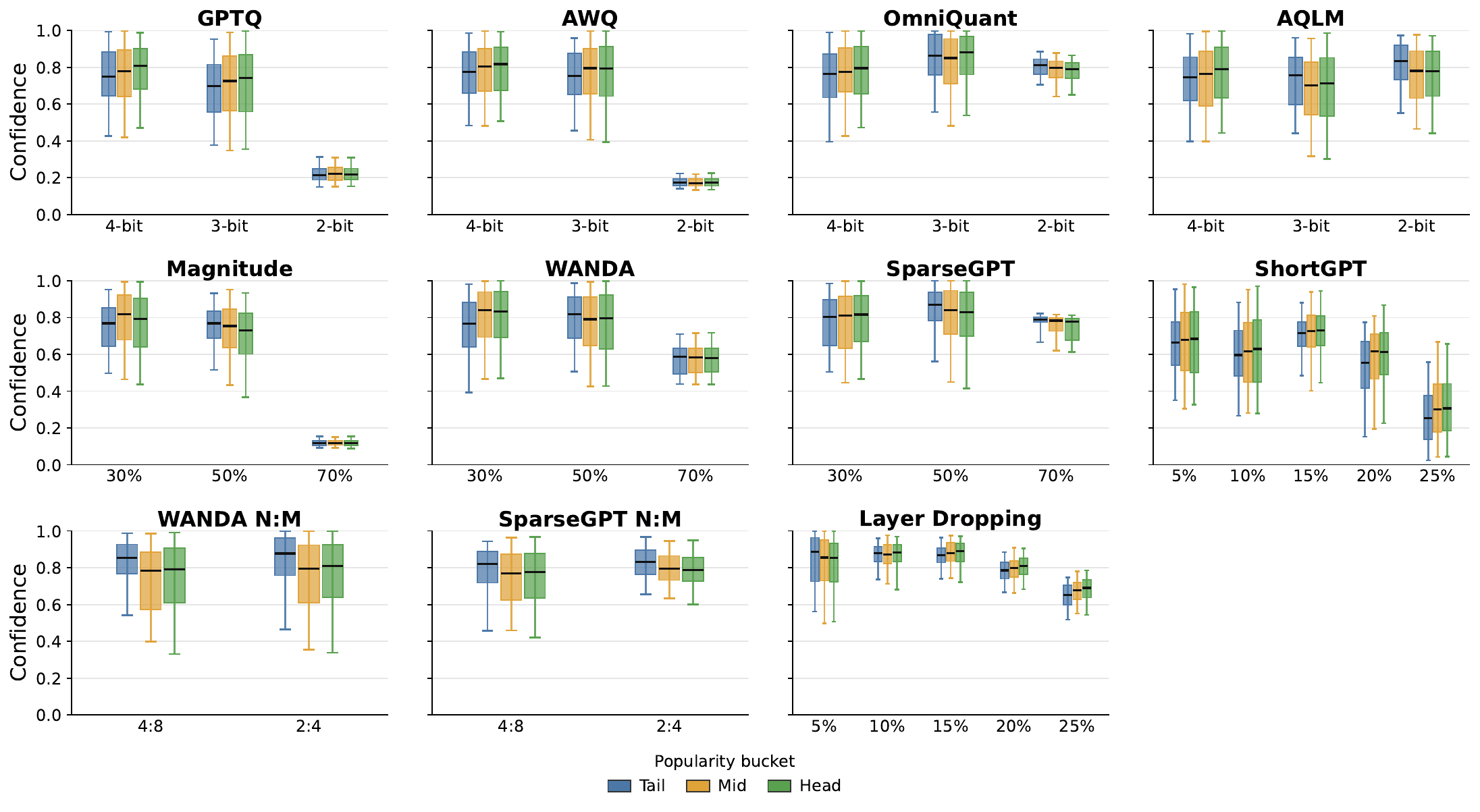}
    \caption{Calibrated confidence assigned to lost knowledge across different
    popularity groups on Head-to-Tail for \texttt{Qwen3-8B} under all
    compression methods and settings.}
    \label{fig:rq2_head_to_tail_lost_confidence_all_methods_qwen}
\end{figure*}

\begin{figure*}[t!]
    \centering
    \includegraphics[width=0.9\textwidth]
    {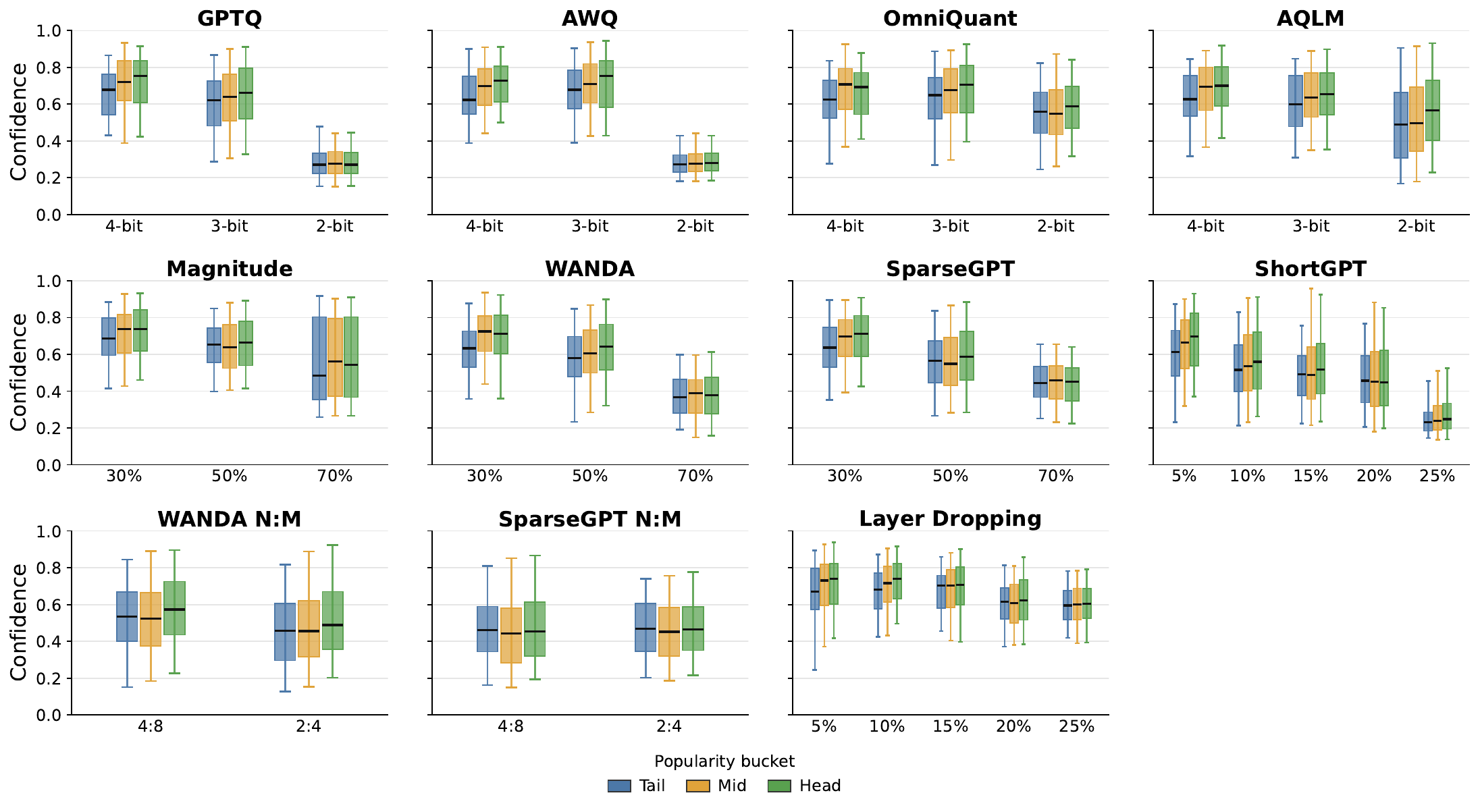}
    \caption{Calibrated confidence assigned to lost knowledge across different
    popularity groups on Head-to-Tail for \texttt{Gemma-2-9B} under all
    compression methods and settings.}
    \label{fig:rq2_head_to_tail_lost_confidence_all_methods_gemma}
\end{figure*}

\paragraph{Popularity-group differences are dataset dependent.}
The Llama examples above assign higher median confidence to lost head
knowledge than to lost tail knowledge, but this ordering is not universal.
Across all 105 model--configuration combinations, the head median exceeds
the tail median in 55 cases on PopQA. The variation across models and methods
is visible in
figures~\ref{fig:rq2_popqa_lost_confidence_other_methods_llama},
\ref{fig:rq2_popqa_lost_confidence_other_methods_qwen}, and
\ref{fig:rq2_popqa_lost_confidence_other_methods_gemma}.
On Head-to-Tail, the head median exceeds the tail median in 78 cases, indicating
a more frequent, although still non-universal, head-over-tail confidence
ordering
in figures~\ref{fig:rq2_head_to_tail_lost_confidence_all_methods_llama},
\ref{fig:rq2_head_to_tail_lost_confidence_all_methods_qwen}, and
\ref{fig:rq2_head_to_tail_lost_confidence_all_methods_gemma}.

The ordering of loss rates differs more sharply across datasets. Head loss
exceeds tail loss in 93 of 105 PopQA combinations
(Figures~\ref{fig:rq2_popqa_knowledge_loss_rate_other_methods_llama},
\ref{fig:rq2_popqa_knowledge_loss_rate_all_methods_qwen}, and
\ref{fig:rq2_popqa_knowledge_loss_rate_all_methods_gemma}),
but in only 48 of 105 Head-to-Tail combinations
(Figures~\ref{fig:rq2_head_to_tail_knowledge_loss_rate_all_methods_llama},
\ref{fig:rq2_head_to_tail_knowledge_loss_rate_all_methods_qwen}, and
\ref{fig:rq2_head_to_tail_knowledge_loss_rate_all_methods_gemma}).
These results reveal popularity-stratified differences in both which
knowledge is lost and the confidence assigned to that loss, but neither
ordering generalizes uniformly across models, methods, and datasets.

\paragraph{Compression trajectories and calibration are method specific.}
Many configurations exhibit increasing knowledge loss as compression becomes
more severe, but their confidence trajectories remain method and model
specific. Figures~\ref{fig:rq2_popqa_knowledge_loss_rate_other_methods_llama}
shows that moving GPTQ from 4 to
3 to 2 bits for Llama raises PopQA loss from $22.3\%$ to $53.0\%$ to
$97.3\%$ and lowers the tail median confidence from $0.50$ to $0.38$
to $0.03$. WANDA at $30\%$, $50\%$, and $70\%$ sparsity similarly
raises loss from $22.6\%$ to $54.7\%$ to $98.7\%$.

AQLM is not monotonic in bit width for Llama.
Figures~\ref{fig:rq2_popqa_knowledge_loss_rate_other_methods_llama}
and~\ref{fig:rq2_head_to_tail_knowledge_loss_rate_all_methods_llama}
show PopQA loss rates of $56.5\%$, $73.1\%$, and $59.2\%$ at 4, 3,
and 2 bits, respectively, compared with corresponding Head-to-Tail
rates of $67.8\%$, $90.1\%$, and $73.3\%$. ShortGPT can also produce
non-monotonic confidence changes. As shown in
Figures~\ref{fig:rq2_popqa_lost_confidence_other_methods_qwen}
and~\ref{fig:rq2_popqa_knowledge_loss_rate_all_methods_qwen}, the tail
median confidence of \texttt{Qwen3-8B} on PopQA rises from $0.62$ at a
$10\%$ pruning ratio to $0.72$ at $15\%$, even as loss increases from
$58.3\%$ to $61.2\%$.

Importance-aware layer dropping is particularly model specific.
Figures~\ref{fig:rq2_popqa_knowledge_loss_rate_all_methods_qwen}
and~\ref{fig:rq2_popqa_knowledge_loss_rate_other_methods_llama} show
that, through a $20\%$ dropping ratio, Qwen PopQA loss remains between
$20.6\%$ and $24.3\%$, whereas the corresponding Llama range is
$42.7\%$--$78.9\%$. Semi-structured WANDA also differs from its
unstructured counterpart. Figure~
\ref{fig:rq2_popqa_knowledge_loss_rate_other_methods_llama} shows that
WANDA 4:8 and 2:4 lose $64.7\%$ and $71.5\%$ of previously correct
Llama PopQA examples, respectively, compared with $54.7\%$ at $50\%$
unstructured sparsity.

Severe compression frequently lowers confidence only after most previously
correct knowledge and nearly all task accuracy have disappeared.
Figures
\ref{fig:rq2_popqa_knowledge_loss_rate_other_methods_llama},
\ref{fig:rq2_head_to_tail_lost_confidence_all_methods_llama}, and
\ref{fig:rq2_head_to_tail_knowledge_loss_rate_all_methods_llama} show
that Llama 2-bit GPTQ has tail and head median confidence near $0.03$
on both datasets, while its loss rate reaches $97.3\%$ on PopQA and
$99.8\%$ on Head-to-Tail. The corresponding accuracies are only
$1.05\%$ and $0.05\%$. Likewise, Llama WANDA at $70\%$ sparsity
reaches $98.7\%$ PopQA loss and $100.0\%$ Head-to-Tail loss while
accuracy falls to $0.50\%$ and $0.03\%$. Lower confidence under these
settings therefore accompanies task collapse rather than improved
practical reliability.

\begin{figure*}[t!]
    \centering
    \includegraphics[width=0.9\textwidth]
    {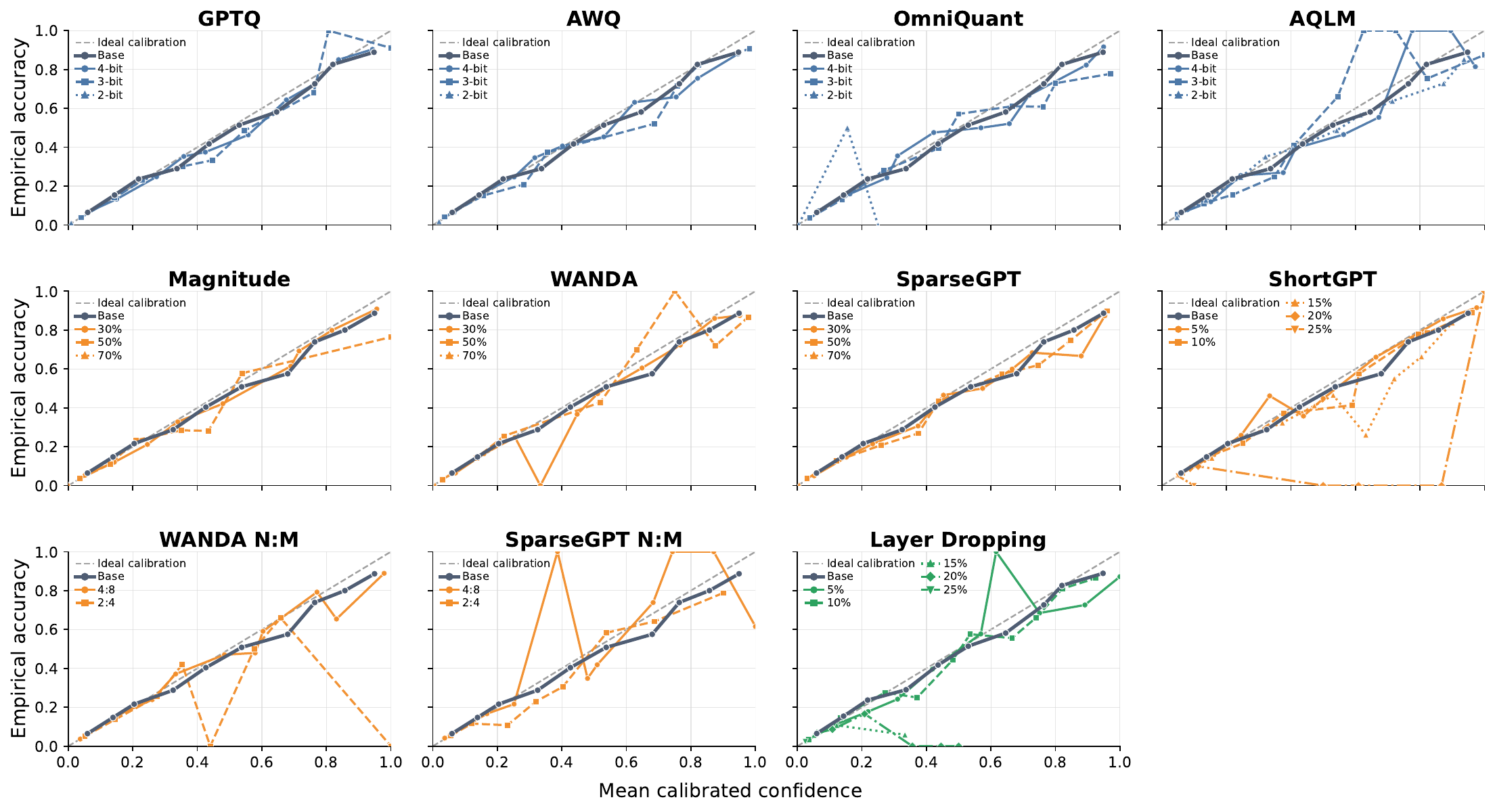}
    \caption{Calibrated reliability diagrams on PopQA for
    \texttt{Llama-3.1-8B-Instruct} across all compression methods and settings.
    The diagonal line represents perfect calibration.}
    \label{fig:rq2_popqa_calibrated_reliability_all_methods_llama}
\end{figure*}

\begin{figure*}[t!]
    \centering
    \includegraphics[width=0.9\textwidth]
    {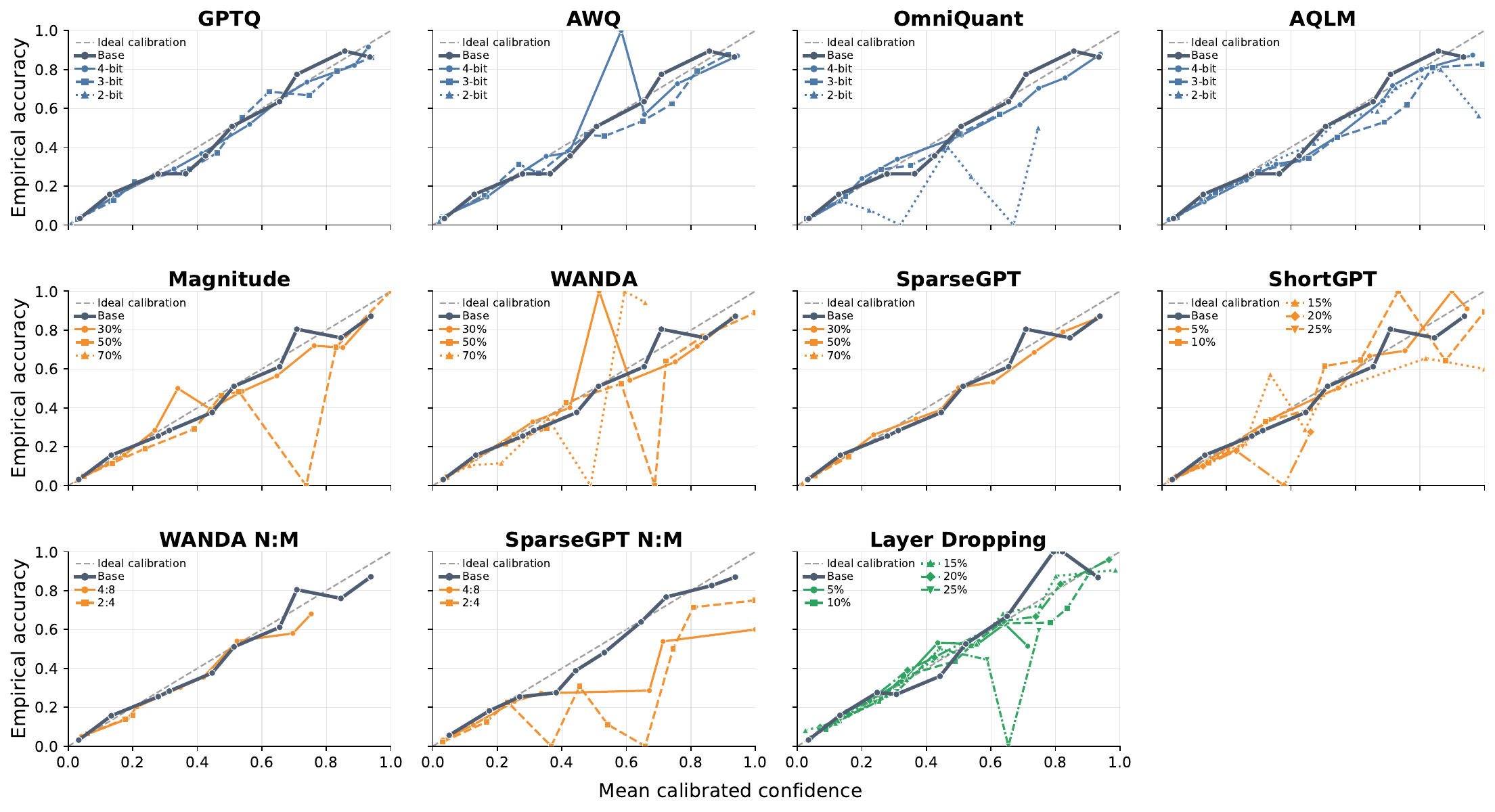}
    \caption{Calibrated reliability diagrams on PopQA for
    \texttt{Qwen3-8B} across all compression methods and settings.
    The diagonal line represents perfect calibration.}
    \label{fig:rq2_popqa_calibrated_reliability_all_methods_qwen}
\end{figure*}

\begin{figure*}[t!]
    \centering
    \includegraphics[width=0.9\textwidth]
    {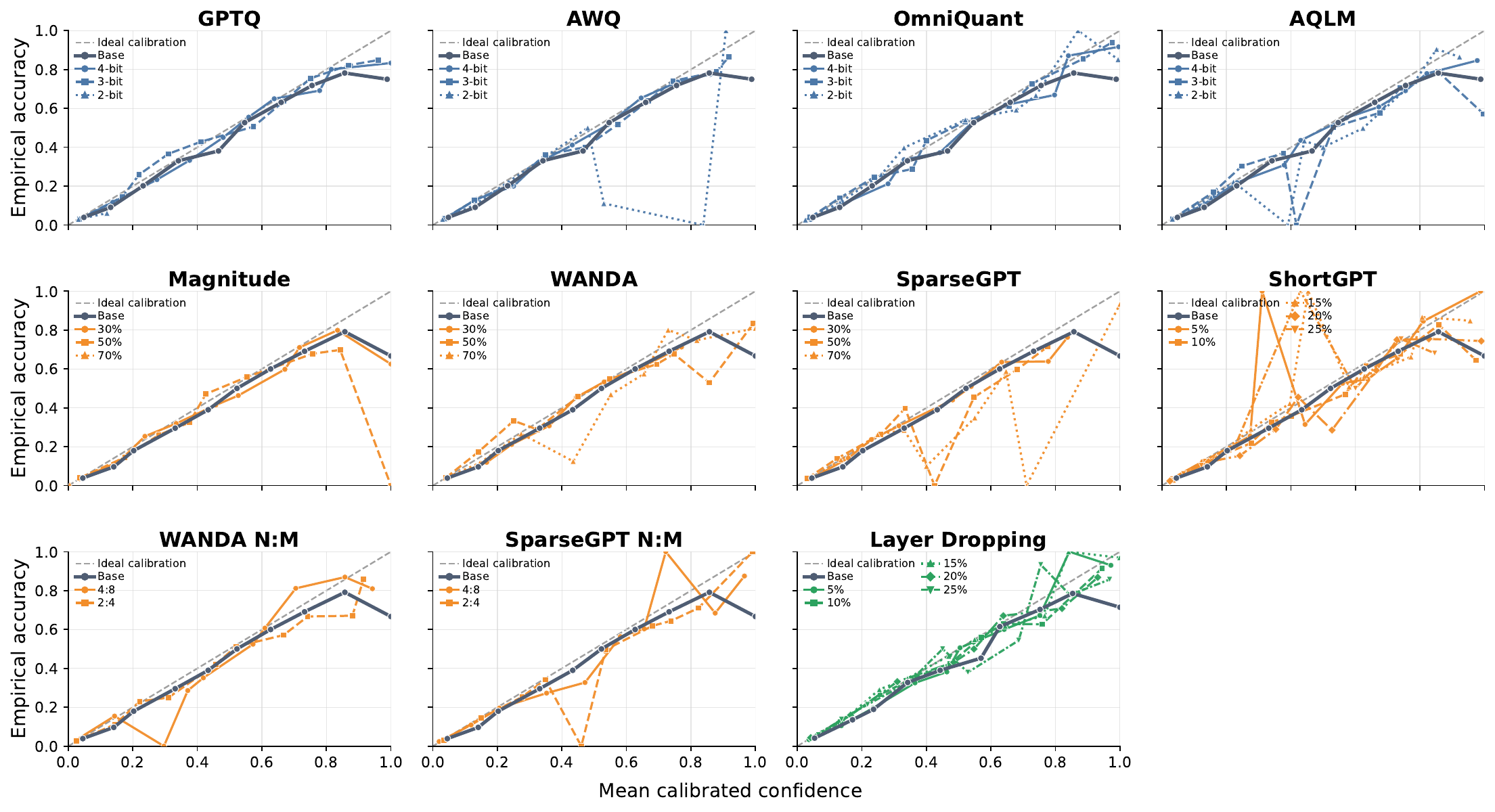}
    \caption{Calibrated reliability diagrams on PopQA for
    \texttt{Gemma-2-9B} across all compression methods and settings.
    The diagonal line represents perfect calibration.}
    \label{fig:rq2_popqa_calibrated_reliability_all_methods_gemma}
\end{figure*}

\begin{figure*}[t!]
    \centering
    \includegraphics[width=0.9\textwidth]
    {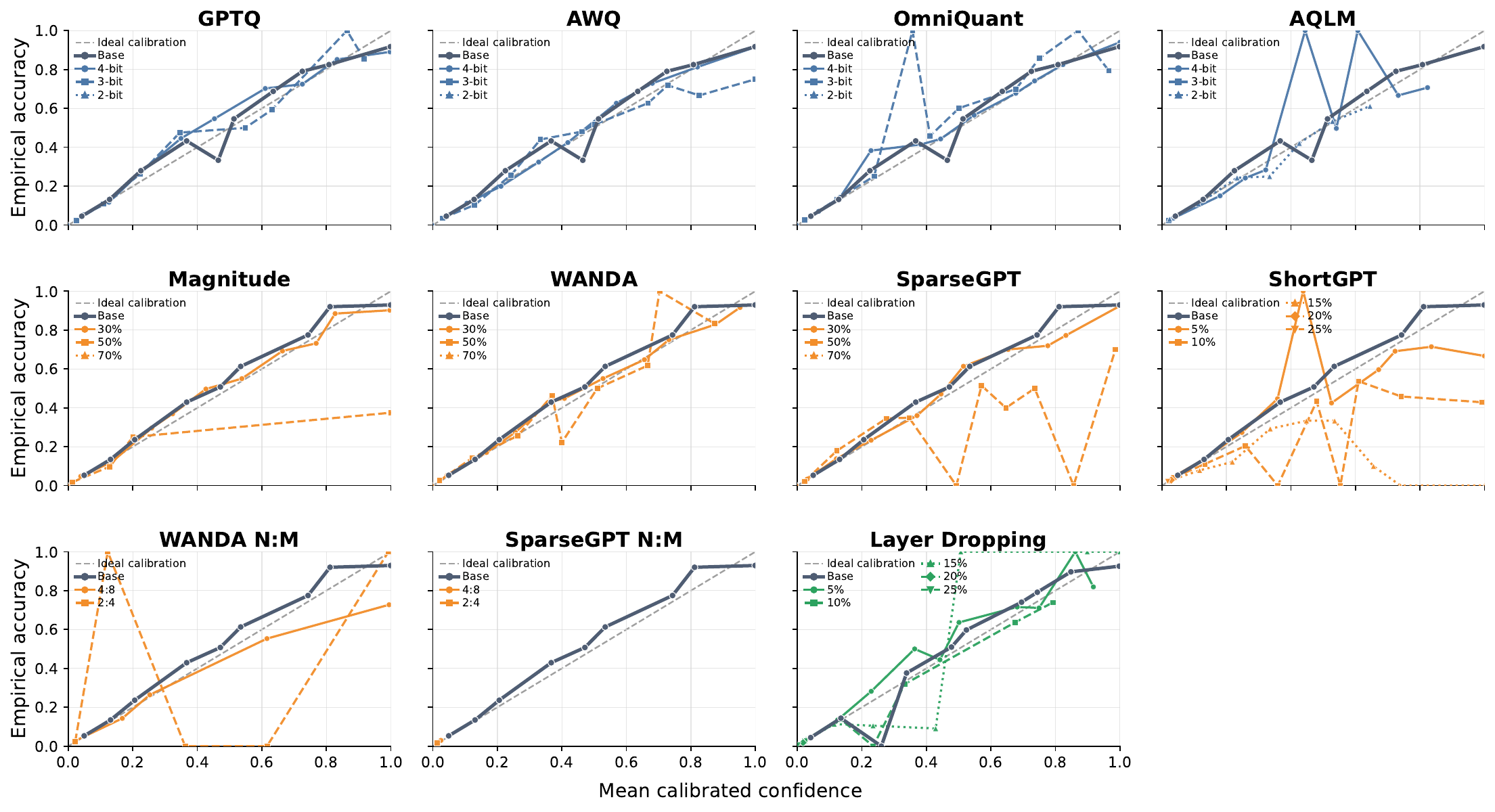}
    \caption{Calibrated reliability diagrams on Head-to-Tail for
    \texttt{Llama-3.1-8B-Instruct} across all compression methods and settings.
    The diagonal line represents perfect calibration.}
    \label{fig:rq2_head_to_tail_calibrated_reliability_all_methods_llama}
\end{figure*}

\begin{figure*}[t!]
    \centering
    \includegraphics[width=0.9\textwidth]
    {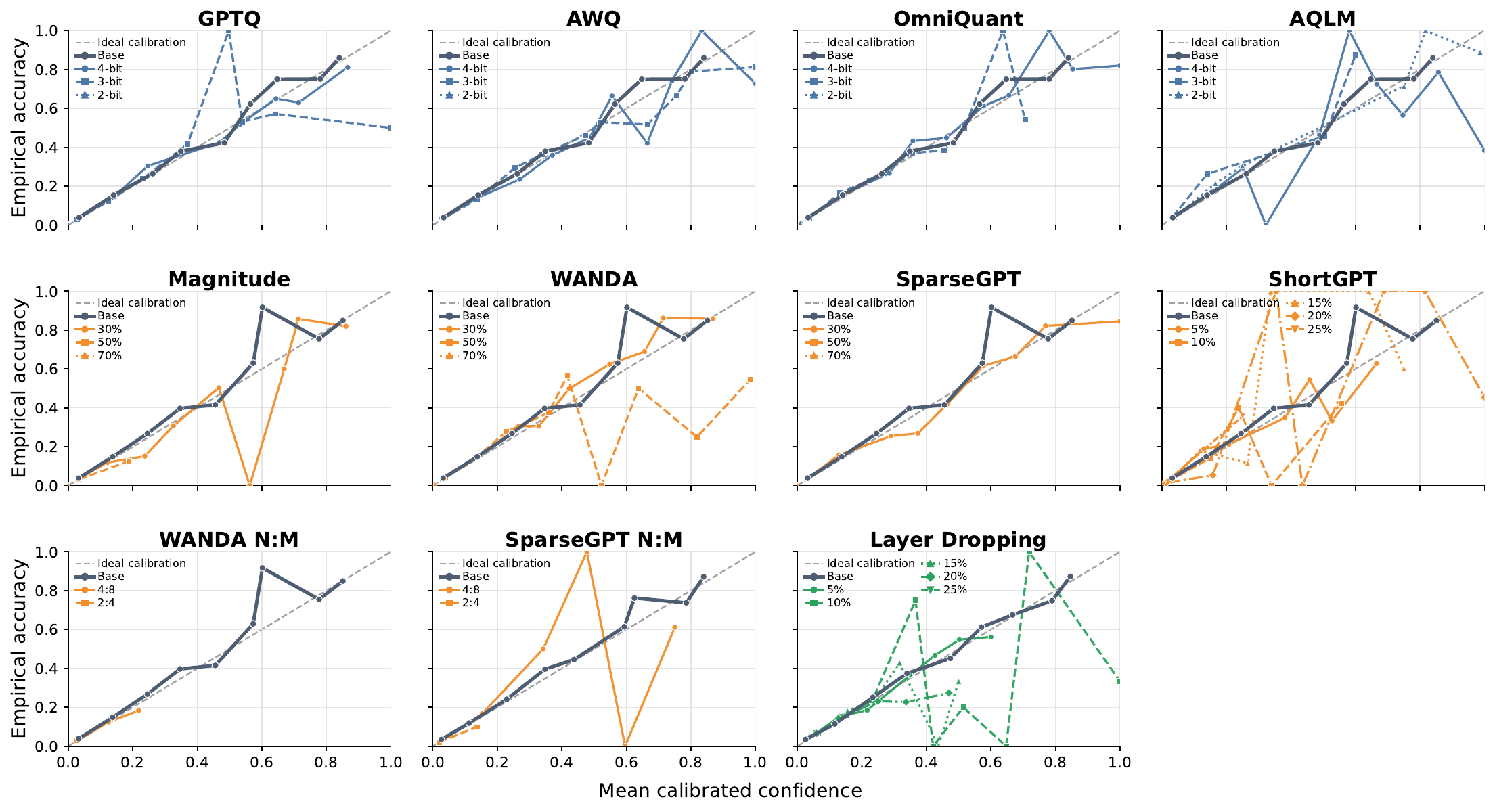}
    \caption{Calibrated reliability diagrams on Head-to-Tail for
    \texttt{Qwen3-8B} across all compression methods and settings.
    The diagonal line represents perfect calibration.}
    \label{fig:rq2_head_to_tail_calibrated_reliability_all_methods_qwen}
\end{figure*}

\begin{figure*}[t!]
    \centering
    \includegraphics[width=0.9\textwidth]
    {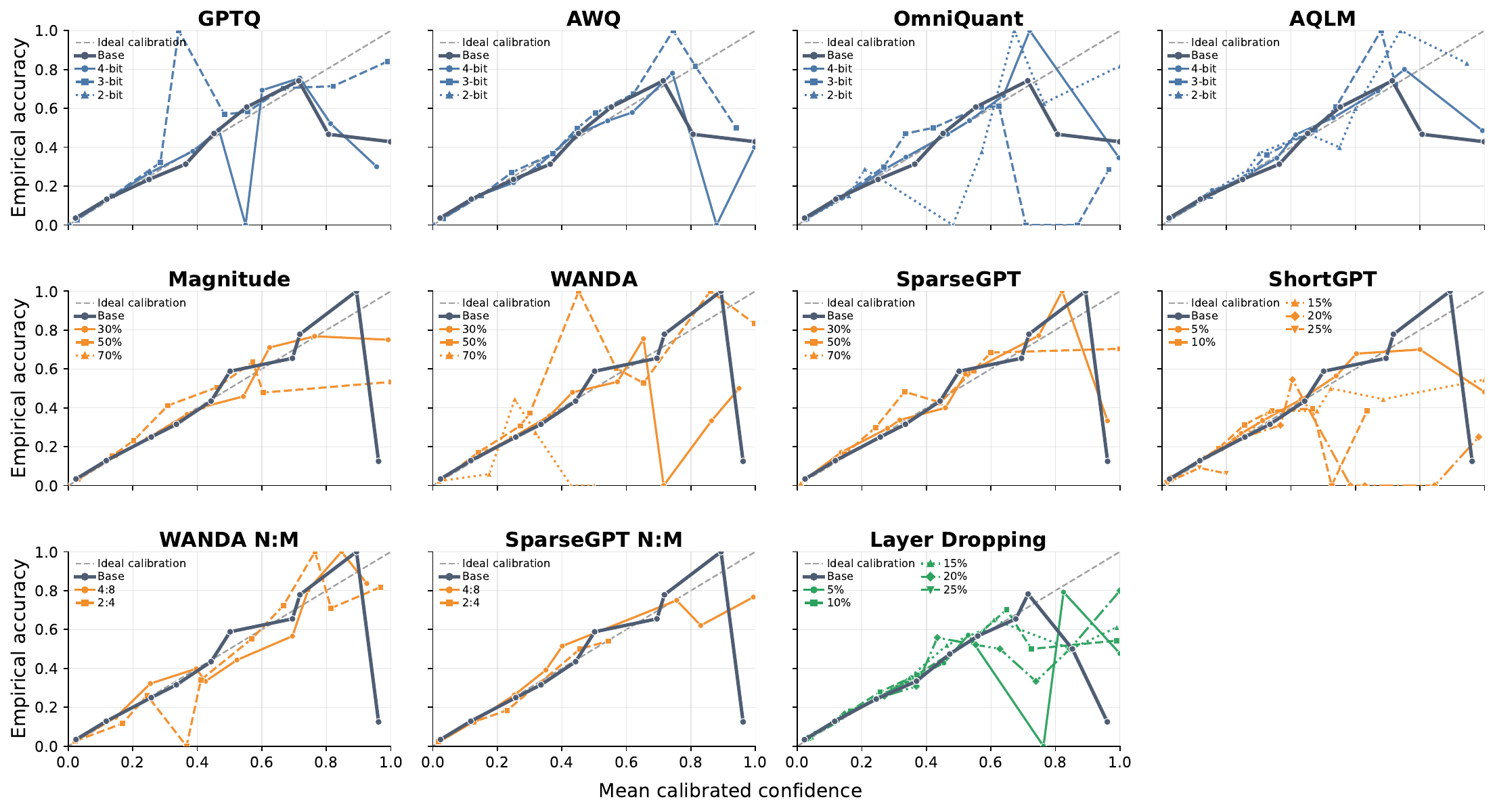}
    \caption{Calibrated reliability diagrams on Head-to-Tail for
    \texttt{Gemma-2-9B} across all compression methods and settings.
    The diagonal line represents perfect calibration.}
    \label{fig:rq2_head_to_tail_calibrated_reliability_all_methods_gemma}
\end{figure*}

A similar qualification applies to ECE. We define
$\Delta\mathrm{ECE}$ as the calibrated ECE of the compressed model minus
that of its paired base model, so a positive value denotes larger ECE and
a negative value denotes smaller ECE. Overall $\Delta\mathrm{ECE}$ ranges
from $-2.52$ to $+0.43$ percentage points on PopQA and from $-1.79$ to
$+0.54$ points on Head-to-Tail. It is negative for 100 of 105 PopQA
combinations and 95 of 105 Head-to-Tail combinations. The corresponding
PopQA reliability results are shown in
Figures~\ref{fig:rq2_popqa_calibrated_reliability_all_methods_llama},
\ref{fig:rq2_popqa_calibrated_reliability_all_methods_qwen}, and
\ref{fig:rq2_popqa_calibrated_reliability_all_methods_gemma}. The
Head-to-Tail reliability results are shown in
Figures~\ref{fig:rq2_head_to_tail_calibrated_reliability_all_methods_llama},
\ref{fig:rq2_head_to_tail_calibrated_reliability_all_methods_qwen}, and
\ref{fig:rq2_head_to_tail_calibrated_reliability_all_methods_gemma}.

Although the predominance of negative values indicates lower ECE on the test
split, it does not by itself establish improved practical reliability or
better knowledge preservation, because some of the largest reductions occur
when correctness and confidence both collapse toward zero. Group-level
changes are also less uniform, ranging from $-7.74$ to $+3.54$ percentage
points on PopQA and from $-3.89$ to $+2.29$ points on Head-to-Tail. No
single popularity group consistently exhibits the largest change across
models, methods, and datasets. Confidence and ECE must therefore be
interpreted jointly with knowledge-loss rate and task accuracy.

\paragraph{RQ2 takeaway.}
Under mild compression, models can remain moderately confident on knowledge
that becomes incorrect, but the magnitude and popularity ordering of this
confidence are dataset and model dependent. Stronger compression generally
increases knowledge loss, while the decline in confidence and ECE under the
most severe settings frequently coincides with near-total task collapse.
Assessing compression-induced reliability therefore requires jointly
considering knowledge-loss rate, confidence on lost knowledge, calibrated
reliability, ECE, and task accuracy rather than interpreting any one statistic
in isolation.

\subsubsection{Additional RQ3 Results}
\label{app:rq3_additional_results}
\begin{figure*}[t!]
    \centering
    \includegraphics[width=0.9\textwidth]
    {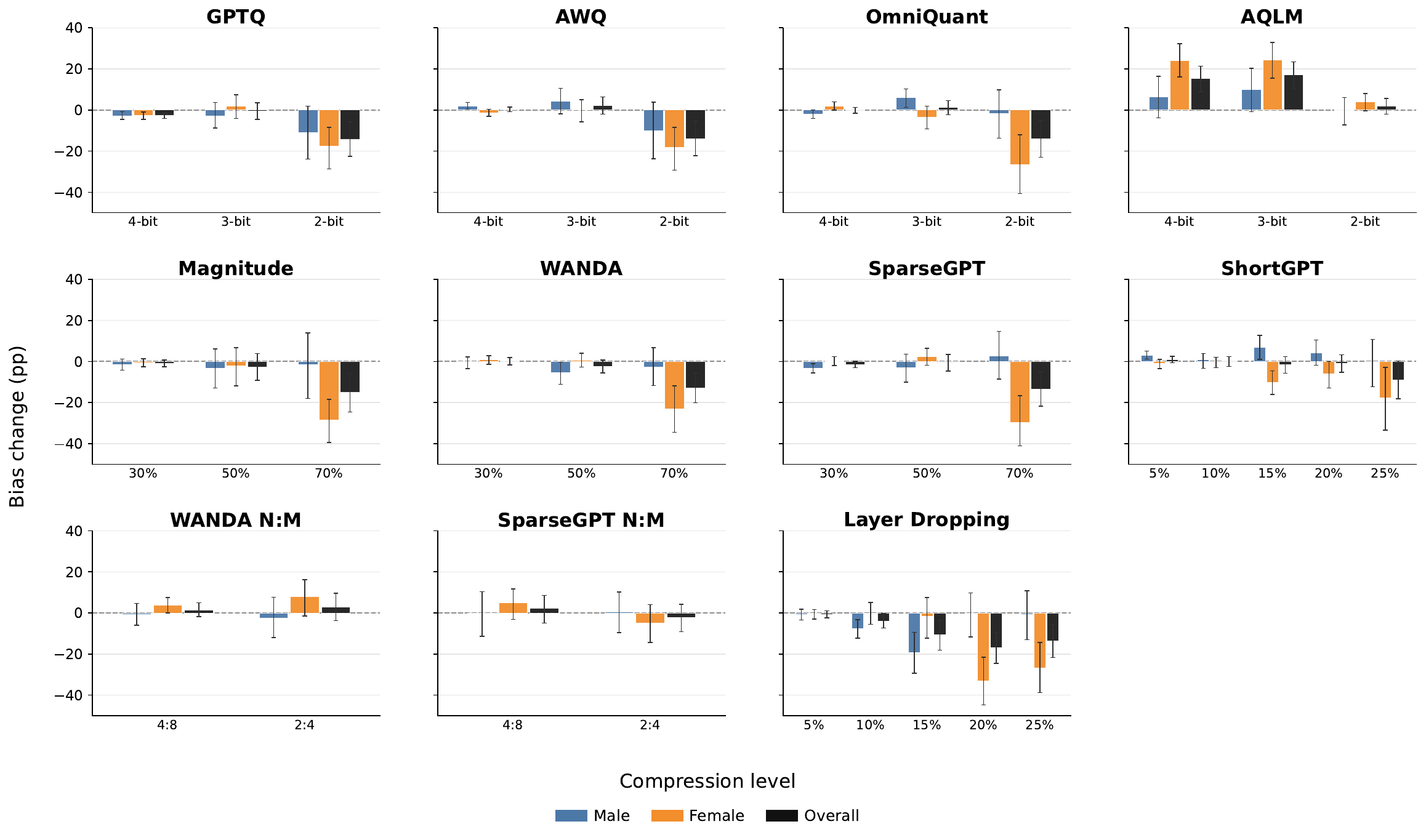}
    \caption{Overall and gender-subgroup changes in stereotypical preference on
    WinoBias for \texttt{Llama-3.1-8B-Instruct} under additional compression
    methods and settings. Blue and orange bars show the changes for male and
    female subgroups, respectively, while black bars show the overall change. Error bars denote 95\%
confidence intervals.
    Positive values indicate increased stereotypical preference after
    compression. The dashed
    vertical line denotes no change from the base model. All values are
    reported in percentage points.}
    \label{fig:rq3_winobias_llama_subgroup_bias_change_other_methods}
\end{figure*}

\begin{figure*}[t!]
    \centering
    \includegraphics[width=0.9\textwidth]
    {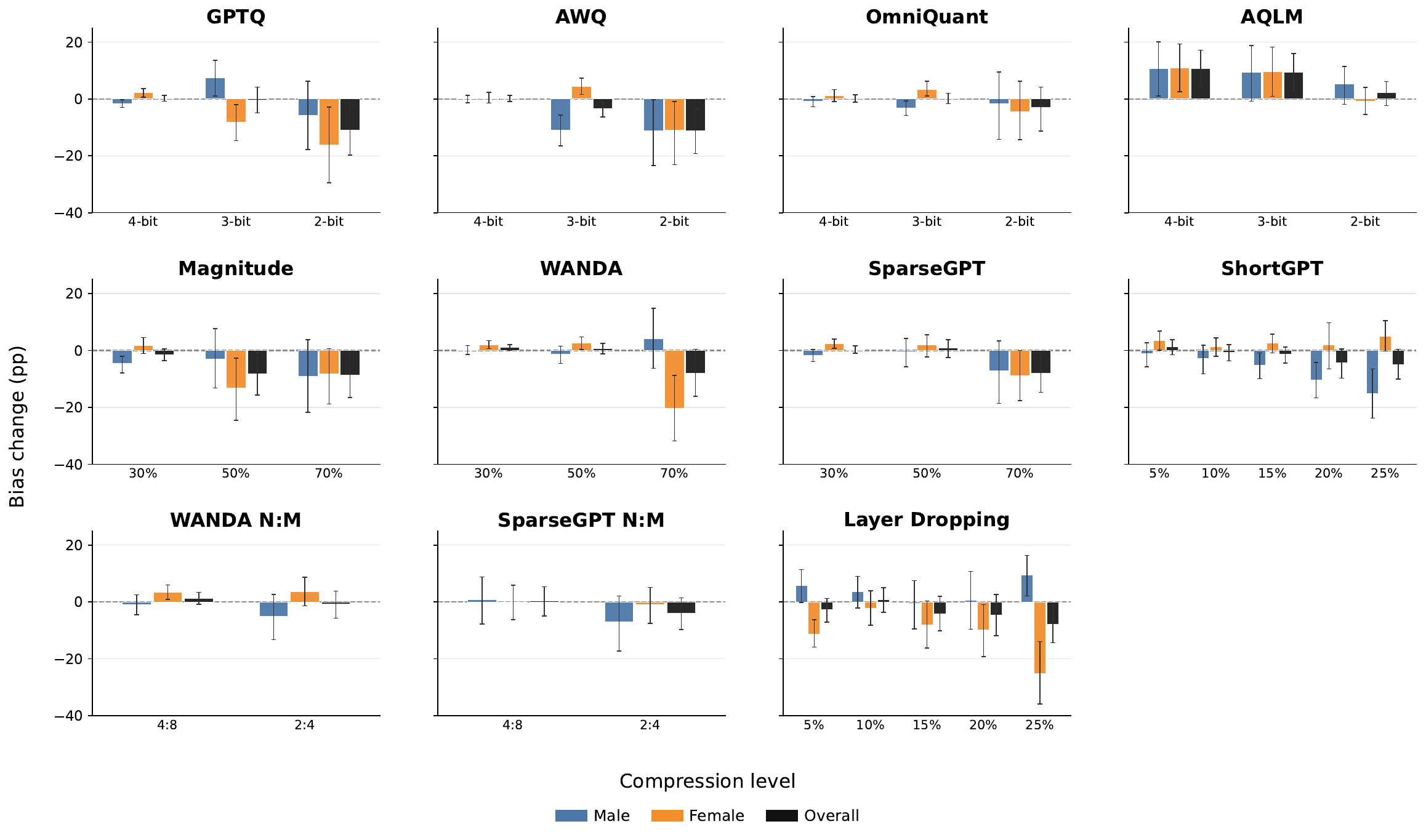}
    \caption{Overall and gender-subgroup changes in stereotypical preference on
    WinoBias for \texttt{Qwen3-8B} under different compression methods and
    settings. Blue and orange bars show the changes for male and female
    subgroups, respectively, while black bars show the overall change. Error bars denote 95\%
confidence intervals.
    Positive values indicate increased stereotypical preference after
    compression. The dashed
    vertical line denotes no change from the base model. All values are
    reported in percentage points.}
    \label{fig:rq3_winobias_qwen_subgroup_bias_change}
\end{figure*}

\begin{figure*}[t!]
    \centering
    \includegraphics[width=0.9\textwidth]
    {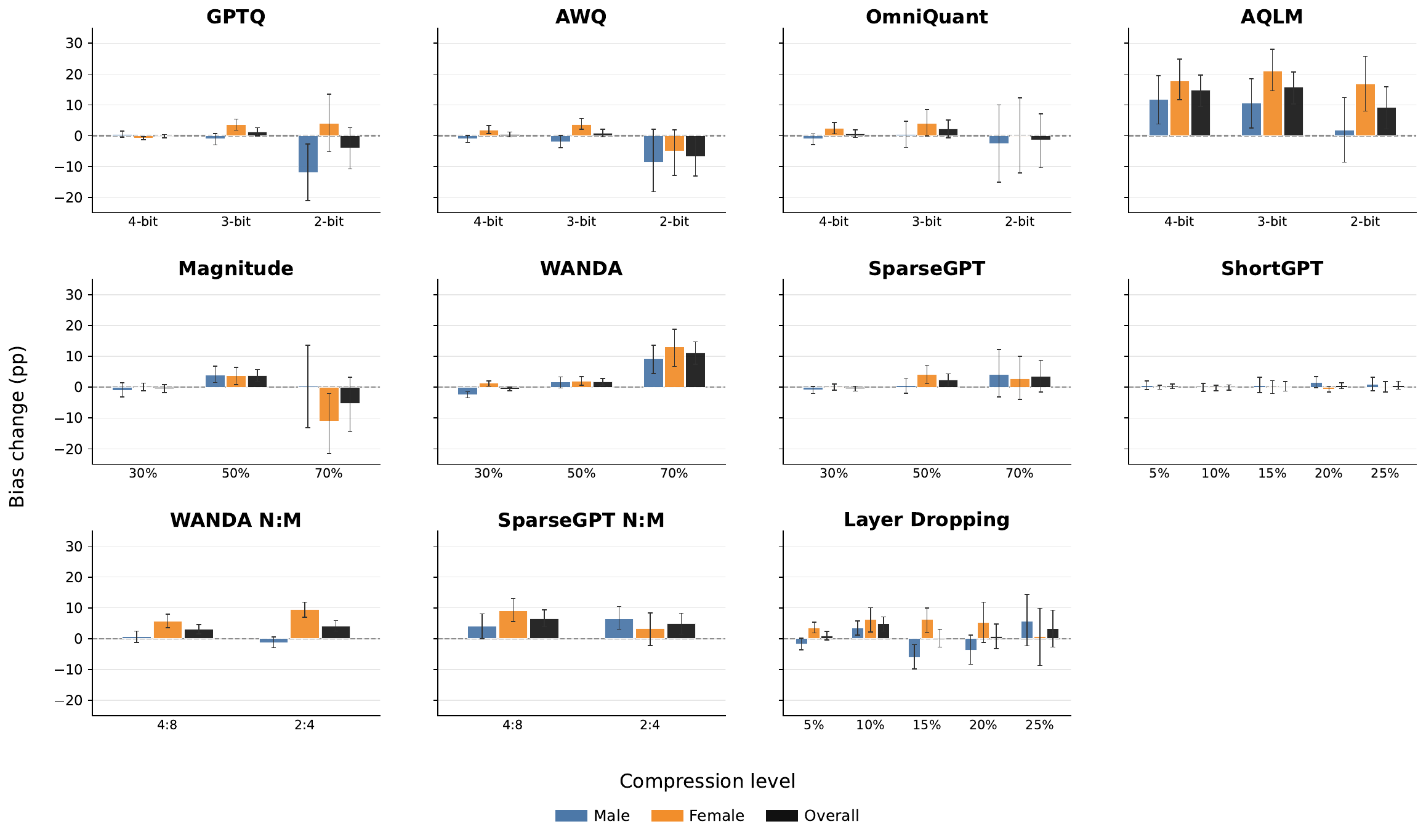}
    \caption{Overall and gender-subgroup changes in stereotypical preference on
    WinoBias for \texttt{Gemma-2-9B-it} under different compression methods and
    settings. Blue and orange bars show the changes for male and female
    subgroups, respectively, while black bars show the overall change. Error bars denote 95\%
confidence intervals.
    Positive values indicate increased stereotypical preference after
    compression. The dashed
    vertical line denotes no change from the base model. All values are
    reported in percentage points.}
    \label{fig:rq3_winobias_gemma_subgroup_bias_change}
\end{figure*}

\begin{figure*}[t!]
    \centering
    \includegraphics[width=0.9\textwidth]
    {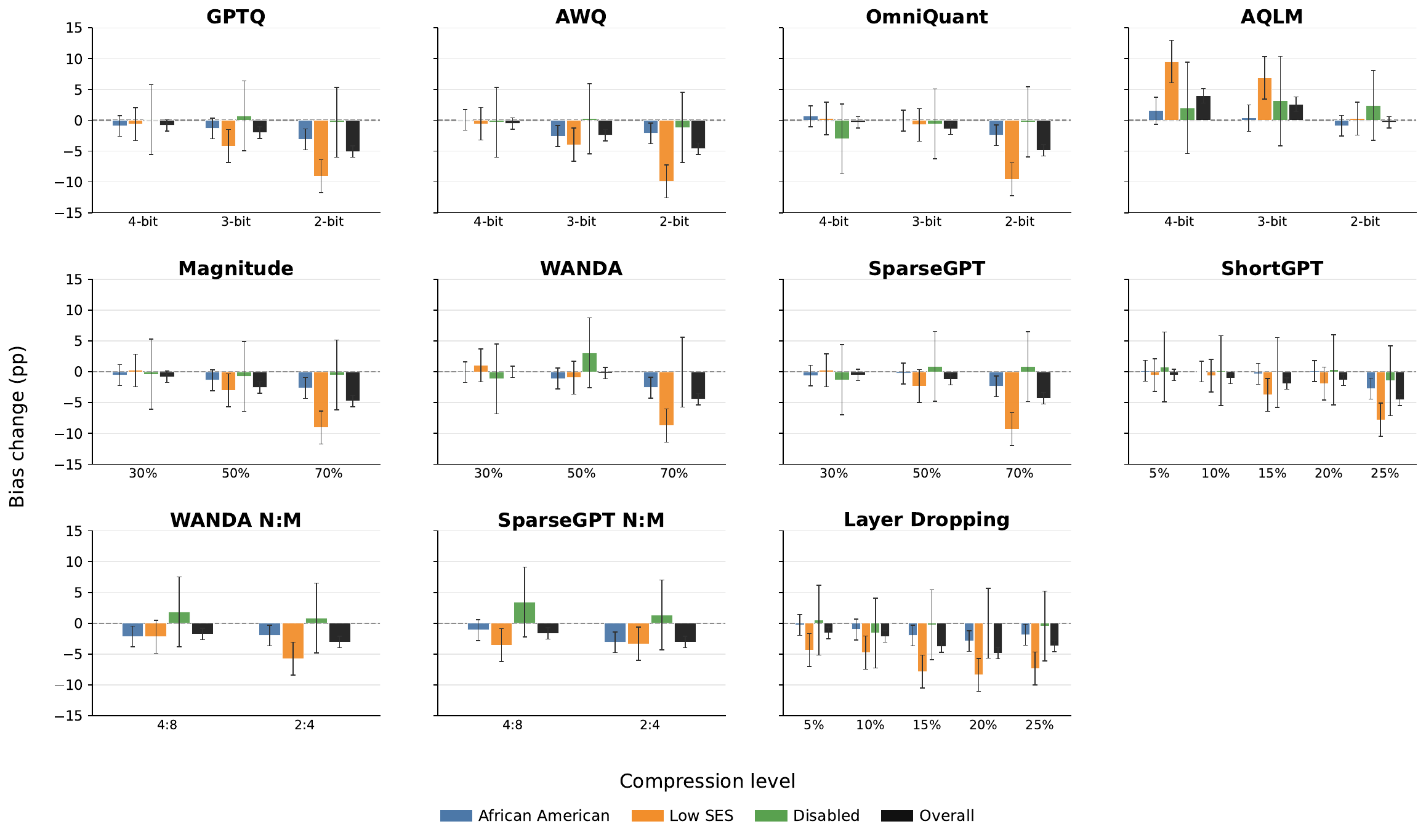}
    \caption{Overall and subgroup-level changes in stereotypical preference on
    BBQ for \texttt{Llama-3.1-8B-Instruct} under different compression methods
    and settings. Colored bars show the changes for different demographic
    subgroups, while black bars show the overall change. Error bars denote 95\%
confidence intervals. Positive values
    indicate increased stereotypical preference after compression. The dashed vertical line
    denotes no change from the base model. All values are reported in
    percentage points.}
    \label{fig:rq3_bbq_llama_subgroup_bias_change}
\end{figure*}

\begin{figure*}[t!]
    \centering
    \includegraphics[width=0.9\textwidth]
    {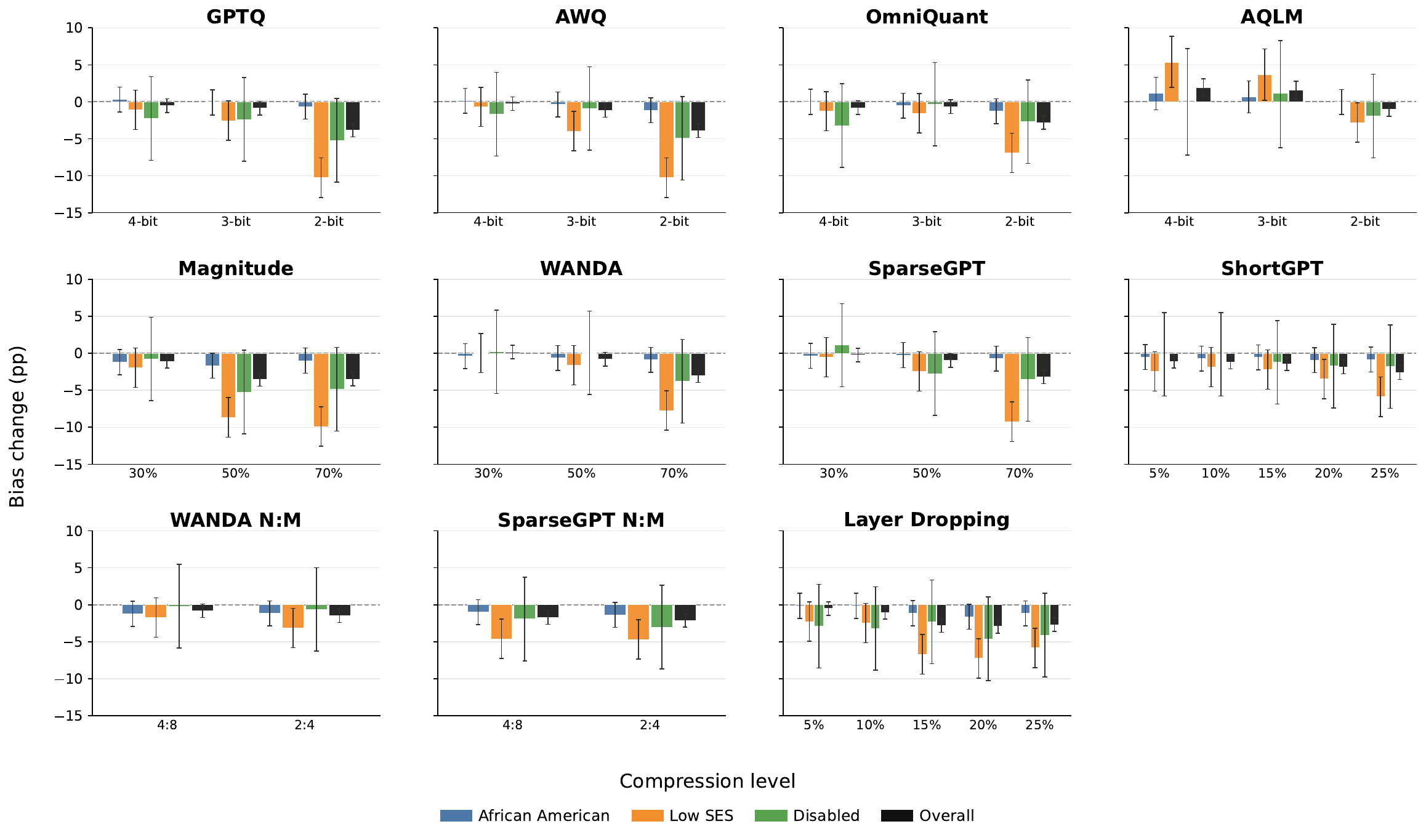}
    \caption{Overall and subgroup-level changes in stereotypical preference on
    BBQ for \texttt{Qwen3-8B} under different compression methods and
    settings. Colored bars show the changes for different demographic
    subgroups, while black bars show the overall change. Error bars denote 95\%
confidence intervals. Positive values
    indicate increased stereotypical preference after compression. The dashed vertical line
    denotes no change from the base model. All values are reported in
    percentage points.}
    \label{fig:rq3_bbq_qwen_subgroup_bias_change}
\end{figure*}

\begin{figure*}[t!]
    \centering
    \includegraphics[width=0.9\textwidth]
    {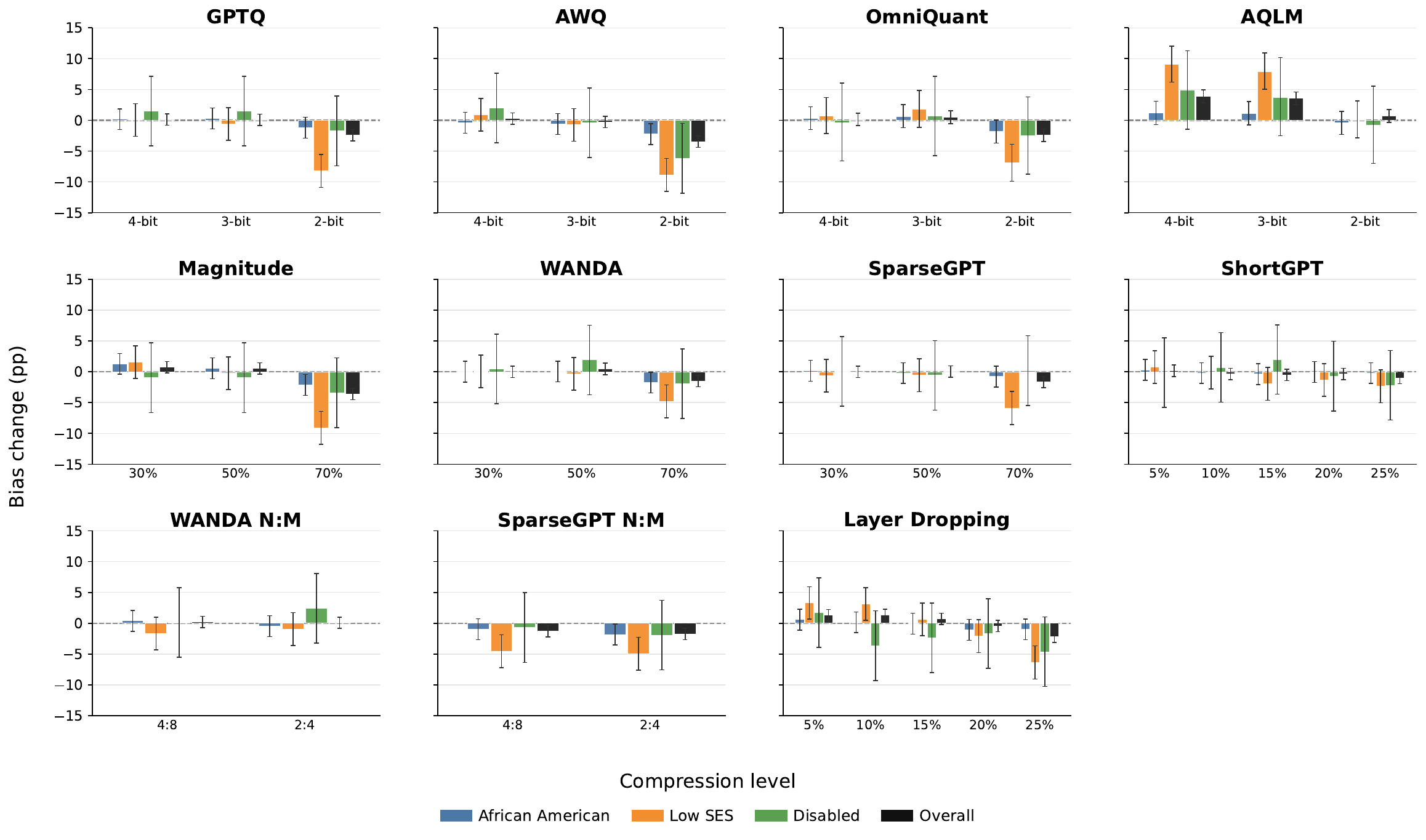}
    \caption{Overall and subgroup-level changes in stereotypical preference on
    BBQ for \texttt{Gemma-2-9B-it} under different compression methods and
    settings. Colored bars show the changes for different demographic
    subgroups, while black bars show the overall change. Error bars denote 95\%
confidence intervals. Positive values
    indicate increased stereotypical preference after compression. The dashed vertical line
    denotes no change from the base model. All values are reported in
    percentage points.}
    \label{fig:rq3_bbq_gemma_subgroup_bias_change}
\end{figure*}

WinoBias and BBQ operationalize stereotypical preference differently. WinoBias focuses on  occupational associations on genders and renders direct comparisons of male and female pronoun groups, whereas BBQ covers multiple social dimensions and fine-grained identity groups. Their values therefore are not interchangeable for the same subgroup construct. Nevertheless, what generalizes across benchmarks is the masking effect of the overall shift.
Figure~\ref{fig:rq3_winobias_subgroup_bias_change} and Figures~\ref{fig:rq3_winobias_llama_subgroup_bias_change_other_methods},~\ref{fig:rq3_winobias_qwen_subgroup_bias_change},~\ref{fig:rq3_winobias_gemma_subgroup_bias_change} extend the WinoBias analysis across compression methods and base models. Figures~\ref{fig:rq3_bbq_llama_subgroup_bias_change},~\ref{fig:rq3_bbq_qwen_subgroup_bias_change},
\ref{fig:rq3_bbq_gemma_subgroup_bias_change} report the corresponding results on BBQ. Since results from collapsed settings have degraded substantially, we place greater interpretation weight on non-collapsed configurations. 

\paragraph{Fine-Grained Subgroup Shifts.}
Tables~\ref{tab:rq3_winobias_largest_subgroup_changes} and
\ref{tab:rq3_bbq_largest_subgroup_changes} report the largest absolute fine-grained subgroup changes.
In WinoBias, the largest changes are predominantly decreases affecting female-coded occupations. For Llama,
\textit{secretary} has the largest absolute change under SparseGPT 2:4, with \(\Delta B_g=-53.1\) pp (95\% CI \([-77.7,-14.1]\)). For Qwen, \textit{sheriff} has the largest absolute change under 50\% magnitude pruning, with \(\Delta B_g=-51.9\) pp
(95\% CI \([-70.7,-24.0]\)). 
SparseGPT N:M accounts for all four displayed Llama entries, whereas 50\% magnitude pruning accounts for four of the six displayed Qwen
entries. For BBQ, the largest changes are predominantly increases concentrated among disability-related identities. The largest displayed shifts for Qwen and Gemma occur under ShortGPT, whereas the largest displayed Llama shifts arise under SparseGPT and
WANDA N:M and occur for cognitive-disability identities.

\paragraph{Model-dependent WinoBias changes.}
The expanded WinoBias results show that the direction and magnitude of bias change depend on the base model, even under the same compression method and nominal setting.
For Llama, the most aggressive GPTQ, AWQ, OmniQuant, Magnitude, WANDA, and SparseGPT settings generally shift the overall score downward, whereas several AQLM configurations shift it upward. Gemma displays a different profile: AQLM again produces positive overall changes, but aggressive WANDA also moves the overall score upward, despite moving it downward for Llama and Qwen. 
At \(70\%\) WANDA sparsity, for example, the observed overall changes are \(-12.9\) pp (95\% CI \([-20.1,-5.5]\)) for Llama,
\(-8.1\) pp (95\% CI \([-16.1,+0.4]\)) for Qwen, and
\(+11.1\) pp (95\% CI \([+7.4,+14.7]\)) for Gemma. 
Although this aggressive setting is best treated as a boundary case because of its severe utility degradation, the sign reversal demonstrates that neither the compression method nor its nominal severity is sufficient to predict the direction of the measured bias change.

\paragraph{Aggregate bias changes conceal non-monotonic subgroup effects}
First, aggregate scores can conceal subgroup-level changes and may even suggest
the opposite direction from that observed for particular subgroups. For
example, Table~\ref{tab:rq3_winobias_largest_subgroup_changes} shows that SparseGPT 4:8 pruning increases Llama's overall stereotypical-preference score by \(2.3\) percentage points, while decreasing the score for the
\textit{Auditor} subgroup by \(50.0\) points. More broadly, a similar aggregate change can arise from markedly different subgroup profiles: the male- and female-pronoun estimates may remain approximately stable, but move together by different magnitudes, or move in opposite directions and partially offset one another. The layer-dropping results further show that these subgroup effects
are not monotonic in compression severity. The relative magnitudes and directions of the male- and female-pronoun shifts fluctuate across layer-dropping rates, with different trajectories for Llama, Qwen, and Gemma.
Compression severity should therefore not be interpreted as a one-dimensional control on stereotypical preference; its effects depend
jointly on the base model, compression configuration, and subgroup.

\paragraph{BBQ subgroup behavior.}
Figures~\ref{fig:rq3_bbq_llama_subgroup_bias_change}--%
\ref{fig:rq3_bbq_gemma_subgroup_bias_change} show the overall BBQ change with the three displayed subgroups: \textit{African American}, \textit{Low SES}, and \textit{Disabled}. Across all plotted configurations, the overall BBQ change ranges from approximately \(-5\) to \(+4\) percentage points and is predominantly negative. The aggregate effect on BBQ is therefore numerically smaller than many of the gender-level movements observed on WinoBias. Nevertheless, a change concentrated within one demographic category contributes only partially to an average computed over the full benchmark. This differs from the most visually direct WinoBias examples, where male- and female-pronoun point estimates move in opposite directions. Therefore, the current BBQ figures support conclusions only at the level of the three displayed aggregates. They do not separately display individual identities within the disability or other social dimensions. 

\paragraph{RQ3 takeaway.}
The additional results provide descriptive evidence that the observed compression-induced changes do not follow a stable directional pattern. Small aggregate changes can coexist with opposing subgroup movements, unequal movements in the same direction, or changes concentrated in a subgroup. Consequently, stronger claims about subgroups require fine-grained estimates accompanied by sample sizes and an analysis that accounts for selecting the largest effect from many candidate subgroups.

\begin{table*}[t]
\centering
\renewcommand{\arraystretch}{1.12}

\begin{tabular*}{\textwidth}{
@{\extracolsep{\fill}}
r
l
l
l
r
r
r
r
@{}
}
\hline
\textbf{Rank}
& \textbf{Gender}
& \textbf{Occupation}
& \makecell[l]{\textbf{Compression}\\\textbf{Configuration}}
& \makecell[r]{\textbf{Base}\\$\boldsymbol{B_g}$ \textbf{(\%)}}
& \makecell[r]{\textbf{Comp.}\\$\boldsymbol{B_g}$ \textbf{(\%)}}
& \makecell[r]{$\boldsymbol{\Delta B_g}$\\\textbf{(pp)}}
& \makecell[r]{$\boldsymbol{\Delta B}$\\\textbf{(pp)}} \\
\hline

\multicolumn{8}{@{}l}{\texttt{Llama-3.1-8B-Instruct}} \\

1
& Female
& Secretary
& SparseGPT N:M (2:4)
& 71.9
& 18.8
& $-53.1$
& $-2.2$ \\

3
& Female
& Auditor
& SparseGPT N:M (4:8)
& 82.1
& 32.1
& $-50.0$
& $+2.3$ \\

8
& Male
& Analyst
& SparseGPT N:M (2:4)
& 50.0
& 2.5
& $-47.5$
& $-2.2$ \\

9
& Female
& Clerk
& SparseGPT N:M (2:4)
& 78.6
& 32.1
& $-46.4$
& $-2.2$ \\

\hline
\multicolumn{8}{@{}l}{\texttt{Qwen3-8B}} \\

2
& Male
& Sheriff
& Magnitude (50\%)
& 59.6
& 7.7
& $-51.9$
& $-8.2$ \\

4
& Female
& Clerk
& Magnitude (50\%)
& 67.9
& 17.9
& $-50.0$
& $-8.2$ \\

5
& Female
& Nurse
& ShortGPT (20\%)
& 77.8
& 27.8
& $-50.0$
& $-4.2$ \\

6
& Male
& Developer
& SparseGPT N:M (2:4)
& 71.9
& 21.9
& $-50.0$
& $-4.0$ \\

7
& Female
& Attendant
& Magnitude (50\%)
& 61.8
& 12.7
& $-49.1$
& $-8.2$ \\

10
& Female
& Counselor
& Magnitude (50\%)
& 66.1
& 19.6
& $-46.4$
& $-8.2$ \\

\hline
\end{tabular*}

\caption{The ten largest absolute occupation-level bias changes on WinoBias
among the retained compression configurations. Base and compressed subgroup
bias scores are reported as percentages, while subgroup changes
$\Delta B_g$ and overall changes $\Delta B$ are reported in percentage
points.}
\label{tab:rq3_winobias_largest_subgroup_changes}
\end{table*}

\begin{table*}[t]
\centering
\setlength{\tabcolsep}{2.4pt}
\renewcommand{\arraystretch}{1.12}

\begin{tabularx}{\textwidth}{
@{}
r
l
>{\raggedright\arraybackslash}X
>{\raggedright\arraybackslash}p{0.22\textwidth}
r
r
r
r
@{}
}
\hline
\textbf{Rank}
& \textbf{Category}
& \textbf{Subgroup}
& \makecell[l]{\textbf{Compression}\\\textbf{Configuration}}
& \makecell[r]{\textbf{Base}\\$\boldsymbol{B_g}$ \textbf{(\%)}}
& \makecell[r]{\textbf{Comp.}\\$\boldsymbol{B_g}$ \textbf{(\%)}}
& \makecell[r]{$\boldsymbol{\Delta B_g}$\\\textbf{(pp)}}
& \makecell[r]{$\boldsymbol{\Delta B}$\\\textbf{(pp)}} \\
\hline

\multicolumn{8}{@{}l}{\texttt{Llama-3.1-8B-Instruct}} \\

8
& Disability
& People with cognitive disabilities or mental illness
& SparseGPT N:M (4:8)
& 40.6
& 59.4
& $+18.8$
& $-1.7$ \\

9
& Disability
& People with cognitive disabilities or mental illness
& WANDA N:M (2:4)
& 40.6
& 59.4
& $+18.8$
& $-3.1$ \\

\hline
\multicolumn{8}{@{}l}{\texttt{Qwen3-8B}} \\

1
& Disability
& Down's syndrome
& ShortGPT (10\%)
& 37.5
& 75.0
& $+37.5$
& $-1.2$ \\

2
& Disability
& Down's syndrome
& ShortGPT (15\%)
& 37.5
& 75.0
& $+37.5$
& $-1.4$ \\

3
& Disability
& Down's syndrome
& ShortGPT (5\%)
& 37.5
& 62.5
& $+25.0$
& $-1.1$ \\

4
& Disability
& People with cerebral palsy
& SparseGPT (50\%)
& 56.3
& 81.3
& $+25.0$
& $-1.0$ \\

10
& Disability
& Down's syndrome
& Magnitude (30\%)
& 37.5
& 56.3
& $+18.8$
& $-1.1$ \\

\hline
\multicolumn{8}{@{}l}{\texttt{Gemma-2-9B-it}} \\

5
& Disability
& Down's syndrome
& ShortGPT (15\%)
& 50.0
& 75.0
& $+25.0$
& $-0.5$ \\

6
& Disability
& Down's syndrome
& ShortGPT (25\%)
& 50.0
& 75.0
& $+25.0$
& $-1.0$ \\

7
& Nationality
& Italian
& WANDA N:M (2:4)
& 57.5
& 37.5
& $-20.0$
& $+0.1$ \\

\hline
\end{tabularx}

\caption{The ten largest absolute subgroup-level bias changes on BBQ among
the retained compression configurations. Base and compressed subgroup bias
scores are reported as percentages, while subgroup changes $\Delta B_g$ and
overall changes $\Delta B$ are reported in percentage points.}
\label{tab:rq3_bbq_largest_subgroup_changes}
\end{table*}

\end{document}